%% file: Rat_Arxiv.tex
\documentclass[11pt]{article} 

\usepackage{url}
\usepackage{mathtools}
\usepackage{fullpage}

\usepackage{amsmath,amssymb,amsthm, amsfonts, bbm}
\usepackage{mathbbol}
\usepackage{mathrsfs}  
\usepackage{multicol}
\usepackage{rotating}
\usepackage{enumitem, comment, xifthen}

\usepackage{xcolor}
\usepackage{booktabs}
\usepackage[mathscr]{euscript}
\usepackage[most]{tcolorbox}

\usepackage[
    backend=bibtex,
    citestyle=alphabetic,
    bibstyle=alphabetic,
    maxalphanames=3,
    minalphanames=1,
    giveninits=true,
    natbib=true
]{biblatex}
\usepackage{algorithmic}
\usepackage{algorithm}

\usepackage{hyperref}[links]
\hypersetup{
  colorlinks,
  linkcolor={red!72!black},
  citecolor={blue!72!black},
  urlcolor={magenta!72!black}
}

\usepackage[nameinlink]{cleveref}

\usepackage{rat_macros}

\newcommand{\PenSmall}{\ensuremath{\Pen_\tnum}}

\newcommand{\kit}{\ensuremath{k}}
\newcommand{\kmax}{\ensuremath{K}}
\newcommand{\Kmax}{\kmax}
\newcommand{\stepsize}{\ensuremath{\eta}}

\newcommand{\funit}[1]{\ensuremath{f^{#1}}}

\newcommand{\LossBarM}{\ensuremath{\LossBar_m}}

\newcommand{\OracleHat}{R_\numtarget}

\newcommand{\LossPL}{\ensuremath{L_\numtarget^{\scaleto{\text{SM}}{4pt}}}}

\newcommand{\resid}{\ensuremath{e}}
\newcommand{\RESCALE}{\ensuremath{\tfrac{1}{m}}}

\newcommand{\PicardIter}[1]{\ensuremath{\ProxAlpha \big( #1 - \stepsize \GradHat(#1) \big)}}

\usepackage{xspace}
\newcommand{\RAT}{\texttt{RaT}\xspace}
\renewcommand{\PL}{\texttt{SM}\xspace}
\newcommand{\vstar}{\ensuremath{v^{*}}}

\newcommand{\TeachMat}{\ensuremath{\mathbf{T}}}
\newcommand{\UseTeachMat}{\TeachMat}

\newcommand{\fhatrat}{\ensuremath{\widehat{f}_{\scaleto{\RAT}{4pt}}}}
\newcommand{\fhatpl}{\ensuremath{\widehat{f}_{\scaleto{\PL}{4pt}}}}

\newcommand{\thetahatrat}{\ensuremath{\widehat{\theta}_{\scaleto{\RAT}{4pt}}}}
\newcommand{\thetahatpl}{\ensuremath{\widehat{\theta}_{\scaleto{\PL}{4pt}}}}

\newcommand{\Cmat}{\ensuremath{\mathbf{C}}}

\newcommand{\var}{\ensuremath{\operatorname{var}}}

\newcommand{\MyGrad}{\ensuremath{\nabla \LossBar_\numtarget}}

\newcommand{\Yspace}{\ensuremath{\mathcal{Y}}}

\newcommand{\LossBarQ}{\ensuremath{\LossBar_{\Qprob}}}

\renewcommand{\Xtil}{\tilde{X}}

\renewcommand{\PhiMat}{\ensuremath{\mathbf{\Phi}}}
\renewcommand{\PhiMatTil}{\ensuremath{\tilde{\PhiMat}}}

\newcommand{\SigMat}{\ensuremath{\mathbf{\Sigma}}}

\newcommand{\NewBiasPL}{\ensuremath{B^2_{\PL}(\fdagger
    + \gstar)}}
\newcommand{\NewBiasRat}{\ensuremath{B^2_{\RAT\:}(\gstar)}}
\newcommand{\NewVarRat}{\ensuremath{V_{\RAT\:}}}

\newcommand{\myfigdir}{.}

\newcommand{\stureg}{\ensuremath{\gamma}}

\newcommand{\KerFunTil}{\ensuremath{\tilde{\KerFun}}}

\newcommand{\newnoise}{\ensuremath{v}}
\newcommand{\newnoisepl}{\ensuremath{\newnoise_{\scaleto{\PL}{4pt}}}}
\newcommand{\newnoiserat}{\ensuremath{\newnoise_{\scaleto{\RAT}{4pt}}}}

\newcommand{\Kerfun}{\KerFun}

\newcommand{\NewMat}{\Cmat}
\newcommand{\regpar}{\ensuremath{\lambda}}
\newcommand{\bandwidth}{\ensuremath{h}}

\newcommand{\dostu}[1]{\ensuremath{#1^{\; \stureg}}}
\newcommand{\fhatratstu}{\dostu{\fhatrat}}
\newcommand{\fhatplstu}{\dostu{\fhatpl}}

\newcommand{\MSE}{\operatorname{MSE}}
\newcommand{\frobnorm}[1]{\ensuremath{|\!|\!| #1 |\!|\!|_{\operatorname{F}}}}

\newcommand{\cov}{\ensuremath{\operatorname{cov}}}

\newcommand{\PLBiasSquared}{B_{\PL}^2}

\newcommand{\hs}{\hspace*{0.02in}}
\newcommand{\FinRatBias}{B_{\RAT}}
\newcommand{\FinRatVar}{V_{\RAT}}

\newcommand{\Gteacher}{\Gclass}
\newcommand{\Fstudent}{\Fclass}

\newcommand{\FstudentClass}{\Fclass_{\scaleto{\mbox{stud}}{4pt}}}
\newcommand{\GteacherClass}{\Gclass_{\scaleto{\mbox{teach}}{4pt}}}

\newcommand{\kteacher}{\ensuremath{k}}
\newcommand{\kstudent}{\ensuremath{\tilde{k}}}

\newcommand{\TV}{\operatorname{TV}}

\newcommand{\imsize}{0.12\textwidth}

\newcommand{\imfigsize}{0.4\textwidth}

\newcommand{\SM}{\PL}

\newcommand{\zbar}{z^{k+1}}
\newcommand{\kstar}{k^*}
\newcommand{\vhat}{\ensuremath{\widehat{v}}}
\newcommand{\vtau}{\ensuremath{v^\tau}}

\begin{document}

\begin{center}
  {\bf{\LARGE

      \input{title_rat.tex} 
      }}

\end{center}

  \begin{center}
    \begin{tabular}{ccc}
      Kakei Yamamoto$^\dagger$ && Martin
      J. Wainwright$^\star$\\ \texttt{kakei@mit.edu} &&
      \texttt{mjwain@mit.edu}
    \end{tabular}

\vspace*{0.2in}
    \begin{tabular}{c}
      Lab for Information and Decision Systems \\ Statistics and Data
      Science Center \\ EECS$^{\dagger, \star}$ and
      Mathematics$^\star$ \\ Massachusetts Institute of Technology
    \end{tabular}
  
  \vspace*{0.25in} \today
  \vspace*{0.25in}

  \begin{abstract}
    \input{abstract_rat.tex}
  \end{abstract}
\end{center}


\section{Introduction}
\label{SecIntro}

The student--teacher paradigm, in which one predictive model is used
to guide the training of another, has become increasingly prevalent
over the past decade.  It has proven useful for a variety of problems,
including semi-supervised learning, model compression and
distillation, as well as adaptation to distribution shift.  The core
idea in all of these settings is to use the predictions of a
pre-trained teacher model, often relatively complex and/or opaque in
nature, to train a student model that might be more computationally
efficient, easier to interpret, or better behaved on a new covariate
population.

In broad terms, the idea of model distillation is relatively old,
emerging alongside the rise of neural networks, boosting, and ensemble
methods in the 1990s.  It has attracted renewed interest over the past
decade, fueled by the explosion in the complexity of prediction
methods, most notably with deep neural networks. Complex models can
generate high-quality predictions, but may lack interpretability, or
require significant memory and computation to generate predictions.  A
natural idea, then, is to train a simpler model (the student) to mimic
the predictions of the more complex model (the teacher). This form of
student-teacher interaction has led to a substantial and evolving line
of work
(e.g.,~\cite{BreSha96,BucCarNic06,BaCar14,HinVinDea15,FroHin17,FurEtAl18,VidEtAl20}).
A notable early example is the ``born-again tree'' of Breiman and
Shang~\cite{BreSha96}, where a single decision tree is used to
approximate the predictions of a tree ensemble; see also the
contemporaneous work~\cite{craven1996extracting}, as well as the
follow-up papers~\cite{FurEtAl18,VidEtAl20}.  Bucila et
al.~\cite{BucCarNic06} formalized the idea of model compression in
more general settings, and studied how to generate synthetic
covariates.  For classification problems, there is a choice between
having the student match the teacher's predicted class labels, known
as ``hard'' information, versus ``soft'' information such as predicted
probabilities or logits.  For classification, Hinton et
al.~\cite{HinVinDea15} proposed minimizing the KL divergence to the
teacher's probabilities, and this use of soft information has proven
to be superior in general~\cite{FroHin17,arazo2020pseudo}.
Student--teacher interactions are also used to address the problem of
covariate shift: here a teacher trained on a source distribution is
used to guide training on a (potentially different) target
distribution.  This approach appears in domain adaptation and
semi-supervised learning, where teacher predictions provide auxiliary
information to the student, especially relevant in regions of the
covariate space lacking responses or labels.  Finally, on a
contemporary note, model distillation was an important component of
training the DeepSeek large-language model~\cite{deepseek2025r1}, and
OpenAI alleges that ChatGPT was used as a teacher.

\paragraph{Soft matching and teacher biases:} 
In standard student--teacher approaches, the student is trained to
match the teacher’s outputs, either by least-squares (for real-valued
outputs), or cross-entropy loss (for probabilities arising from
classification problems). We refer to direct imitation procedures of
this type as \emph{student (soft) matching} (\PL).  When the teacher's
predictions are accurate, then direct imitation by the student can
yield strong performance. In particular, as observed from the earliest
work~\cite{BreSha96}, a student trained on a teacher's outputs can
outperform a student trained on the original data, since the teacher
effectively denoises the response data.  On the other hand, if the
teacher’s predictions are biased or systematically incorrect, then the
\PL approach propagates these errors to the student. As a result, even
with abundant data, the student can inherit persistent prediction
error from the teacher. This undesirable phenomenon has been referred
to as confirmation bias, and the focus of this paper is new
methodology and theory for tackling it.

Many classes of complex predictive models are known to exhibit
particular biases. For examples, tree-based
methods~\cite{CARTBook,breiman2001random} and stump-based boosting
methods~\cite{breiman1998arcing,friedman2001greedy,chen2016xgboost}
allow for step-function discontinuities, but with a strong preference
for axis-aligned changes.  On the other hand, local smoothing
methods~\cite{watson1964smooth,nadaraya1964estimating,fan1996local}
are strongly biased against discontinuity, but without any axis
preference.  Regression procedures based on reproducing kernel Hilbert
spaces~\cite{Gu02,Wahba,Wai19} use explicit penalization, which
induces bias depending on the covariate distribution and kernel
function.  Shallow neural networks using rectified linear units
(ReLUs) are biased towards functions with low oscillation, easily
expressible sums of piecewise linear
functions~\cite{telgarsky2016benefits}.  Deep neural networks, with
many layers of hidden units, also exhibit various forms of inductive
bias, notably a spectral bias towards lower frequency components of
functions~\citep{rahaman2019spectral,basri2020frequency,xu2020frequency}.

Any such teacher biases can degrade the quality of the trained
student, and various heuristics have been explored to mitigate this
effect, including temperature scaling of the teacher's
outputs~\cite{HinVinDea15}, confidence-based
filtering~\cite{arazo2020pseudo}, noise injection
schemes \cite{XieLuHovyLe20,arazo2020pseudo}, and weighting and
ensembling methods~\cite{TarVal17}. All of these approaches seek to
mitigate the effect of teacher biases, but do not alter the underlying
imitation objective.

\paragraph{Residual-as-teacher:}  In this paper, we propose and
analyze an alternative mechanism for leveraging teacher predictions
that is designed to mitigate bias.  Instead of applying the teacher to
generate fresh responses for the student, we instead use the teacher
to estimate the \emph{errors} in the student's predictions.  This
change naturally leads to an iterative procedure, known as the
\emph{residual-as-teacher} (\RAT) algorithm, in which the student
model is successively refined by the teacher's feedback.  Although
this residual feedback echoes boosting-style
updates~\cite{friedman2000additive,friedman2001greedy,buhlmann2003boosting,zhang2005boosting},
the \RAT procedure does not construct a stagewise additive model;
rather, our analysis shows that the \RAT algorithm is a surrogate for
a proximal update scheme over the student function class, with fixed
points corresponding to the optima of an oracle penalized optimization
problem (cf. equation~\eqref{EqnDefnFdagger}).  Moreover, we show
through a combination of theoretical analysis and empirical study that
\RAT estimates are superior to the \SM approach in mitigating the
teacher's biases.

\begin{figure}[h!]
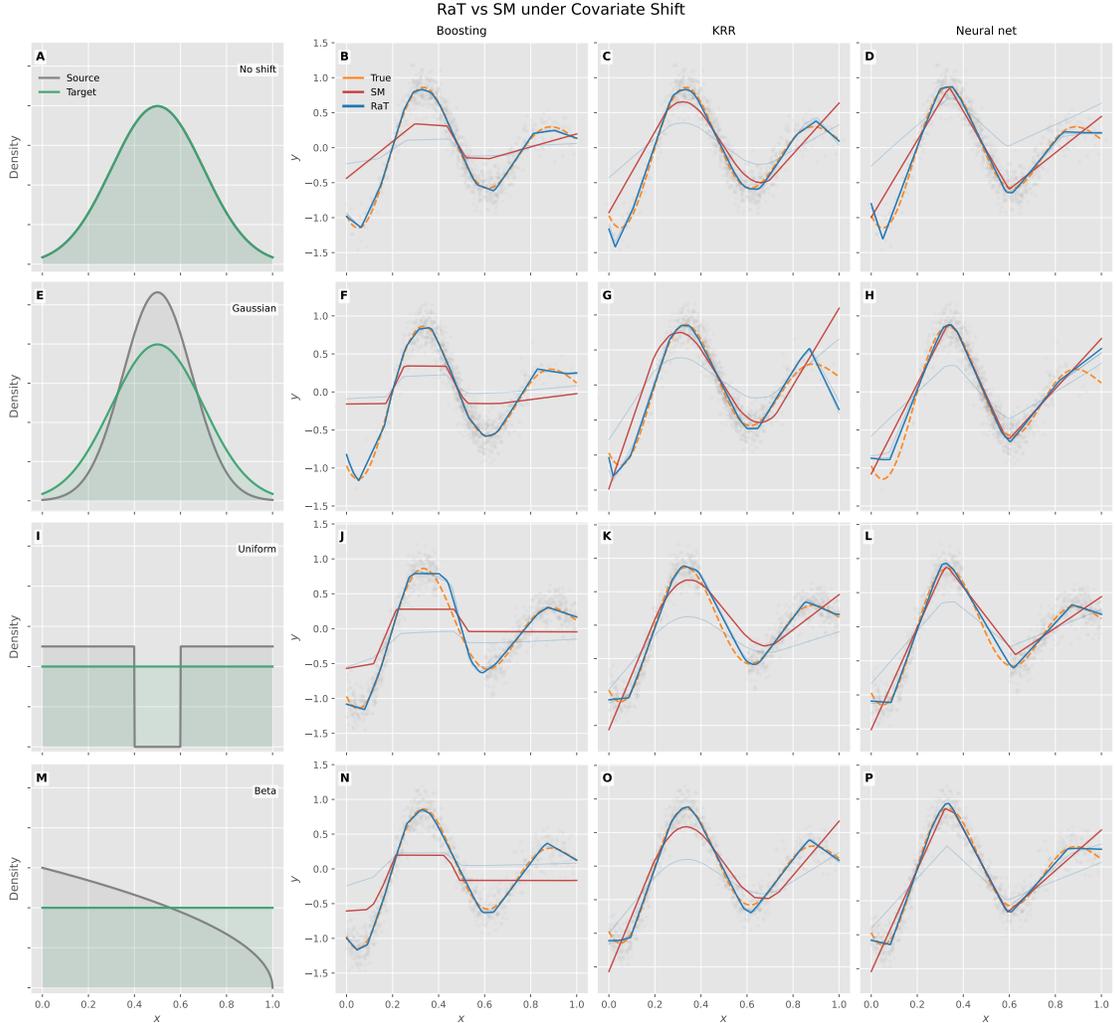

\begin{center}
    \widgraph{0.9\textwidth}{\myfigdir/fignew_intro_rat_vs_pl}
    \caption{Comparison of the \RAT and \PL estimates when the student
      class $\Fclass$ is a two-layer neural network with $128$ hidden
      units.  Shown are results for four different covariate shifts
      (rows 1--4), and three different teacher models (columns 2--4).
      Each row corresponds to a different source–target distribution
      pair, shown in the leftmost column via their marginal
      densities. The remaining columns report results for three
      teacher classes: Boosting, Kernel ridge regression (KRR), and
      ReLU neural network fits. In each panel, the true regression
      function is shown as a dotted orange line, while the estimates
      obtained by \PL and \RAT are shown in red and blue,
      respectively. Gray points indicate source samples. Intermediate
      \RAT iterates are shown as faint curves to illustrate the
      refinement process.}
    \label{FigRATvsPL}
\end{center}
\end{figure}

As a preview of our results, \Cref{FigRATvsPL} illustrates the
difference between student matching (\SM) and the \RAT approach for a
simple one-dimensional regression problem based on the least-squares
loss.  In all cases, the student is a two-layer ReLU neural network
with $h = 128$ hidden units, and is trained on the target data via
least-squares regression.  The teacher is fit on source data only,
and~\Cref{FigRATvsPL} shows results for three different teachers:
\begin{itemize}[leftmargin=*,itemsep=0pt]
\item gradient boosting with depth $2$ trees, and $8$ rounds of boosting;
\item kernel ridge regression using a Gaussian kernel function with
  bandwidth $\sigma = 0.5$ and regularization parameter $\gamma = 3$, and
\item a two-layer ReLU neural network with $10$ hidden units.
\end{itemize}
In all cases, these teacher classes exhibit significant forms of bias:
the boosting scheme involves shallow trees with limited rounds; the
KRR procedure has a fixed and overly large regularization parameter;
and the small number of hidden units in the ReLU network induces bias,
since the ReLU units are piecewise-linear.  The \SM procedure trains
the student to match the teacher's predictions (see~\Cref{SecPseudo}
for a precise description), whereas \RAT iteratively fits the teacher
to the student's residuals on source data, and uses these residual
estimates to refine the student (see~\Cref{SecRat} for the exact
procedure).  The resulting differences are quite clear in the figure:
\SM tends to inherit systematic teacher bias, while \RAT progressively
corrects it.  The analysis of this paper provides a firm theoretical
grounding for these empirical observations.

\paragraph{Our contributions:}  In this paper, we introduce the \RAT
procedure and analyze it in detail, including both the statistical
properties of fixed points, and the computational properties of the
iterative updates themselves.  In~\Cref{ThmStat}, we prove
non-asymptotic upper bounds on the excess risk of the \RAT estimate;
apart from leading to statistical consistency and rate guarantees,
these bounds elucidate the way in which the teacher's bias enters the
estimates.  We also prove a related result (\Cref{PropPL}) that
provides risk bounds for the student soft-matching (\SM) estimator,
and reveals that the teacher's bias enters in an entirely different
way.  This suggests that there should be a fundamental gap between the
two procedures, and \Cref{ThmSeparation} formalizes this intuition.
We consider student--teacher pairs based on kernel ridge regression
(KRR) estimators, in which the teacher has a fixed bias whereas the
student estimator can be tuned, and there is covariate shift, meaning
that the teacher and student observe covariates from different
distributions.  By suitable tuning, despite the teacher bias and
covariate shift, the \RAT estimator is able to achieve the
minimax-optimal risk, while the \SM estimator has prediction error
lower bounded by a universal constant.  In~\Cref{ThmComp}, we study
the convergence properties of iterative schemes designed to compute
the \RAT fixed point.  Finally, we complement our theoretical results
with numerical studies on both synthetic data, and on
covariate-shifted versions of ImageNette dataset for image
classification.

\paragraph{Paper organization:}  The remainder of this paper is organized as follows.
We begin in~\Cref{SecSetup} by setting up the student--teacher
distillation more precisely, allowing for the possibility of covariate
shift between student and teacher.  \Cref{SecMain} is devoted to
statements of our main theoretical results, including excess risk
bounds (\Cref{ThmStat} in~\Cref{SecRisk}); comparison with student
matching (\Cref{SecPseudo}); a separation result (\Cref{ThmSeparation}
in~\Cref{SecSeparation}); and algorithmic guarantees (\Cref{ThmComp}
in~\Cref{SecComp}).  In~\Cref{SecNumerical}, we provide numerical
experiments that support our theory.  \Cref{SecNumSeparation} gives
results on the separation between \SM/\RAT for synthetic problems,
whereas we report results for a covariate-shifted version of the
ImageNette dataset in~\Cref{SecNumImage}.  Proofs of our three main
theorems are given in~\Cref{SecProofs}, with certain more technical
calculations given in the appendices; we prove all the remaining
propositions and corollaries in the appendices as well.


\section{Problem and method formulation}
\label{SecSetup}

In this section, we begin in~\Cref{SecProblem} with a more precise
specification of the student--teacher estimation problem of interest,
before describing the residual-as-teacher estimator that we analyze in
this paper (\Cref{SecRat}).

\subsection{Target/source data and student/teacher pairs}
\label{SecProblem}

We consider a general prediction problem, involving a covariate or
feature vector $x \in \Xspace$, and a response or output $y
\in \Yspace$.  When $y$ is real-valued, the prediction problem is a
form of regression, whereas when $y$ takes discrete values, we are
solving a classification problem.  Our goal is to learn a mapping
$f: \Xspace \rightarrow \Yspace$ such that $f(x)$ is a ``good
predictor'' of the response $y$.  As usual, we formalize the quality
of $f$ in terms of a scalar-valued \emph{loss function} $\ell(f(x),
y)$.  Standard examples include the least-squares loss $\ell(f(x), y)
= (1/2) \big( f(x) - y \big)^2$ for $y \in \real$, and the logistic
regression loss $\ell(f(x), y) = \log(1 + e^{f(x)}) - y f(x)$ for
binary labels $y \in \{0,1 \}$. \\

\vspace*{-0.02in}
\noindent At the core of the set-up are two pairs:
\begin{itemize}[leftmargin=*,itemsep=0pt, topsep=3pt]
\item the student--teacher pair of (potentially different) prediction
  classes $\FstudentClass \equiv \Fclass$ and $\GteacherClass \equiv
  \Gclass$, used to construct functions mapping covariates to
  predictions, and
\item the target--source pair $(\Qprob_X, \Prob_X)$ of (potentially
  different) distributions over the covariates used by the student and
  teacher, respectively.
\end{itemize}
In the standard approach to covariate shift, there is no notion of
student--teacher pair, whereas in the typical student--teacher set-up,
there is no covariate shift.  The set-up described here allows for
both possibilities. \\

The \emph{source data} consists of labeled pairs $\{(x_i, y_i)
\}_{i=1}^\numobs$ drawn from an unknown distribution \mbox{$\Prob_X
  \times \Prob_{Y \mid X}$} over the joint space $\Xspace
\times \Yspace$.  Second, the \emph{target data} consists of target
covariates $\{\xtil_j \}_{j=1}^m$ drawn according to $\Qprob_X$.
Given these two datasets, the \emph{student} operates over the
function class $\FstudentClass$, and has the goal of learning a
function $f \in \FstudentClass$ that performs well in predicting at
the target covariates $\{\xtil_j \}_{j=1}^m$.  On the other hand, the
\emph{teacher} is defined by a function class $\GteacherClass$, and
implements empirical risk minimization (ERM) procedures based on the
source data.

\paragraph{Smoothed target risk:} 
For the bulk of this paper, we condition on the target covariates
$\{\xtil_j \}_{j=1}^m$, and we measure the student's performance in
terms of the \emph{smoothed target risk}, given by
\begin{subequations}
\begin{align}
  \label{EqnSmoothedRisk}
  \LossBarM(f) & = \sum_{j=1}^\numtarget \Exs_{Y} \big[
    \ell(f(\xtil_j), Y) \mid X = \xtil_j \big],
\end{align}
where we take expectations over each conditional distribution $Y \mid
X = \xtil_j$.  When the samples $\xtil_j$ are drawn i.i.d. from
$\Qprob_X$, we observe that $(1/\numtarget) \LossBarM(f)$ is an
unbiased estimate of the population quantity
\begin{align}
\LossBarQ(f) & \defn \Exs_{(X,Y) \sim \Qprob_X \times \Prob_{Y \mid
    X}} \big[ \ell(f(X), Y) \big].
\end{align}
\end{subequations}
Consequently, if we can bound $\LossBarM(\fhat)$ for an estimate
$\fhat$, then we can use standard empirical process theory
(e.g.,~\cite{vandeGeer,vanderVaart96,Wai19}) to obtain bounds on
$\LossBarQ(\fhat)$.  For this reason, we focus primarily on
$\LossBarM$ in our analysis.

\paragraph{Student oracle estimand:}

In many applications, the natural object to minimize is a constrained
or regularized version of the smoothed target risk $\LossBarM$.  In
particular, let $\Pen: \Fclass \rightarrow \real$ be some penalty
function defined on the student class $\Fclass$.  We focus on
approximating the best regularized estimate
\begin{align}
\label{EqnDefnFdagger}  
  \fdagger & \defn \arg \min_{f \in \Fclass} \big \{ \LossBarM(f) +
  \Pen(f) \big \},
\end{align}
which we refer to as the \emph{student oracle estimand}.

One special case of penalty function $\Pen$ is a $\{0,
\infty\}$-valued function for membership in some subset
$\widetilde{\Fclass} \subset \Fclass$ of the full student class.  In
this case, the student oracle estimand $\fdagger$ corresponds to a
projection operation, under the smoothed target risk $\LossBarM$, onto
the constraint set $\widetilde{\Fclass}$.  As a simple but concrete
example, consider functions that are linear in some feature map $x
\mapsto (\varphi_1(x), \ldots, \varphi_k(x)) \in \real^k$, say of the
form $f_\theta(x) = \sum_{a=1}^k \theta_a \varphi_a(x)$ for some
weight vector $\theta \in \real^k$.  In this case, the full student
class takes the form $\Fclass \defn \big \{f_\theta \mid \theta \in
\real^d \}$, and the penalty function could define projection onto the
sparse sub-class $\widetilde{\Fclass} \defn \{f_\theta \in \Fclass
\mid \|\theta\|_1 \leq R \}$ for some radius $R$.  This formulation is
useful when the goal is to discover student functions that are more
interpretable, in that they use some highly relevant subset of all
possible features.  Our general formulation~\eqref{EqnDefnFdagger}
encompasses this particular case, as well as a wide range of other
examples.

\subsection{Residual-as-teacher}
\label{SecRat}

Having defined the target estimand $\fdagger$, now let us describe the
\RAT estimator analyzed in this paper.  The estimate is defined as a
fixed point of an operator that can be decomposed into two steps: (i)
a proximal update implemented by the student, and (ii) estimation of
residuals by the teacher.  We describe each of these in turn, and then
define the notion of a \RAT fixed point, as well as a natural
procedure (Picard iteration) for computing a fixed point.

\paragraph{Student proximal update and oracle consistency:}
We begin with the proximal update associated with the student.  For a
stepsize $\stepsize > 0$ and vector $u \in \real^m$, we define the
\emph{proximal mapping} $u \mapsto \ProxAlpha(u) \in \Fclass$ via
\begin{align}
\label{EqnStudentProximal}  
\ProxAlpha(u) & \defn \arg \min_{f \in \Fclass} \Big \{ \frac{1}{2
  \stepsize} \|u - f(\xtil_1^m) \|_2^2 + \Pen(f) \Big \} \quad
\mbox{where $f(\xtil_1^m) \defn \big(f(\xtil_1), \ldots, f(\xtil_m)
  \big) \in \real^m$.}
\end{align}
An important fact is that this update operator is directly related to
the smoothed target risk~\eqref{EqnSmoothedRisk}, and the student
oracle estimand~\eqref{EqnDefnFdagger}, as we now describe.  In
particular, define the functional gradient vector $\nabla \LossBarM(f)
\in \real^m$ with components
\begin{align}
\label{EqnDefnSFG}  
  \big[ \nabla \LossBarM(f) \big]_j & \defn \Exs_{Y} \Big[
    \frac{\partial}{\partial z} \ell(z, Y) |_{z = f(\xtil_j)} \mid X =
    \xtil_j \Big].
\end{align}
Under convexity conditions, for any $\stepsize > 0$, the student
oracle estimand $\fdagger$ defined in equation~\eqref{EqnDefnFdagger}
satisfies the fixed point relation:
\mygraybox{%
  \begin{flalign}
\label{EqnFdaggerFix}    
\textbf{Oracle self-consistency:}\qquad & \qquad \qquad \qquad
\fdagger \; \; = \; \; \underbrace{\ProxAlpha\big(\fdagger(\xtil_1^m)
  - \stepsize \nabla \LossBarM(\fdagger)\big)}_{\equiv
  \Top_\stepsize(\fdagger)} &&
\end{flalign}
}
\noindent This consistency condition follows by analysis of a proximal
gradient scheme for computing the minimizer~\eqref{EqnDefnFdagger}.
See~\Cref{AppConsistency} for the details. \\

The fixed point relation~\eqref{EqnFdaggerFix}, while of conceptual
value, is not practically useful, since the oracle functional gradient
$\nabla_f \LossBarM(f)$ cannot be computed: it depends on expectations
over the unknown conditional distribution $Y \mid X = \xtil_j$.
However, this gradient can be approximated by applying the teacher
class to the student's empirical residuals, as we now describe.

\paragraph{Estimation of residuals:}
Given a student function $f \in \Fclass$, the \emph{empirical
residual} associated with the source sample $(x_i, y_i)$, is given by
\begin{subequations}
\begin{align}
 \label{EqnDefnResidual}
\underbrace{\resid(f(x_i), y_i)}_{\equiv \resid_i} \defn
\frac{\partial}{\partial z} \ell(z, y_i) |_{z = f(x_i)} \qquad
\mbox{for $i = 1, 2, \ldots, \numobs$.}
\end{align}
Using the residuals $\{\resid_i \}_{i=1}^\numobs$ as response targets,
we solve the least-squares regression
\begin{align}
\label{EqnResidRegression}  
\ghat & \in \arg \min_{g \in \Gclass} \Big \{ \sum_{i=1}^\numobs
\big(\resid_i - g(x_i) \big)^2 \Big \},
\end{align}
thereby obtaining the best-fitting teacher $\ghat \in \Gclass$ to the
student's residuals.  Evaluating this estimate over the target samples
$\{\xtil_j\}_{j=1}^m$ yields the $m$-vector
\begin{align}
  \label{EqnFuncGradSurrogate}
[\GradHat(f)]_j \defn \ghat(\xtil_j) \qquad \mbox{for $j = 1, \ldots,
  \numtarget$.}
\end{align}
\end{subequations}

\paragraph{Residual-as-teacher estimate:}  We have defined two operators:
the student proximal update~\eqref{EqnStudentProximal}, and the
mapping~\eqref{EqnFuncGradSurrogate} from student residuals to
$\GradHat(f)$.  The composition of these two steps defines an operator
on the student function space $\Fclass$, and for a stepsize $\stepsize
> 0$, we define the \RAT estimate in terms of the fixed point
relation:
\mygraybox{%
  \begin{flalign}
\label{EqnFhatRatFix}  
\textbf{\RAT self-consistency:}\qquad & \qquad \qquad \qquad \fhatrat
\; \; = \; \; \underbrace{\ProxAlpha \big( \fhatrat(\xtil_1^m) -
  \stepsize \GradHat(\fhatrat) \big)}_{\equiv
  \TopHat_\stepsize(\fhatrat)}.  &&
\end{flalign}
}
As discussed in~\Cref{AppConsistency}, this definition is sensible in
that fixed points exist under relatively mild conditions, and
moreover, that the set of fixed points do \emph{not} depend on the
stepsize $\stepsize > 0$.  In particular, if a function $\fhatrat$
satisfies the fixed point relation~\eqref{EqnFdaggerFix} for some
stepsize $\stepsize >0$, then the fixed point relation holds for any
stepsize $\tilde{\stepsize} > 0$.  See~\Cref{SecRatFix} for details.

In certain cases, the \RAT fixed point is unique, and can be computed
in closed form; see the discussion in~\Cref{SecSeparation} for a broad
class of examples based on kernel student--teacher pairs.  In the
general setting, the fixed point needs to be computed by iterating
between the student--teacher updates.  The simplest such scheme is
given by the following form of Picard iteration.  Given an initial
choice $f^0 \in \Fclass$ of student function, it generates a sequence
of student functions $\{f^k\}_{k \geq 0}$ as follows:

\mygraybox{\textbf{Proximal \RAT algorithm:} Given stepsize $\stepsize
  > 0$ and initial student function $f^0$, repeat for $k = 0, 1, 2,
  \ldots, K$:
  \begin{enumerate}
  \item[(1)] Train teacher to predict
    residuals~\eqref{EqnResidRegression}, and evaluate $\GradHat(f^k)
    \in \real^m$ via equation~\eqref{EqnFuncGradSurrogate}.
  \item[(2)] Perform the approximate proximal gradient update:
    \begin{align}
\label{EqnPicardProximal}      
      f^{k+1} & = \ProxAlpha \Big( f^k(\xtil_1^m) - \stepsize
      \GradHat(f^k) \Big) \qquad \mbox{where $f^k(\xtil_1^m) \equiv
        (f^k(\xtil_1), \ldots, f^k(\xtil_m))$.}
    \end{align}
  \end{enumerate}
}
\noindent We analyze the convergence of these updates, including the
choice of stepsize, in~\Cref{SecComp}; in particular,
see~\Cref{ThmComp}. \\

\paragraph{Connection to boosting:}  The \RAT updates are related to boosting
procedures~\cite{friedman2000additive,friedman2001greedy}, because
both rely on regression to estimate the residuals of the current fit.
However, the goals of \RAT and boosting are fundamentally different:
the \RAT procedure seeks to approximate the fixed point $\fdagger$,
the best penalized fit within the student class, whereas boosting is a
stage-wise procedure for generating a sequence of additive fits.  In a
boosting algorithm, the effective function class grows with the number
of iterations, so that its statistical behavior depends critically on
controlling this growth, typically via early
stopping~(e.g.,~\cite{buhlmann2003boosting,zhang2005boosting,RasWaiYu14,WeiYanWai19}).
In contrast, at each round, the \RAT
procedure~\eqref{EqnPicardProximal} applies a proximal update that
projects each iterate back into the fixed student class
$\Fclass$. Consequently, all iterates remain within $\Fclass$, so that
the function complexity remains fixed.  Moreover, the RAT estimate
$\fhat$ itself can characterized as a fixed
point~\eqref{EqnFhatRatFix}, independent of any algorithm used to
obtain it.  Overall, \RAT is not a sequential procedure for additive
modeling, but instead a teacher-aided way to approximate a proximal
fixed point over the fixed class $\Fclass$.  As our analysis shows,
its statistical performance governed by the accuracy of the
teacher-induced gradient combined with the complexity of the student
class.

\section{Main results}
\label{SecMain}

Having set up the \RAT procedure, we are now ready to state our main
results.  We begin in~\Cref{SecRisk} by discussing the statistical
properties of the \RAT estimate~\eqref{EqnFhatRatFix}, before
comparing and contrasting it with the student soft-matching (\SM)
procedure in~\Cref{SecPseudo}.  \Cref{SecSeparation} analyzes the \SM
and \RAT methods in detail for kernel-based student--teacher pairs,
and reveals a significant performance gap. Finally, in~\Cref{SecComp},
we give computational guarantees on the proximal \RAT algorithm.


\subsection{Risk bounds relative to oracle}
\label{SecRisk}

In this section, we derive bounds on the excess risk and the
estimation error of any \RAT fixed point $\fhatrat$, as defined by the
condition~\eqref{EqnFhatRatFix}, relative to the oracle minimizer
$\fdagger$, as defined by equation~\eqref{EqnDefnFdagger}.  In stating
these results (and throughout the paper), for a real-valued function
$f: \Xspace \rightarrow \real$ on the covariate space, and a vector $u
\in \real^m$, we adopt the shorthand notation
\begin{align*}
  \qempinner{f}{u} \defn \frac{1}{m} \sum_{j=1}^m f(\xtil_j) u_j \quad
  \mbox{and} \quad \qempnorm{f}^2 \defn \frac{1}{m} \sum_{j=1}^m
  f^2(\xtil_j).
\end{align*}

\mygraybox{
\begin{theorem}
\label{ThmStat}    
Suppose that oracle risk function $\Oracle(f) = \RESCALE \big \{
\LossBar_m(f) + \Pen(f) \big \} $ is convex in the fitted values, and
let $\fdagger$ be a minimizer.  Then for any \RAT fixed point
$\fhatrat$:
\begin{enumerate}
\item[(a)] The \RAT excess risk is bounded as
\begin{subequations}      
\begin{align}
\label{EqnRiskBound}  
\Oracle(\fhatrat) - \Oracle(\fdagger) & \leq \qempinner{\fhatrat -
  \fdagger}{\nabla \LossBar_m(\fhatrat) - \GradHat(\fhatrat)}.
\end{align}
\item[(b)] When $\LossBar_\numtarget$ is $\scon$-strongly convex, any
  \RAT estimate satisfies the bound
\begin{align}  
\label{EqnEstBound}  
  \qempnorm{\fhatrat - \fdagger}^2 & \leq \frac{1}{\scon}
  \qempinner{\fhatrat - \fdagger}{\MyGrad(\fhatrat) - \GradHat(\fhatrat)}.
\end{align}
\end{subequations}
\end{enumerate}
\end{theorem}
}

\noindent See~\Cref{SecProofThmStat} for the proof of this claim. \\

Recall that both the oracle estimate $\fdagger$ and the \RAT fixed
point $\fhatrat$ are defined by fixed point relations, namely
equations~\eqref{EqnFdaggerFix} and~\eqref{EqnFhatRatFix},
respectively.  With this context, Theorem~\ref{ThmStat} can be
understood as a stability result for proximal fixed points: it shows
that the statistical accuracy of any \RAT solution can be controlled
via the accuracy of the teacher-induced gradient operator at that
solution. In particular, it reduces the statistical analysis of \RAT
to bounding the gradient estimation error, without requiring analysis
of an iterative procedure used to compute the fixed point.  Notably,
the bound depends only on the gradient error evaluated at the fixed
point, rather than uniformly over the function class; this fact is
critical for obtaining sharp rates in later sections.

\paragraph{Specialization to least-squares:} As an important special
case, \Cref{ThmStat} has implications for the least-squares loss
$\ell(f(x), y) = (1/2) \big(f(x) - y \big)^2$.  It is $\scon$-strongly
convex with $\scon = 1$, so that that the bound~\eqref{EqnEstBound} is
in force.  In this case, the smoothed functional gradient at $f(x)$ is
given by $f(x) - \Exs[Y \mid X = x]$, so that the oracle gradient has
components
\begin{align*}
[\MyGrad(\fhatrat)]_j & = \fhatrat(\xtil_j) - \fstar(\xtil_j) \qquad
\mbox{where $\fstar(x) = \Exs[Y \mid X = x]$.}
\end{align*}
The teacher $\GradHat(\fhatrat)$ is trained via regression of the
source residuals $\{\resid_i\}_{i=1}^\numobs$ onto the source
covariates $\{x_i\}_{i=1}^n$. For the least-squares loss, the source
residuals are given by
\begin{align*}
  \resid_i & = \fhatrat(x_i) - y_i \qquad \mbox{for $i = 1, \ldots,
    \numobs$.}
\end{align*}
A very crude form of consistency can be obtained by applying
Cauchy--Schwarz to the bound~\eqref{EqnEstBound}.  Doing so and
simplifying terms yields
\begin{align*}
\qempnorm{\fhatrat - \fdagger}^2 \leq \qempnorm{\MyGrad(\fhatrat) -
  \GradHat(\fhatrat)}^2.
\end{align*}
Thus, we see that the procedure will be consistent if we can ensure
that the teacher-based surrogate gradient $\GradHat(\fhatrat)$ is
uniformly consistent as an estimate of the true gradient
$\MyGrad(\fhatrat)$.  (To be clear, this Cauchy--Schwarz argument is
very loose in general, and our later analysis provides a far more
precise analysis.)

\paragraph{Bounds for approximate fixed points:}
Our theory can also be extended to give guarantees for a function that
satisfies the \RAT fixed point relation in an approximate sense.
Define the \RAT operator
\begin{subequations}
\begin{align}
\label{EqnRatOperator}
\TopHat_\stepsize(f) & \defn \ProxAlpha \big(f(\xtil_1^m) - \stepsize
\GradHat(f) \big)
  \end{align}
along with the \emph{proximal operator defect}
\begin{align}
\label{EqnRatDefect}    
\Defect(f) \defn \TopHat_\stepsize(f) - f,
\end{align}
\end{subequations}
which measures the deficiency in a function $f$ as a \RAT fixed point
(cf.  equation~\eqref{EqnFhatRatFix}).

For any function $f^k$, define the update $f^{k+1} \defn
\TopHat_\stepsize(f^k)$.  We then have the following extension
of~\Cref{ThmStat}, bounding the excess risk of $f^{k+1}$.
Under the assumptions of part (a), we have
\begin{subequations}
\begin{align}
\label{EqnRiskBoundFbar}
\Oracle(\funit{\kit+1}) - \Oracle(\fdagger)
& \leq \qempinner{\funit{\kit+1} - \fdagger}{
\MyGrad(\funit{\kit+1}) - \GradHat(\funit{\kit}) -
\frac{1}{\stepsize}\Defect(\funit{\kit})},
\end{align}
whereas under the assumptions of part (b), we have
\begin{align}
\label{EqnExcessRiskFbar}
\qempnorm{\funit{\kit+1} - \fdagger}^2 & \leq \frac{1}{\scon}
\qempinner{\funit{\kit+1} - \fdagger}{ \MyGrad(\funit{\kit+1}) -
  \GradHat(\funit{\kit}) - \frac{1}{\stepsize}\Defect(\funit{\kit})}.
\end{align}
\end{subequations}
We establish these more general claims as part of the proof
of~\Cref{ThmStat} in~\Cref{SecProofThmStat}.  When $f^k = \fhatrat$,
then we have $f^k = f^{k+1}$ and $\Defect(\funit{k}) = 0$, so that
with this particular choice, the bounds~\eqref{EqnRiskBoundFbar}
and~\eqref{EqnExcessRiskFbar} imply the bounds~\eqref{EqnRiskBound}
and~\eqref{EqnEstBound}, respectively.

\paragraph{Relaxed convexity requirements:}  Let us clarify how
the convexity assumptions can be weakened.  First, assuming that
$\LossBarM$ is convex in the fitted values $f(x)$ is relatively mild;
it holds for many standard losses (e.g., least-squares, logistic
regression, log loss etc.).  In assuming that $\Oracle$ is convex, we
are requiring that, in addition, the penalty function, and the set of
fitted values $f(x)$ obtained as $f$ varies over the student class
$\Fclass$, are convex.  This latter requirement is more stringent.
From the proof, however, the bound~\eqref{EqnRiskBound} only requires
convexity along the line joining $\fhatrat$ and $\fdagger$, meaning a
directional and localized form of convexity.  Meanwhile, the
bound~\eqref{EqnEstBound} exploits a non-expansivity property for the
proximal operator $\ProxAlpha$, holding locally between $\fhatrat$ and
$\fdagger$; see equation~\eqref{EqnFirmExpand} for the precise
definition.  Extensions to nonconvex settings may be possible under
prox-regularity assumptions, which ensure local single-valuedness and
Lipschitz continuity of the proximal
mapping~\cite{PoliquinRockafellar1996, RockafellarWets1998}.  Last, we
have many empirical results based on potentially non-convex teachers,
including various types of neural nets
(see~\Cref{FigRATvsPL}~and~\Cref{FigReLUBeta} for some
examples). These numerical studies suggest that the \RAT estimator can
be well-behaved in a broader setting than the theoretical set-up given
here.


\subsection{Some comparison with student soft-matching}
\label{SecPseudo}

As noted in the introduction, in the standard approach to
student--teacher estimation, the student is trained to directly
approximate the predicted outputs of the teachers.  Here we describe
this direct matching approach more precisely, and then prove some
guarantees for it that reveal key differences compared to the \RAT
estimate.  We begin by describing the \SM approach for a least-squares
regression problem with real-valued responses $y \in \real$, where it
is simplest to describe.  We then give the extension to classification
problems involving logits.

\paragraph{Student soft-matching (\SM) for regression:}  
For a scalar regression problem, the \SM approach uses the teacher to
construct a pseudo-response $\yhat_j \in \real$ for each target
covariate $\xtil_j$.  These pseudo-responses are used to construct the
synthetic dataset $\{(\xtil_j, \yhat_j) \}_{j=1}^\numtarget$, and the
empirical objective
\begin{subequations}
\begin{align}
  \label{EqnDefnPLLoss}
  \LossPL(f) & \defn \sum_{j=1}^\numtarget \loss(f(\xtil_j), \yhat_j),
\end{align}
which is meant to act as a surrogate to the smoothed target risk
$\LossBarM$ from equation~\eqref{EqnSmoothedRisk}.  Given our goal of
estimating the oracle estimand $\fdagger$ from
equation~\eqref{EqnDefnFdagger}, a natural \SM estimator, and the one
that we analyze, is given by
\begin{align}
\label{EqnDefnPLEst}  
\fhatpl & \defn \arg \min_{f \in \Fclass} \Big \{ \LossPL(f) + \Pen(f)
\Big \}.
\end{align}
\end{subequations}

\paragraph{\SM for classification:} Now let us describe the extension of
this approach to a $K$-ary classification problem.  Suppose that we
use the standard ``one-hot'' encoding of the $K$-classes, meaning that
class labels are vectors $y \in \real^K$, taking values in the set
$\mathcal{L} = \{e^1, \ldots, e^K \}$ of standard basis
vectors.\footnote{In particular, the vector $e^\ell \in \real^K$ is
zero in all positions except the $\ell^{th}$, where it takes the value
one.}  In this case, for each target sample $j = 1, \ldots, m$, the
``hard'' matching approach uses the teacher to construct synthetic
one-hot label vectors $\yhat_j \in \mathcal{L}$, and then trains the
student to match these labels.  The ``soft'' matching approach, which
has proven to be better behaved in practice, is to use the teacher to
predict either the logits or the probabilities of the class labels.
In the latter case, the vector $\yhat_j$ belongs to the
$K$-dimensional probability simplex, and the student is trained to
minimize the Kullback--Leibler divergence to these predictions.  This
leads to an instance of the \SM empirical loss~\eqref{EqnDefnPLLoss},
where
\begin{align}
\loss(f(\xtil_j), \yhat_j) & \defn \sum_{a=1}^K \yhat_{ja} \log
\Big(\frac{\yhat_{ja}}{f_a(\xtil_j)} \Big)
\end{align}
defines the KL divergence between the teacher probabilities, and the
student predicted probability $f(\xtil_j) \in \real^K$.

\subsubsection{A general bound for the \PL estimate}

We now state a guarantee for the \SM estimate $\fhatpl$ defined in
equation~\eqref{EqnDefnPLEst}.  It involves the \SM gradient vector
$\nabla \LossPL(f) \in \real^m$, with entries given by
\begin{align}
  [\nabla \LossPL(f)]_j = \frac{\partial}{\partial z} \ell(z, \yhat_j)
  |_{z = f(\xtil_j)} \qquad \mbox{for $j = 1, \ldots, m$.}
\end{align}
  More precisely, we have
\mygraybox{
\begin{proposition}
\label{PropPL}
Under the conditions of~\Cref{ThmStat}, the \PL estimate $\fhatpl$
based on teacher predictions $\yhat$ satisfies the bound
\begin{subequations}
\begin{align}
\label{EqnPLRiskBound}  
\Oracle(\fhatpl) - \Oracle(\fdagger) & \leq \qempinner{\fhatpl -
  \fdagger}{\nabla \LossBar_m(\fhatpl) - \nabla \LossPL(\fhatpl)},
\quad \mbox{and} \\
\label{EqnPLEstBound}  
  \qempnorm{\fhatpl - \fdagger}^2 & \leq \frac{1}{\scon}
  \qempinner{\fhatpl - \fdagger}{\nabla \LossBar_\numtarget(\fhatpl) -
    \nabla \LossPL(\fhatpl)}.
\end{align}
\end{subequations}
\end{proposition}
}
\noindent This claim follows by arguments similar to those used to
prove~\Cref{ThmStat}; see~\Cref{SecProofPropPL} for details. \\

Observe that the \SM bound~\eqref{EqnPLEstBound} and \RAT
bound~\eqref{EqnEstBound} have a parallel structure, but using
\emph{two different estimates} of the true gradient $\nabla
\LossBarM(f)$---namely, $\nabla \LossPL(f)$ and $\GradHat(f)$,
respectively.  To gain intuition, it is useful to compute these two
quantities for the least-squares loss given by \mbox{ $\loss(f(x_i),
  y_i) = (1/2) \big(f(x_i) - y_i \big)^2$,} for which we have the
residual or functional gradient $f(x_i) - y_i$.

For both methods, we can think of the teacher as a function $\Teacher:
\real^n \rightarrow \real^m$ that maps an $n$-vector from the source
domain to an $m$-vector in the target domain.  Given a vector $y =
(y_1, \ldots, y_n) \in \real^\numobs$ of source responses, the \PL
method uses the teacher to compute a vector $\yhat = \Teacher(y)$ of
pseudo-responses.  Consequently, the bounds of~\Cref{PropPL} will
apply in terms of the \SM gradient $\nabla \LossPL(f) \in \real^m$
with elements
\begin{subequations}
  \begin{align}
\label{EqnPLGradApprox}    
[\nabla \LossPL(f)]_j = f(\xtil_j) - [\Teacher(y)]_j.
\end{align}

On the other hand, the \RAT estimate uses the teacher in a rather
different way.  With the least-squares loss, the residual associated
with the pair $(x_i, y_i)$ is given by $f(x_i) - y_i$.  Thus, the \RAT
procedure uses the teacher to compute the gradient estimate
$\GradHat(f) \in \real^m$, with entries given by
\begin{align}
\label{EqnRatGradApprox}    
  [\GradHat(f)]_j = \Big[\Teacher \Big( f(x_1^n) - y \Big) \Big]_j
  \qquad \mbox{where $f(x_1^n) = (f(x_1), \ldots, f(x_n))$.}
\end{align}
\end{subequations}
These two expressions make clear the differences: the \PL procedure
applies the teacher to the responses $y \in \real^n$, whereas the \RAT
procedure applies the teacher to the residual vector $f(x_1^n) - y$.

\subsubsection{Explicit bias-variance decomposition}

This difference in how the teacher is
applied---viz. equations~\eqref{EqnPLGradApprox}
versus~\eqref{EqnRatGradApprox}---has important consequences for the
biases of the resulting estimates.  Let us now state a result that
reveals these differences more clearly.  For a regression problem with
responses $Y$, we can write $\Exs[Y \mid X = x] = \fdagger(x) +
\gstar(x)$, where $\fdagger$ is the student oracle
estimand~\eqref{EqnFdaggerFix}, and the term $\gstar$, when non-zero,
captures mis-specification in the student model.  With this
decomposition, we can write each source observation $y_i \in \real$ in
the form
\begin{align}
\label{EqnGenerativeModel}  
y_i & = \fdagger(x_i) + \gstar(x_i) + w_i, \qquad \mbox{for $i = 1,
  \ldots, \numobs$,}
\end{align}
where each $w_i$ is a conditionally zero-mean noise variable.

To decompose the teacher into bias and variance components, for any
function $h$, we define $\TeacherBar(h) \defn \Exs_w \Teacher(h + w)$,
where the expectation is taken over the random noise vector $w = (w_1,
\ldots, w_n)$.  We then define two bias terms
\begin{subequations}
\begin{align}
  \NewBiasPL & \defn \qempnorm{\TeacherBar(\fdagger(x_1^n) +
    \gstar(x_1^n)) - (\fdagger + \gstar)}^2, \quad \mbox{and}
  \\ \NewBiasRat & \defn \qempnorm{\TeacherBar(\Delta(x_1^n) -
    \gstar(x_1^n)) - (\Delta - \gstar)}^2 \qquad \mbox{where $\Delta
    \defn \fhatrat - \fdagger$.}
\end{align}
\end{subequations}
In addition, we define the zero-mean random noise vectors
\begin{subequations}
\begin{align}
  \newnoisepl & \defn \Teacher \big(\fdagger(x_1^n) + \gstar(x_1^n) +
  w \big) - \TeacherBar(\fdagger + \gstar), \quad \mbox{and} \\
  \newnoiserat & \defn \Teacher(\Delta(x_1^n) - \gstar(x_1^n) - w) -
  \TeacherBar(\Delta - \gstar).
\end{align}
\end{subequations}
Given these definitions, we have the following guarantee:%
\mygraybox{
  \begin{corollary}
\label{CorPLversusRat}    
For least-squares loss, the \PL estimate satisfies the bound
  \begin{subequations}
    \begin{align}
    \label{EqnSimplePLBound}
 \qempnorm{\fhatpl - \fdagger}^2 & \leq 2 \qempinner{\fhatpl -
   \fdagger}{\newnoisepl} +  \NewBiasPL,
    \end{align}
whereas the \RAT estimate satisfies the bound
\begin{align}
  \label{EqnSimpleRatBound}
 \qempnorm{\fhatrat - \fdagger}^2 & \leq 2 \qempinner{\fhatrat -
   \fdagger}{\newnoiserat} +  \NewBiasRat,
\end{align}
\end{subequations}
\end{corollary}
}
\noindent See~\Cref{SecProofCorPLversusRat} for the proof.\\

\Cref{CorPLversusRat} highlights how the \PL estimator involves a bias
component that depends on the full function $\fstar = \fdagger +
\gstar$, whereas the bias in the \RAT bound involves only the
mis-specification component $\gstar$.  This difference has important
implications for the (in)consistency of the \PL procedure.  Notably,
in~\Cref{SecSeparation}, we exhibit broad classes of problems for
which the \RAT estimate is consistent, whereas the \PL estimate
remains inconsistent.

The various terms in~\Cref{CorPLversusRat} take a simpler form when
the teacher procedure is \emph{response-linear}, meaning that its
behavior can be represented by a matrix $\UseTeachMat \in \real^{m
  \times n}$ that depends only on the covariates $\{x_i\}_{i=1}^n$ and
$\{\xtil_j \}_{j=1}^\numtarget$.  Various non-parametric smoothers,
among them kernel ridge regression (KRR)~\cite{Gu02,Wahba,Wai19},
Nadaraya-Watson
smoothing~\cite{watson1964smooth,nadaraya1964estimating}, and local
polynomial regression methods~\cite{fan1996local}, have this
response-linear form.  As a concrete example, consider the teacher
that performs kernel ridge regression using a pre-specified kernel
function \mbox{$\KerFun: \Xspace \times \Xspace \rightarrow \real$,}
and a given regularization parameter $\lambda > 0$.  In this case, the
teacher's mapping from source to target space is described by $m
\times n$ dimensional matrix
\begin{align}
\label{EqnKRRTeacher}  
\UseTeachMat_\lambda & \defn \Kmat_{mn} \big(\Kmat_{nn} + n \lambda
\IdMat_n \big)^{-1}
\end{align}
where the $n \times n$ source kernel matrix $\Kmat_{nn}$ has
$(i,j)^{th}$ entry $\KerFun(x_i, x_j)$, whereas the $m \times n$
target-source kernel matrix $\Kmat_{mn}$ has $(i,j)^{th}$ entry
$\KerFun(\xtil_j, x_i)$.

For any response-linear teacher, with the KRR being one example, a
straightforward calculation shows that we have the equivalence
\begin{subequations}
\begin{align}
\newnoisepl \equiv \newnoiserat \; = \; \TeachMat(w) \in \real^m.
\end{align}
Consequently,  for an estimate $\fhat \in \{\fhatpl, \fhatrat \}$,
we have the upper bound
\begin{align}
\label{EqnSimpleUpper}      
\qempnorm{\fhat - \fdagger}^2 & \leq 2 \, \underbrace{\qempinner{\fhat
    - \fdagger}{\UseTeachMat(w)}}_{\mbox{Stochastic error}} + B^2
\end{align}
\end{subequations}
where the squared-bias term $B^2$, given by $\NewBiasPL$ and
$\NewBiasRat$, determines the main differences between the two
estimators.  In~\Cref{SecTeacherEffect}, we discuss the form that the
stochastic error term in equation~\eqref{EqnSimpleUpper} takes for
different response-linear teachers.


\subsection{Exact MSEs and \PL-\RAT separation}
\label{SecSeparation}

In our analysis thus far, we have studied the gap between an estimate
$\fhat$, either the \RAT estimate $\fhatrat$ or the \SM estimate
$\fhatpl$, and the oracle student estimand $\fdagger$.  In practice,
one actually has a family of such oracle estimands, say of the form
\begin{align}
\label{EqnTemp}  
(\fdagger)^\stureg & = \arg \min_{f \in \Fclass} \Big \{ \LossBarM(f)
+ \stureg \Pen(f) \Big \} \qquad \mbox{for some regularization
  parameter $\stureg > 0$.}
\end{align}
Note that $(\fdagger)^\stureg$ is, at least in general, different from
the ground truth function $\fstar$, due to the effect of the student
regularization term $\stureg \Pen(f)$, and the difference
$(\fdagger)^\stureg - \fstar$ is the approximation error. In practice,
given an estimator $\fhat$, one adjusts the choice of $\stureg$ so as
to trade off the approximation error with statistical error.

In this section, we study this issue when the student class consists
of a family of kernel ridge estimators (KRR); here the \emph{student
regularization} parameter $\stureg > 0$ corresponds to the weight on
the squared Hilbert norm that serves as the regularizer. Our main
result is to establish that a \emph{biased teacher} leads to a
fundamental separation between the performance of the \RAT and \PL
estimators.  More precisely, we exhibit a simple class of problems for
which the \RAT procedure, with a suitable choice of regularization
parameter $\stureg$, provides a consistent estimate of $\fstar$ as the
source sample size $\numobs$ grows.  This consistency is guaranteed
even when the teacher introduces a constant level of bias $\lambda >
0$, independent of the sample size $\numobs$.  On the other hand, the
\PL procedure, even when the student regularization parameter $\stureg
> 0$ is chosen arbitrarily, \emph{cannot correct} the teacher bias.
Underlying this result are exact expressions for the mean-squared
errors of the \RAT and \PL estimates using student KRR estimators
(see~\Cref{PropExactMSE} in~\Cref{SecExact}).  We then use these
expressions to establish our separation result (\Cref{ThmSeparation})
in~\Cref{SecSeparation}.

\paragraph{Set-up:}  We begin with the problem set-up studied in this
section.  We make $n$ i.i.d. source observations of the form
\begin{align}
y_i & = \fstar(x_i) + w_i, \quad \mbox{for $i = 1, \ldots, \numobs$,}
\end{align}
where the noise terms $w_i$ are zero-mean with $\var(w_i) = \sigma^2$.
Our goal is to estimate the unknown function $\fstar$ via the
teacher-student interaction, and we use the least-squares loss for
this regression problem.

We assume that the student class is defined by a kernel function
$\KerFunTil: \Xspace \times \Xspace \rightarrow \real$, along with the
associated reproducing kernel Hilbert space $\Hil$.  For a
regularization parameter $\stureg > 0$, we consider the
\emph{$\stureg$-regularized ensemble} of student proximal updates with
$\Pen(f) = \frac{\stureg \numtarget}{2} \|f\|_\Hil^2$.  Thus, we have
a family $\{ \fhatratstu \}_{\stureg > 0}$ of \RAT estimates, indexed
by the choice of student regularization parameter $\stureg > 0$.
Similarly, we have an ensemble $\{\fhatplstu \}_{\stureg > 0}$ of \PL
estimates.  Note that the oracle student function
$(\fdagger)^\stureg$, defined via equation~\eqref{EqnTemp} with
$\Pen(f) = (\stureg/2) \numtarget \|f\|_\Hil^2$, is not equal to
$\fstar$, due to bias introduced by the penalty.

Suppose that the original regression problem is well-specified,
meaning that we can write the unknown regression function $\fstar$ in
the form $\fstar(\cdot) = \sum_{j=1}^m \thetastar_j \KerFunTil(\cdot,
\xtil_j)$ for a weight vector $\thetastar \in \real^m$.  In this
setting, if we were to make full observations of labeled target pairs
$ \{ (\xtil_j, \ytil_j) \}_{j = 1}^m$, it is then possible to estimate
$\fstar$ consistently by making a suitable choice of the
regularization parameter.  Our goal is to explore whether or not, for
the \RAT and \PL estimators, such consistency is possible given only
labeled source observations.

\subsubsection{Exact expressions}
\label{SecExact}

We begin by deriving exact expressions for the \RAT and \PL fixed
estimates, along with their mean-squared errors (MSEs).  Our analysis
applies to an arbitrary response-linear teacher, as represented by a
matrix $\TeachMat \in \real^{m \times n}$.  By construction, each of
the \RAT and \PL solutions correspond to functions of the form
\begin{align}
\label{EqnStudentRep}
f_\theta(\cdot) \defn \sum_{j=1}^{\numtarget} \theta_j
\KerFunTil(\cdot,\xtil_j) \qquad \mbox{for some weight vector $\theta
  \in \real^m$.}
\end{align}
We use $\thetahatrat$ and $\thetahatpl$ to denote the corresponding
weight vectors, so that $\fhatrat \equiv f_{\thetahatrat}$ and
$\fhatpl \equiv f_{\thetahatpl}$.

\paragraph{\RAT and \PL Fixed Points:}

In this case, for any teacher matrix $\UseTeachMat \in \real^{m \times
  n}$, it is possible to give explicit expressions for the \RAT and
\PL weight vectors $\thetahatrat$ and $\thetahatpl$.  More
specifically, define the target kernel matrix $\KmatTil_{\numtarget
  \numtarget} \in \real^{m \times m}$ with $(j, j')$-th entry given by
$\KerFunTil(\xtil_j, \xtil_{j'})$.  We also define the source-target
matrix $\KmatTil_{\numsource \numtarget} \in \real^{\numsource \times
  \numtarget}$ with $(i,j)$-th entry $\KerFunTil(x_i, \xtil_j)$.  In
terms of these matrices, it can be shown that the unique \PL fixed
estimate is given by
\begin{subequations}
  \begin{align}
  \label{EqnThetaHatPL}
  \thetahatpl = (\KmatTil_{\numtarget\numtarget} + \numtarget \stureg
  \IdMat_{\numtarget})^{-1} \TeachMat (y).
\end{align}
This follows in a straightforward way, since the PL procedure simply
applies the student regression method---in this case, the
$\stureg$-regularized KRR procedure---to the teacher's predictions
$\TeachMat(y) \in \real^m$.

Starting from the definition~\eqref{EqnFhatRatFix} of the \RAT fixed
point $\fhatrat$, it can be shown that it takes the form $\fhatrat
\equiv f_{\thetahatrat}$, where
\begin{align}
 \label{EqnThetaHatRAT}
\thetahatrat & \defn \big( \Aop + \numtarget \stureg
\IdMat_{\numtarget} \big)^{-1} \TeachMat(y), \qquad \mbox{where $\Aop
  \defn \TeachMat \KmatTil_{\numsource \numtarget} \in \real^{m \times
    m}$.}
\end{align}
\end{subequations}
See~\Cref{SecProofPropExactMSE} for the derivation of these relations.
For any estimate $\fhat \equiv f_{\thetahat}$, its target-based MSE,
relative to the true function $\fstar$, can be expressed as
\begin{align}
\MSE(\fhat) \defn \Exs_w \qempnorm{\fhat - \fstar}^2 = \Exs_w
\Big[\frac{1}{m} \sum_{j = 1}^m \big(\fhat(\xtil_j) - \fstar(\xtil_j)
  \big)^2 \Big].
\end{align}
We use $\MSE(\fhatpl)$ and $\MSE(\fhatrat)$ to denote this
mean-squared error for the \PL and \RAT estimators, respectively.
\mygraybox{
\begin{proposition}[Exact MSEs for \PL/\RAT]
\label{PropExactMSE}    
Under the previous assumptions, we have
\begin{subequations}
  \begin{align}
\label{EqnMSEDecompRAT}    
\MSE(\fhatrat) & = \frac{1}{m}\| \numtarget\gamma \KmatTil_{mm} (\Aop
+ \numtarget\gamma\IdMat)^{-1} \thetastar \|_2^2 +
\frac{\sigma^2}{\numtarget} \frobnorm{ \KmatTil_{\numtarget\numtarget}
  (\Aop + \numtarget\gamma\IdMat)^{-1} \TeachMat}^2
  \end{align}
  where the matrix $\Aop$ was previously
  defined~\eqref{EqnThetaHatRAT}, as well as
  \begin{align}
\label{EqnMSEDecompPL}    
 \MSE(\fhatpl) & = \frac{1}{m}\| \underbrace{\KmatTil_{mm} ( \IdMat
   - \Bmat_\stureg^{-1} \Aop) \thetastar}_{\mbox{\PL bias vector}}
 \|_2^2 + \frac{\sigma^2}{\numtarget} \frobnorm{
   \KmatTil_{\numtarget\numtarget} \Bmat_\stureg^{-1} \TeachMat}^2
  \end{align}
where $\Bmat_\stureg \defn \KmatTil_{\numtarget\numtarget} +
\numtarget \stureg \IdMat$. The \PL bias vector further decomposes
into distribution-shift and regularization components:
\begin{align}
  \label{EqnPLBiasSplit}
  \KmatTil_{\numtarget\numtarget}(\IdMat - \Bmat_\stureg^{-1} \Aop)
  \thetastar = \underbrace{(\KmatTil_{\numtarget\numtarget} -
    \Aop)\thetastar}_{\text{shift bias}} +
  \underbrace{\numtarget\gamma \Bmat_\stureg^{-1} \Aop
    \thetastar}_{\text{ridge bias}}.
\end{align}
\end{subequations}
  \end{proposition}
}
\noindent We prove these results in~\Cref{SecProofPropExactMSE}.


\subsubsection{Teacher-induced separation between \RAT and \PL}

\Cref{PropExactMSE} suggests that there might be a fundamental gap
between the performance of the \RAT and \PL estimators.  This
intuition turns out to be correct, and in this section, we give a
result that formalizes this performance gap in a particular setting.

Recall that~\Cref{PropExactMSE} applies to a student model class based
on a kernel function $\KerFunTil$.  Given this set-up, one natural
model of a biased teacher is one that performs kernel ridge regression
using the kernel function $\KerFunTil$ and the source data, along with
a \emph{teacher regularization parameter} $\lambda > 0$.  This leads
to a family of $\lambda$-biased teacher matrices, given by
\begin{align}
\TeachMat_\lambda & = \Kmat_{m n} \big( \Kmat_{nn} + \lambda n
\IdMat_n)^{-1}
\end{align}
where $\Kmat_{mn}$ has $(j,i)^{th}$-entry $\KernelTil(\xtil_j, x_i)$,
and $\Kmat_{nn}$ has $(i, i')^{th}$-entry $\KernelTil(x_i, x_{i'})$.
The third relevant kernel matrix is the student kernel matrix
$\KmatTil_{mm}$ previously defined for the target covariates
$\{\xtil_j\}_{j=1}^m$, which has $(j,j')^{th}$-entry $\KerFun(\xtil_j,
\xtil_{j'})$. \\

In order to vary the richness of the student/teacher classes, we
consider kernel matrices whose eigenspectra decay in a polynomial
manner.  More precisely, the matrix $\Kmat$ exhibits
\mbox{$\alpha$-eigendecay} if its eigenvalues satisfy the
relation\footnote{More precisely, there are universal constants $0 <
c_0 \leq c_1 < \infty$ such that $c_0 j^{-2 \alpha} \leq
\sigma_j(\Kmat) \leq c_1 j^{-2 \alpha}$.}  $\sigma_j(\Kmat) \asymp
j^{-2 \alpha}$.  With this notion, we study the following set-up:
\begin{itemize}[leftmargin=*]
\item The bias of the teacher model is captured by the parameter
  $\lambda > 0$. The richness of the kernel class relative to the
  source covariates is defined by the eigenspectrum of the matrix
  $\Kmat_{nn}$, assumed to have $\alpha$-polynomial decay for
  some $\alpha > 1/2$.
\item Similarly, the student model has bias parameter $\stureg > 0$,
  and the spectrum of the target kernel matrix $\Kmat_{mm}$ has
  $\beta$-polynomial decay for some \mbox{$\beta \in \big(1/2, \; 1/2
    + \alpha \big)$.}
\end{itemize}

Given a fixed teacher bias parameter $\lambda$, our goal is to
understand the differences between \PL and \RAT when the student bias
parameter $\stureg$ is \emph{freely adjustable}.  In the following
statement, the quantities $c_0,c_1,c_2>0$ are all universal constants
independent of $(\numsource,R,\sigma^2)$.
\newcommand{\labcol}[1]{\makebox[7em][l]{\textbf{#1}}}
\mygraybox{
  \begin{theorem}[Separation result for \PL/\RAT, with a matching lower bound]
    \label{ThmSeparation}
Consider a $\lambda$-biased teacher for some $\lambda > 0$, and a
source sample size $\numsource \geq \lambda^{ -1 -\frac{1}{2\alpha}}$.
Then there is a Hilbert space $\Hil$ with kernel $\KerFunTil$ and
source-target covariates for which $(\Kmat_{nn}, \KmatTil_{mm})$
exhibit $(\alpha, \beta)$-eigendecay such that, for any radius $R \geq
1$, we have:
\begin{subequations}
\begin{align}
\label{EqnRATShiftRateCorrect}
\mbox{{\bf{\RAT upper:}}} & \quad \sup_{\|\fstar\|_\Hil \leq R }\Exs_w
\qempnorm{ \fhatratstu - \fstar}^2 \leq c_0 \;R^{2}
\Big(\frac{\sigma^2}{ \numsource R^2} \Big)^{\frac{2\beta}{2\alpha+1}}
\quad \mbox{with $\gamma = \tfrac{1}{\lambda} \Big(\frac{\sigma^2}{R^2
    n} \Big)^{\frac{2(\alpha+\beta)}{2\alpha+1}}$},\\
\label{EqnPLShiftRateIncorrect}
\mbox{{\bf{\PL lower:}}} & \quad \sup_{\|\fstar\|_\Hil \leq R}
\inf_{\stureg > 0}\Exs_w\qempnorm{ \fhatplstu - \fstar}^2  \geq c_1
R^2 > 0.
\end{align}
Moreover, for source-target pairs $m = n \geq \max \big\{ \lambda^{ -1
  -\frac{1}{2\alpha}}, (R^2/\sigma^2)^{\frac{1}{2 \alpha}} \big \}$,
\emph{any estimator} $\ftil$ based on labeled source sample
$\{(x_i,y_i)\}_{i=1}^{\numsource}$ and the unlabeled target covariates
$\{\xtil_j\}_{j=1}^{\numsource}$ has MSE lower bounded as
\begin{align}
\label{EqnMinimaxShiftLower}
\mbox{\bf{Minimax lower:}} \qquad \sup_{\|\fstar\|_\Hil \le R}\Exs_w \qempnorm{\fhat-\fstar}^2 \geq c_2
\min \Big \{ \, R^2 \Bigl(\frac{\sigma^2}{\numsource
  R^2}\Bigr)^{\frac{2\beta}{2\alpha+1}}, R^2 \Big \}.
\end{align}
\end{subequations}
\end{theorem}
}
\noindent See \Cref{SecProofThmSeparation} for the proof.

At a high level, the main take-away is that the \RAT procedure, with a
properly adjusted level of student regularization, achieves the
minimax-optimal rate for estimating the true regression function
$\fstar$.  In contrast, the \SM estimate suffers from a constant level
bias, preventing consistency. \\

\noindent Beyond this performance gap, the \RAT consistency result
exhibits three distinct regimes that are worthy of further comment:
\paragraph{No covariate shift:} First of all, the setting of ``no
covariate shift'' can be captured by setting $\alpha = \beta$.  In
this case, viewing $(\sigma, R)$ as constants, the \RAT estimate
achieves estimation error that scales as $(1/\numobs)^{\frac{2
    \alpha}{2 \alpha +1}}$.  Note that this is the standard minimax
rate for a kernel that exhibits $\alpha$-polynomial decay
(e.g.,~\cite{Wai19}).  As a special case, if we use spline kernels,
then the RKHS corresponds to a Sobolev space with smoothness $\alpha >
1/2$.

Setting $\alpha \neq \beta$ induces covariate shift, and
interestingly, its effect can either be malign or benign.  (It is far
more common to view covariate shift as harmful than helpful.)  For the
sake of these comparisons, suppose that we have the same number of
target samples as source samples (i.e., $m = n$).  In this case, if we
were given $m$-labeled target samples, then training the student
directly would yield a rate of $(1/n)^{\frac{2 \beta}{2 \beta + 1}}$.
Let us compare this guarantee to the minimax-optimal guarantee of
$(1/n)^{\frac{2 \beta}{2 \alpha + 1}}$ that can be achieved when only
the source data has labels.

\paragraph{Malign covariate shift:}  First, if $\alpha > \beta$, then we
have ${\frac{2 \beta}{2 \beta + 1}} > \frac{2 \beta}{2 \alpha + 1}$,
so that the idealized estimator based on labeled target samples
exhibits faster convergence.  This is an example of harmful covariate
shift, in that it would have been more helpful to obtain labeled
samples from the target covariate distribution (instead of the
source).

\paragraph{Benign covariate shift:}  The opposite conclusion holds if $\alpha < \beta$:
here it turns out to be beneficial to observed labeled source data
instead of target data.  Intuitively, when $\alpha < \beta$, the
source covariate distribution $\Prob_X$ places more mass on the kernel
spectrum with larger eigenvalues, and it is these eigen-functions that
are most relevant in estimation. (Regularization in kernel ridge
regression amounts to down-weighting the impact of directions
associated with smaller eigenvalues.)


\subsection{Computational guarantees}
\label{SecComp}

In the previous section, we analyzed a particular student--teacher
pair for which the \RAT estimate can be computed in closed form
(viz. equation~\eqref{EqnThetaHatRAT}). In general, computing the \RAT
estimate requires iterating between the student--teacher steps, and
the simplest such scheme is the Picard iteration described previously
in~\Cref{SecRat}.  Given a stepsize $\stepsize > 0$, it begins with an
initial student function $\funit{0} \in \Fclass$, and then generates
the sequence
\begin{align}
  \label{EqnPicard}
  \funit{\kit+1} = \ProxAlpha \Big(\funit{\kit}(\xtil_1^m) - \stepsize
  \GradHat(\funit{\kit}) \Big) \qquad \mbox{for $\kit = 0, 1, 2,
    \ldots$,}
\end{align}
using the student proximal update $\ProxAlpha: \real^m \rightarrow
\Fclass$ from equation~\eqref{EqnStudentProximal}, and the
gradient estimator $\GradHat$ from equation~\eqref{EqnFuncGradSurrogate}.

\Cref{FigConvergence} illustrates some representative convergence
behavior that we observe in practice for this algorithm.  Here we are
measuring performance in terms of the operator defect
\begin{align}
\label{EqnNewDefect}
  \Defect(f) \defn \ProxAlpha \big(f(\xtil_1^m) - \stepsize
  \GradHat(f) \big) - f,
\end{align}
which measures how close $f \in \Fclass$ is to being a \RAT fixed
point.  Panels (a) and (b) correspond to a stepsize $\stepsize = 0.10$
and $\stepsize = 0.20$, respectively, and we show results for training
a $10$-class logistic classifier as the student, using three different
types of neural network teachers.  The teachers are labeled with pairs
$(h_1, h_2)$, corresponding to the number of units in the two hidden
layers of the architecture.  For this particular problem, at one
extreme, the choice $(h_1, h_2) = (16, 16)$, shown in light purple,
leads to a teacher neural net with sufficient flexibility to yield a
relatively accurate gradient approximation, whereas the choice $(h_1,
h_2) =(4,4)$, shown in red, leads to a teacher that is relatively
impoverished. The theory to be developed will give sufficient
conditions under which: (a) we are guaranteed to obtain convergence of
this type; and (b) the role of teacher-provided gradient
approximation.

\newcommand{\cfig}{0.38}
\begin{figure}[h]
  \begin{center}
    \begin{tabular}{ccc}
      \widgraph{\cfig \textwidth}{\myfigdir/fig_shot_noise_sev4_stepsize_0.1_defect}
      & \hspace*{0.08in} &
      \widgraph{\cfig
        \textwidth}{\myfigdir/fig_shot_noise_sev4_stepsize_0.2_defect}
      \\
      (a) Stepsize $\stepsize = 0.10$ && (b) Stepsize $\stepsize =
      0.20$
    \end{tabular}
    \caption{Convergence behavior of the Picard
      iteration~\eqref{EqnPicard} when using a neural network teacher
      to construct the gradient estimate $\GradHat$, and for
      multinomial classifier over $K = 10$ classes
      (see~\Cref{SecNumImage} for more details.)  Each plot shows the
      operator defect norm $\|\Defect(\funit{\kit})\|$ versus the
      iteration number $k$.  The three different curves correspond to
      three-layer neural-net teachers with three different architectures:
      hidden units $(h_1, h_2)$ are marked in the labels.  Solid lines
      correspond to the mean defect over $T = 100$ random trials, with
      the shaded areas showing 95\% confidence intervals.  Stepsizes
      $\stepsize$ are marked below each plot.}
    \label{FigConvergence}    
  \end{center}
\end{figure}

Let us now turn to some theoretical results.  As previously discussed,
the update~\eqref{EqnPicard} can be understood as mimicking a proximal
gradient update, using the exact gradient $\nabla \LossBarM(f)$ for
the oracle objective function~\eqref{EqnDefnFdagger}.  From known
results on proximal methods, the convergence properties of this method
depend on the co-coercivity and monotonicity properties of the
gradient
operator~\cite{bauschke2017convex,beck2017first,nesterov2018lectures}.
Accordingly, it is natural that the convergence properties of the
update~\eqref{EqnPicard} should depend on analogous properties for
the approximate gradient estimator $\GradHat$ provided by the teacher.
We introduce these properties next.

\paragraph{Structural properties of gradient estimator:}
Given a parameter $\liphat > 0$, we say that $\GradHat$ is $(\liphat,
\epsstable)$-approximately co-coercive at $\ftil$ if
\begin{subequations}
\begin{align}
  \label{EqnApproxCocoercive}
  \qempinner{f - \ftil}{\GradHat(f) - \GradHat(\ftil)} & \geq
  \frac{1}{\liphat} \Big \{ \qempnorm{\GradHat(f) - \GradHat(\ftil)}^2
  - \epsstable^2 \Big \} \qquad \mbox{for all $f \in \Fclass$.}
\end{align}
For $\epsstable = 0$, this definition corresponds to the standard
definition of co-coercivity for an operator, so we have a relaxation
of the standard condition.  Similarly, for a parameter $\sconhat > 0$,
we say that $\GradHat$ is $(\sconhat, \epsstable)$-approximately
monotone at $\ftil$ if
\begin{align}
\label{EqnApproxMonotone}
    \qempinner{f - \ftil}{\GradHat(f) - \GradHat(\ftil)} & \geq \sconhat
    \Big \{ \qempnorm{f - \ftil}^2 - \epsstable^2 \Big \} \qquad
    \mbox{for all $f \in \Fclass$.}
\end{align}
\end{subequations}
In terms of these conditions, we have:
\mygraybox{
\begin{theorem}[Computational guarantees for \RAT]
\label{ThmComp}   
Suppose that the gradient estimator $\GradHat$ is approximately
$(\liphat, \epsstable)$-co-coercive~\eqref{EqnApproxCocoercive} at
$\fhat$. Then after $\Kmax$ iterations with stepsize $\stepsize =
1/\Liphat$, we have
\begin{subequations}
\begin{align}
\label{EqnWeakPicard}    
\min_{\kit = 0, \ldots, \Kmax} \qempnorm{\Defect(\funit{\kit})}^2 &
\leq \tfrac{2}{\Kmax + 1} \qempnorm{\funit{0} - \fhat}^2 + 4
\tfrac{\epsstable^2}{\liphat^2},
  \end{align}
  where $\fhat$ is any \RAT fixed point.  If in addition $\GradHat$ is
  $(\sconhat, \epsstable)$-approximately
  monotone~\eqref{EqnApproxMonotone} at $\fhat$, then after $\Kmax$
  iterations with stepsize $\stepsize = \sconhat/\liphat^2$, we have
  \begin{align}
     \label{EqnStrongPicard}
    \qempnorm{\Defect(\funit{\Kmax})}^2 & \leq 2 \Big( 1 -
    \tfrac{\sconhat^2}{\liphat^2} \Big)^\Kmax \qempnorm{\funit{0} -
      \fhat}^2 + 8 \epsstable^2.
  \end{align}
\end{subequations}
\end{theorem}
}
\noindent See~\Cref{SecProofThmComp} for the proof. \\

Observe that the guarantee~\eqref{EqnStrongPicard} guarantees
geometric convergence (up to the radius $8 \epsstable^2$).  However,
it requires both the co-coercivity and monotonicity conditions to
hold.  On the other hand, the guarantee~\eqref{EqnWeakPicard} requires
only co-coercivity, but yields a slower $1/K$ convergence rate.  Both
rates have analogues for standard gradient-based optimization: the
$1/K$ rate holds for a function with Lipschitz gradient, otherwise
known as a smooth function, whereas the geometric rate holds for a
strongly convex and smooth function.

Let us describe how it is possible to show that the
co-coercivity~\eqref{EqnApproxCocoercive} and
monotonicity~\eqref{EqnApproxMonotone} conditions hold for various
problems.  To give the intuition, recall that $\GradHat(f)$ acts as an
approximation to the gradient $\nabla \LossBarM(f)$ of the smoothed
target risk~\eqref{EqnSmoothedRisk}.  When $\LossBarM$ is a
$\scon$-strongly convex function, then its gradient $\nabla \LossBarM$
satisfies the strong monotonicity condition
\begin{align}
\qempinner{f - \ftil}{\nabla \LossBarM(f) - \nabla \LossBarM(\ftil)} &
\geq \scon \qempnorm{f - \ftil}^2.
\end{align}
As long as $\GradHat(f)$ approximates $\nabla \LossBarM(f)$, then we
can expect that $\GradHat$ satisfies the approximate strong
monotonicity condition~\eqref{EqnApproxMonotone} with a suitable
$(\sconhat, \epsstable)$.  Similar comments apply to when $\nabla
\LossBarM$ is a $\beta$-smooth function, and the co-coercivity
condition.  We leave in-depth study for future work.


\section{Numerical results}
\label{SecNumerical}

In this section, we describe the results of various numerical
studies that complement our theoretical analysis.

\subsection{Separation between \PL and \RAT procedures}
\label{SecNumSeparation}

In~\Cref{ThmSeparation}, we stated and proved a result that reveals a
fundamental separation between the \PL and \RAT estimators: while the
\RAT procedure is consistent when the student model is appropriately
tuned, the \PL estimate is inconsistent, even when the same student
tuning is available.  In this section, we describe a suite of
numerical simulations that further explore this separation result.
These experiments serve three purposes: (i) to verify the statistical
error bounds in a setting that closely matches the theorem; (ii) to
demonstrate that the qualitative behavior persists for more general
kernel models, even when the theorem assumptions do not strictly hold;
and (iii) to test whether the same phenomenon appears for other
student--teacher interactions, such as those based on neural networks.

\subsubsection{Verification of the theoretical rate}
We first construct a synthetic class of problems that approximately
matches the construction used in the proof of \Cref{ThmSeparation}.
Specifically, in a univariate setting, we consider the target
distribution $\Qprob_X \sim N(0,1)$, and the family of source
distributions $\Prob_X \sim N(0, \sigma^2_P)$ for an adjustable
variance $\sigma^2_P > 0$.  We construct a kernel using a
finite-dimensional Hermite feature map; by computing the empirical
kernel matrices, we can study their eigendecay, and observe
approximate polynomial eigendecay with exponents $(\alpha,\beta)$.
\begin{figure}[h!]
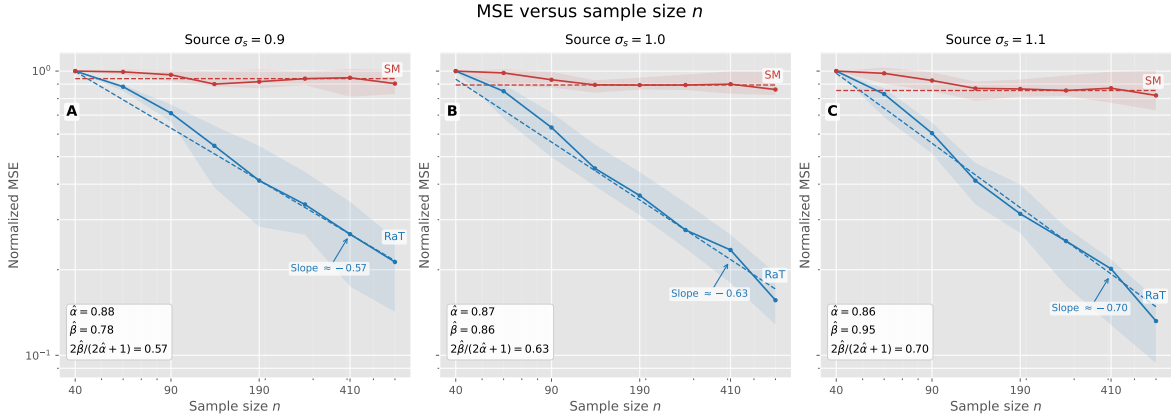

  \begin{center}
  \widgraph{0.95\textwidth}{\myfigdir/fignew_rat_vs_pl_hermite_gaussian}
  \caption{Mean-squared errors of \RAT and \PL under Gaussian
    covariate shift using a Hermite feature model, target distribution
    $\Target=N(0,1)$, and source distributions
    $\Source=N(0,\sigma_P^2)$ with $\sigma_P\in\{0.9,1.0,1.1\}$. Solid
    curves show the median MSE with interquartile bands over
    repetitions, and dashed lines indicate the predicted scaling
    laws. Consistent with \Cref{ThmSeparation}, \RAT achieves the rate
    $\numsource^{-2\beta/(2\alpha+1)}$, whereas \PL exhibits a
    non-vanishing error floor due to distribution shift.  }
\end{center}
 \label{FigHermiteGaussian}
\end{figure}
Figure~\ref{FigHermiteGaussian} shows that the \RAT estimator follows
the predicted rate $\numsource^{-2\beta/(2\alpha+1)}$, while the \PL
estimator exhibits a non-vanishing error due to the shift bias from
equation\eqref{EqnPLShiftRateIncorrect}.  The actual values of
$(\alpha, \beta)$, which determine the observed rate, depends on the
interaction between the source and target eigenspectra. In the setting
of \emph{benign covariate shift} with $\alpha < \beta$, as shown in
the right-most panel of~\Cref{FigHermiteGaussian}, we see that the
\RAT estimator can be faster than naive MSE rates determined solely by
$\alpha$ or $\beta$, i.e., $\numsource^{-2\beta/(2\alpha+1)} < \min
\{\numsource^{-2\alpha/(2\alpha+1)},
\numsource^{-2\beta/(2\beta+1)}\}$.


\subsubsection{Separation for more general kernels}
Next we consider a more generic kernel regression setting with target
distribution $\Target=\mathrm{Unif}[0,1]$, and source distribution
$\Source=\mathrm{Beta}(a,1)$.  We use kernel ridge regression with the
Laplace kernels, i.e., $\Kerfun(x,y) = \exp(-\|x-y\|_2/\nu_s)$ and
$\KernelTil(x,y) = \exp(-\|x-y\|_2/\nu_t)$ with $0 < \nu_t < \nu_s$.
Although the exact eigendecay assumptions of~\Cref{ThmSeparation} no
longer hold, the qualitative separation between \PL and \RAT is still
observed.
\begin{figure}[h!]
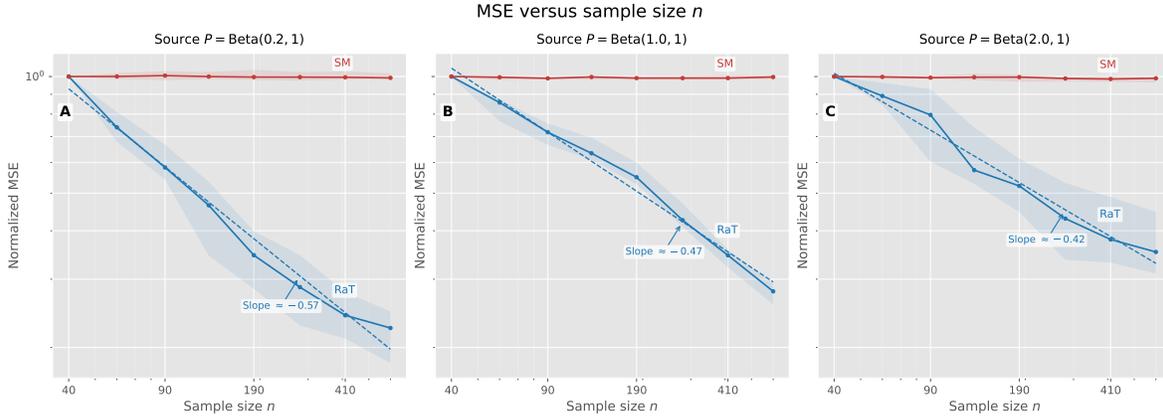

  \begin{center}
  \widgraph{0.95\textwidth}{\myfigdir/fignew_rat_vs_pl_laplace_beta}
  \caption{Mean-squared errors of the \RAT and \PL estimators for
    Laplace kernel ridge regression, using target distribution
    $\Target=\mathrm{Beta}(1,1)$ and source distributions
    $\Source=\mathrm{Beta}(\alpha,1)$ with $\alpha\in{0.2, 1.0,
      2.0}$. The teacher and student use Laplace kernels with
    differing bandwidths, so that the teacher is mis-specified
    relative to the student.  Solid curves show the median MSE with
    interquartile bands over repeated trials. Each curve is normalized
    by its value at the smallest sample size. The dashed line
    represents a least-squares linear fit on a log--log scale to the
    \RAT curve.}
  \end{center}
  \label{FigLaplaceBeta}
\end{figure}
\Cref{FigLaplaceBeta} shows that across all configurations, the \PL
estimator exhibits a persistent error floor, whereas the \RAT
estimator continues to improve as the source sample size increases.
This behavior is consistent with the theoretical analysis: \PL suffers
from a shift-induced bias, while \RAT progressively corrects this bias
through iterative feedback from the teacher model.
\subsubsection{Neural-network estimators}

Finally, we set up an experiment to examine whether similar phenomena
appear when using non-kernel pairs of student-teachers.  Concretely,
consider a regression task in which both the teacher and student are
one-hidden-layer ReLU networks with $k$ hidden units, corresponding to
functions of the form
\begin{align}
\label{EqnReluNet}  
h_{a, b, w}(x) & = \sum_{j=1}^k w_j \varphi \big( \inprod{a_j}{x} +
b_j \big) \qquad \mbox{for $(a_j, b_j) \in \real^d \times \real$, and
  $w \in \real^k$,}
\end{align}
and $\varphi(t) \defn \max(0, t)$ is the ReLU function.  We studied a
teacher class $\Gteacher$ based on $\kteacher = 2$ hidden units; this
very small choice induces a strong bias, analogous to the role played
by the explicit penalty $\lambda$ in a kernel procedure.  We construct
the student class $\Fstudent$ with the number of hidden units $\kstudent$
as a tuning parameter to be adjusted; it plays the role of $\stureg$
from our earlier experiments.

\begin{figure}[h!]
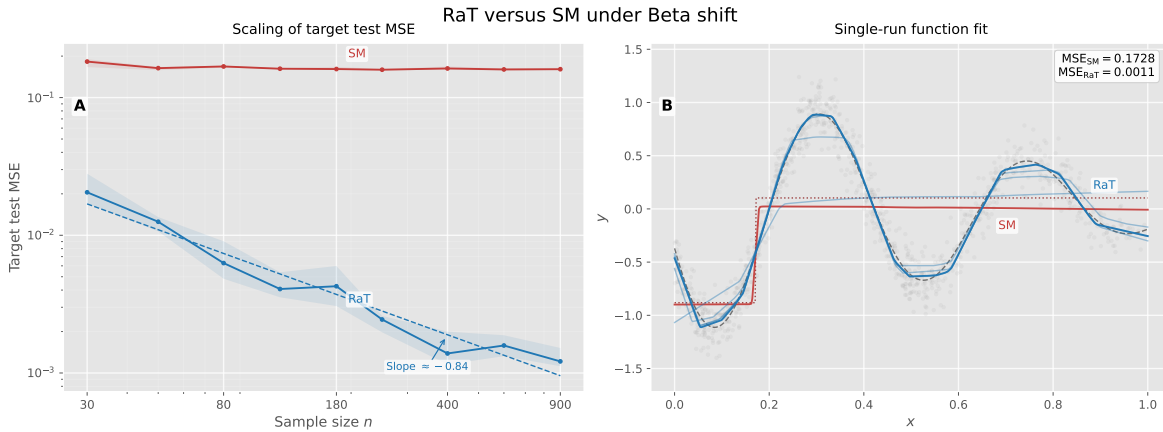

\centering
\widgraph{0.95\textwidth}{\myfigdir/fignew_rat_vs_pl_nn_beta}
\caption{Mean-squared errors (MSE) of \RAT and \PL estimators for
  neural network students with a regression-stump teacher, and Beta
  covariate shift.  (a) Target test MSE versus source sample size on a
  log-log scale. Solid lines show the median over repeated runs, and
  shaded regions indicate the interquartile range. The dashed line is
  a least-squares linear fit in log–log coordinates to the \RAT median
  curve.  (b) Function estimates for a single run, with ground truth
  function (black dashed), source samples (gray points), \PL teacher
  (dark red dotted), and final function estimates (solid blue for
  \RAT, solid red for \PL).  Faint blue curves correspond to
  intermediate \RAT iterates. The reported MSE values are computed on
  held-out target data. }
\label{FigReLUBeta}
\end{figure}
As seen in~\Cref{FigReLUBeta}, we observe the same qualitative
behavior in this neural net setting: the \PL estimator exhibits a
substantially higher error floor, while the \RAT estimator continues
to improve as the sample size increases.  These results suggest that
the separation mechanism identified in~\Cref{PropExactMSE} is not
limited to kernel methods but reflects a more general phenomenon of
iterative bias correction.


\subsection{Covariate shift in ImageNette}
\label{SecNumImage}

We now turn to experiments in a real data setting that further
elucidate the differences between the \SM and \RAT approaches.  These
experiments made use of the ImageNette
dataset~\cite{Howard_Imagenette_2019}, a subset of the full Imagenet
dataset restricted to $K = 10$ classes that are relatively easy to
distinguish.  After pre-processing, each covariate $x$ corresponds to
a tensor with dimensions $224 \times 224 \times 3$, representing an
image with $224 \times 224$ pixels, and $3$ color bands.  The
responses $y \in \real^{10}$ are all standard-basis vectors, with $y =
e_j$ meaning that the class label is equal to $j$.  There are $N =
13394$ samples in total, and in the experiments reported here, we used
random splits into 70\% training and 30\% validation.

Beginning with a ResNet18 neural network~\cite{he2016deep} trained on
the full ImageNet dataset, we performed $4$ epochs of fine-tuning
updates on the ImageNette subset, adjusting the weights and biases of
the last two layers.  Doing so yields a network that achieves
classification accuracy $\approx 98\%$ and cross-entropy loss $\approx
0.03$ on the ImageNette dataset, reflecting the easiness of the
original problem.  The final layer of the ResNet18 model contains a
total of $D = 512$ features; by performing PCA using the source
training data, we reduced these features to a total of $d = 40$
principal components, and both student and teacher models operate in
this $40$-dimensional space.  We remark that this PCA step has a very
minor effect on accuracy, for instance increasing cross-entropy loss
from $\approx 0.03$ to $\approx 0.07$.
\begin{figure}[h]
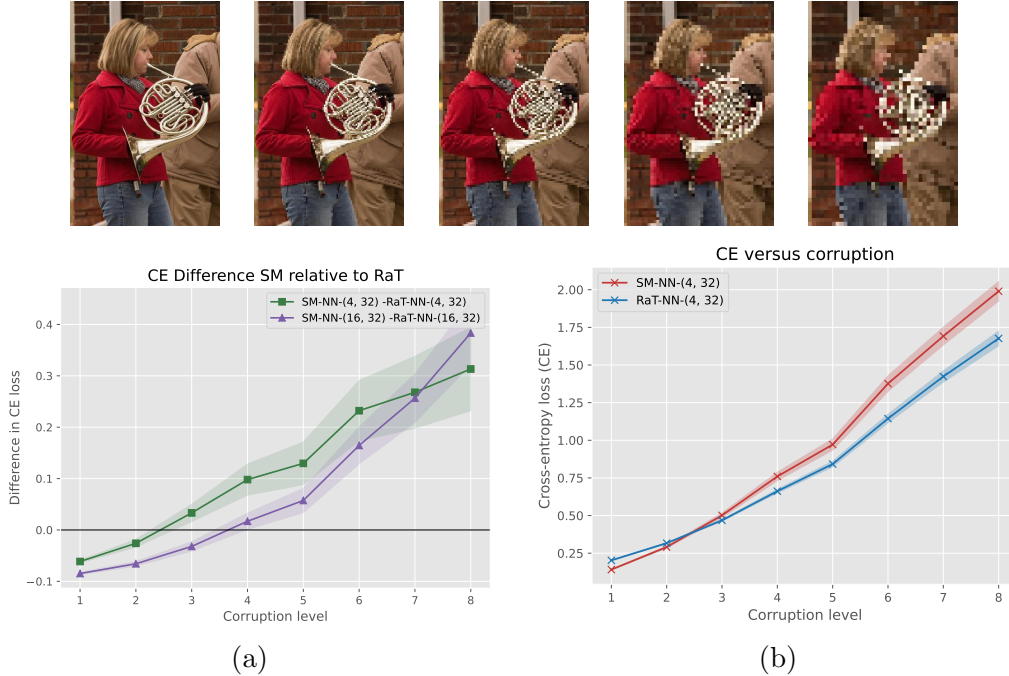

  \centering
  \begin{tabular}{ccccc}
    \widgraph{\imsize}{\myfigdir/fig_img_class_6_n03394916_51752_orig} & \hs
    \widgraph{\imsize}{\myfigdir/fig_img_class_6_n03394916_51752_pixelate_sev_2} & \hs
    \widgraph{\imsize}{\myfigdir/fig_img_class_6_n03394916_51752_pixelate_sev_4} & \hs
    \widgraph{\imsize}{\myfigdir/fig_img_class_6_n03394916_51752_pixelate_sev_7} & \hs
    \widgraph{\imsize}{\myfigdir/fig_img_class_6_n03394916_51752_pixelate_sev_10} 
  \end{tabular}
  \begin{tabular}{cc}
    \widgraph{\imfigsize}{\myfigdir/fig_pixelate_splits100_fl4_sl32_relative_differences}
    &
    \widgraph{\imfigsize}{\myfigdir/fig_pixelate_splits100_fl4_sl32_ind0_raw_mse}
    \\ (a) & (b)
  \end{tabular}
    \caption{Top row: a sample image from the ``French horn'' class,
      and its corrupted versions using the ``pixelate'' corruption.
      (a) Plots of the difference in cross-entropy (CE) loss between
      the \SM and \RAT procedures versus the corruption level.  Two
      types of neural nets are shown: three-layer architectures with
      $(h_1, h_2) = (4, 32)$ hidden units, or $(h_1, h_2) = (16,32)$
      hidden units.  Solid lines correspond to the mean CE difference
      over $T = 100$ random splits into train and validation, and the
      shaded areas correspond to 95\% confidence intervals based on
      these trials.  (b) Similar plots for the raw CE scores of the
      two methods, shown here for the $(4, 32)$-teacher.}
  \label{FigPixelate}
\end{figure}

In order to introduce covariate shift, we then adapted the image
corruption procedures from the Imagenet-C
dataset~\cite{hendrycks2019benchmarking}; here the ``C'' corresponds
to the various types of corruption (15 in total) that can be applied.
The top panels of~\Cref{FigPixelate} and \Cref{FigElastic}
respectively, show five levels of the ``Pixelate'' and ``Elastic
Blur'' corruptions.  By doing so, we obtained for a given corruption
(e.g., ``Pixelate'') and level, a new set of images to act as the
target set.

We explored the behavior of various teachers for this problem,
also acting on the extracted ResNet18 features.  Here we report some
representative results for two different types of teachers,
specifically designed to induce different levels of teacher bias:
\begin{itemize}[leftmargin=*,itemsep=0pt, topsep=1pt]
\item a three-layer ReLU neural network with hidden-unit dimensions
  $(h_1, h_2) = (4, 32)$, and
\item a three-layer ReLU neural network with hidden-unit dimensions
  $(h_1, h_2) = (16, 32)$.
\end{itemize}
We trained each teacher model using the ADAM algorithm with its
default settings, and a weight decay parameter $0.01$.  In all cases,
the student was a $K = 10$ class multinomial classifier operating on
the extracted ResNet18 features, and we trained the student model
using ADAM with these same settings.  For the \SM model, the student
was trained to fit (via cross-entropy loss) the teacher's predicted
probabilities, whereas for the \RAT procedure, the student performed
the proximal updates, using the teacher's predicted logits.
\begin{figure}[h]
  \centering
  \begin{tabular}{ccccc}
    \widgraph{\imsize}{\myfigdir/fig_img_class_8_n03425413_18562_orig} & \hs
    \widgraph{\imsize}{\myfigdir/fig_img_class_8_n03425413_18562_elastic_transform_sev_2} & \hs
    \widgraph{\imsize}{\myfigdir/fig_img_class_8_n03425413_18562_elastic_transform_sev_4} & \hs
    \widgraph{\imsize}{\myfigdir/fig_img_class_8_n03425413_18562_elastic_transform_sev_7} & \hs
    \widgraph{\imsize}{\myfigdir/fig_img_class_8_n03425413_18562_elastic_transform_sev_10} 
  \end{tabular}
  \begin{tabular}{cc}
    \widgraph{\imfigsize}{\myfigdir/fig_elastic_transform_splits100_fl4_sl32_relative_differences} &    
    \widgraph{\imfigsize}{\myfigdir/fig_elastic_transform_splits100_fl4_sl32_ind0_raw_mse}  \\
    (a) & (b) 
  \end{tabular}
  \caption{Top row: a sample image from the ``Gas pump'' class, and
    its corrupted versions using the ``elastic-blur'' corruption.  (a)
    Plots of the difference in cross-entropy (CE) loss between the \SM
    and \RAT procedures versus the corruption level.  Two types of
    neural nets are shown: three-layer architectures with $(h_1, h_2)
    = (4, 32)$ hidden units, or $(h_1, h_2) = (16,32)$ hidden units.
    Solid lines correspond to the mean CE difference over $T = 100$
    random splits into train and validation, and the shaded areas
    correspond to 95\% confidence intervals based on these trials.
    (b) Similar plots for the raw CE scores of the two methods, shown
    here for the $(4, 32)$-teacher.}
  \label{FigElastic}
\end{figure}
The number $h_1$ of hidden units in the first layer controls the
amount of teacher bias: strong ($h_1 = 4$) or medium ($h_1 = 16$).
With only $h_1 = 4$ hidden units in the first layer, the teacher is
forced to perform a drastic dimensionality reduction of the reduced
last layer (from dimension $40$ down to $4$ non-linear features).
This structural constraint embeds bias into the teacher, and we expect
that it becomes more severe as the degree of covariate shift (as
measured by the level of corruption) increases.

\Cref{FigPixelate} and~\Cref{FigElastic} show some representative
results, for the ``pixelate'' and ``elastic-blur'' corruptions
respectively.  In each plot, panel (a) plots the difference in
cross-entropy loss, measured on the validation set of the target
dataset, between the \SM and \RAT procedures, versus the amount of
covariate shift (as measured by the level of corruption applied). The
horizontal black line corresponds to the zero-line, so that above this
line, the \RAT procedure exhibits superior performance.  There are two
curves, corresponding to the two types of teachers, and the shaded
area corresponds to a 95\% confidence interval, constructed using $T =
100$ trials run on random splits of the dataset into training and
validation.  Panel (b) plots the raw values of the cross-entropy (CE)
loss for both \SM and \RAT procedures, again versus the corruption level
and with 95\% CIs shown.

As can be seen from both plots, when the corruption level is
relatively low, then the \SM procedure yields slightly lower CE loss
than the \RAT procedure.  We note that for low levels of corruption,
the problem is relatively easy, since the uncorrupted classifier has
cross-entropy loss $\approx 0.07$.  As the amount of corruption
increases, then the \RAT procedure starts to outperform.
In~\Cref{FigPixelate}, we see out-performance for both the
$(4,32)$-teacher (heavily biased), and the $(16, 32)$-teacher, whereas
in~\Cref{FigElastic}, the out-performance is much stronger for the
more heavily biased teacher.  These observations are qualitatively
consistent with our general theory.


\section{Proofs}
\label{SecProofs}

In this section, we collect the proofs of our three theorems.  We
focus on the main steps, with some more technical details provided in
the appendices.  In addition, the proofs of~\Cref{PropPL},
~\Cref{CorPLversusRat} and~\Cref{PropExactMSE} all are deferred
to~\Cref{SecAdditionalProofs} in the supplement.


\subsection{Proof of~\Cref{ThmStat}}
\label{SecProofThmStat}

We now turn to the proof of~\Cref{ThmStat}.  It suffices to prove the
more general bounds~\eqref{EqnRiskBoundFbar},
and~\eqref{EqnExcessRiskFbar}, since as noted in the discussion
following~\Cref{ThmStat}, they imply the bounds claimed in the theorem
statement.

Recall the penalty function $\Pen: \Fclass \rightarrow \real$ that
defines the student oracle function $\fdagger$, as in
equation~\eqref{EqnFdaggerFix}.  It is convenient to have a version
that acts on vectors $u \in \real^m$, so that we define
\begin{align}
\label{EqnDefnPenSmall}  
 \PenSmall(u) \defn \min \limits_{\{ f \in \Fclass \mid
   f(\xtil_1^\numtarget) = u \}} \Pen(f),
\end{align}
and we introduce the shorthand $\PenSmall(g) \equiv
\PenSmall(g(\xtil_1^\numtarget))$.  With a slight abuse of notation,
we identify a function $f \in \Fclass$ with its fitted-value vector
$f(\xtil_1^m) \in \real^m$ when taking Euclidean inner products and
norms.

Equipped with this notation, we are now ready to prove each of the two
bounds~\eqref{EqnRiskBoundFbar}, and~\eqref{EqnExcessRiskFbar}, in
turn.  Recall that they apply to an arbitrary function $\funit{\kit}
\in \Fclass$, and the one-step update \mbox{$\funit{k+1} =
  \TopHat_\stepsize(\funit{k})$.}  By definition of the defect
operator, we then have $\Defect(\funit{\kit}) = \funit{\kit+1} -
\funit{\kit}$.

\paragraph{Proof of the bound~\eqref{EqnRiskBoundFbar}:}

By assumption, the function $\Oracle(f) =
\frac{1}{\numtarget}\big\{\LossBar_\numtarget(f) + \Pen(f)\big\}$ is
convex in the fitted values.  Therefore, by the subgradient inequality
applied at the point $\funit{\kit+1}$, we have
\begin{align}
\label{EqnApproxRiskStart}
\Oracle(\fdagger) - \Oracle(\funit{\kit+1}) & \geq
\frac{1}{\numtarget} \inprod{\fdagger -
  \funit{\kit+1}}{\MyGrad(\funit{\kit+1}) + \zbar},
\end{align}
valid for any choice of subgradient $\zbar \in \partial
\PenSmall(\funit{\kit+1})$.

We now construct a particular sub-gradient.  By definition of the
proximal update, the fitted vector $\funit{\kit+1}(\xtil_1^m)$
minimizes the objective
\begin{align*}
u \mapsto \frac{1}{2 \stepsize}\|u - \big(\funit{\kit}(\xtil_1^m) -
\stepsize \GradHat(\funit{\kit})\big)\|_2^2 + \PenSmall(u).
\end{align*}
Consequently, the zero-subgradient optimality condition ensures that
there exists some vector \mbox{$z^{k+1} \in \partial
  \PenSmall(\funit{\kit+1})$} such that
\begin{align*}
0 & = \frac{1}{\stepsize}\Big\{\funit{\kit+1} - \big(\funit{\kit} -
\stepsize \GradHat(\funit{\kit})\big)\Big\} + \zbar \; = \;
\frac{1}{\stepsize}\big(\funit{\kit+1} - \funit{\kit}\big) +
\GradHat(\funit{\kit}) + \zbar.
\end{align*}
Since $\Defect(\funit{\kit})=\funit{\kit+1}-\funit{\kit}$, we see that
$\zbar \defn - \GradHat(\funit{\kit}) - \frac{1}{\stepsize}
\Defect(\funit{\kit})$ belongs to the sub-differential $\partial
\PenSmall(\funit{\kit+1})$.  Substituting this choice into our lower
bound~\eqref{EqnApproxRiskStart}, we find that
\begin{align*}
\Oracle(\fdagger) - \Oracle(\funit{\kit+1}) & \geq
\frac{1}{\numtarget} \inprod{\fdagger - \funit{\kit+1}}{
  \MyGrad(\funit{\kit+1}) - \GradHat(\funit{\kit}) -
  \frac{1}{\stepsize}\Defect(\funit{\kit})}.
\end{align*}
Re-arranging both sides yields
\begin{align*}
\Oracle(\funit{\kit+1}) - \Oracle(\fdagger)
& \leq
\qempinner{\funit{\kit+1} - \fdagger}{
\MyGrad(\funit{\kit+1}) - \GradHat(\funit{\kit}) -
\frac{1}{\stepsize}\Defect(\funit{\kit})},
\end{align*}
which proves the claimed bound~\eqref{EqnRiskBoundFbar}.

\paragraph{ Proof of the estimation bound~\eqref{EqnExcessRiskFbar}:}

Our proof is based on the fact that the proximal operator $\ProxAlpha:
\real^\numtarget \rightarrow \real^\numtarget$ is firmly
non-expansive~\cite{parikh2014proximal,bauschke2017convex}, so that
for any vectors $u, v \in \real^\numtarget$, we have
\begin{align}
\label{EqnFirmExpand}  
\empnorm{\ProxAlpha(u) - \ProxAlpha(v)}_2^2 & \leq
\empinner{u-v}{\ProxAlpha(u) - \ProxAlpha(v)}.
\end{align}
Since $\fdagger$ is the oracle minimizer, it satisfies the exact fixed
point relation
\begin{align*}
  \fdagger = \ProxAlpha\big(\fdagger(\xtil_1^m) - \stepsize
  \MyGrad(\fdagger)\big),
\end{align*}
and by definition, we have
\begin{align*}
\funit{\kit+1} & = \ProxAlpha\big(\funit{\kit}(\xtil_1^m) - \stepsize
\GradHat(\funit{\kit})\big).
\end{align*}
We now apply firm non-expansivity~\eqref{EqnFirmExpand} with the
choices
\begin{align*}
u & = \funit{\kit}(\xtil_1^m) - \stepsize \GradHat(\funit{\kit}),
\qquad\mbox{and}\qquad v = \fdagger(\xtil_1^m) - \stepsize
\MyGrad(\fdagger).
\end{align*}
Doing so yields the inequality
\begin{align*}
\empnorm{\funit{\kit+1} - \fdagger}_2^2 & \leq \empinner{
  \big(\funit{\kit} - \stepsize \GradHat(\funit{\kit})\big) -
  \big(\fdagger - \stepsize \MyGrad(\fdagger)\big) }{ \funit{\kit+1} -
  \fdagger }.
\end{align*}
Using the identity
$\funit{\kit} = \funit{\kit+1} - \Defect(\funit{\kit})$, we can rewrite
the right-hand side as
\begin{align*}
\empinner{
\funit{\kit+1} - \fdagger
}{
\funit{\kit+1} - \fdagger
- \Defect(\funit{\kit})
- \stepsize\big(\GradHat(\funit{\kit}) - \MyGrad(\fdagger)\big)
}.
\end{align*}
Cancelling the common term
$\empnorm{\funit{\kit+1} - \fdagger}_2^2$ from both sides and dividing
by $\stepsize > 0$ yields
\begin{align*}
0 & \leq \inprod{\funit{\kit+1} - \fdagger}{
  \frac{1}{\stepsize}\Defect(\funit{\kit}) + \GradHat(\funit{\kit}) -
  \MyGrad(\fdagger)},
\end{align*}
or equivalently,
\begin{align*}
\inprod{\funit{\kit+1} - \fdagger}{\MyGrad(\fdagger)} & \leq
\inprod{\funit{\kit+1} - \fdagger}{ \GradHat(\funit{\kit}) +
  \frac{1}{\stepsize}\Defect(\funit{\kit})}.
\end{align*}
Adding and subtracting $\MyGrad(\funit{\kit+1})$ and re-arranging then
yields
\begin{align}
\label{EqnApproxIntermediate}
\inprod{\funit{\kit+1} - \fdagger}{
\MyGrad(\funit{\kit+1}) - \MyGrad(\fdagger)}
& \leq
\inprod{\funit{\kit+1} - \fdagger}{
\MyGrad(\funit{\kit+1}) - \GradHat(\funit{\kit}) -
\frac{1}{\stepsize}\Defect(\funit{\kit})}.
\end{align}
Since $\LossBar_\numtarget$ is $\scon$-strongly convex, its gradient
$\MyGrad = \nabla \LossBar_\numtarget$ is $\scon$-strongly monotone, so
that the left-hand side is lower bounded as
\begin{align*}
\inprod{\funit{\kit+1} - \fdagger}{
\MyGrad(\funit{\kit+1}) - \MyGrad(\fdagger)}
& \geq \scon \empnorm{\funit{\kit+1} - \fdagger}_2^2.
\end{align*}
Combining this inequality with~\eqref{EqnApproxIntermediate}, and then
rescaling both sides by $1/\numtarget$ to convert to the empirical
norm $\qempnorm{\cdot}$ and inner product $\qempinner{\cdot}{\cdot}$,
we find that
\begin{align*}
\qempnorm{\funit{\kit+1} - \fdagger}^2
& \leq \frac{1}{\scon}
\qempinner{\funit{\kit+1} - \fdagger}{
\MyGrad(\funit{\kit+1}) - \GradHat(\funit{\kit}) -
\frac{1}{\stepsize}\Defect(\funit{\kit})},
\end{align*}
as claimed in equation~\eqref{EqnExcessRiskFbar}.



\subsection{Proof of~\Cref{ThmSeparation}}
\label{SecProofThmSeparation}

The theorem asserts the existence of kernel and source-target pairs
with certain properties.  We begin by describing the specific class of
kernels and source-target pairs that underlies our construction.

\paragraph{A simple feature-based RKHS.}

In order to make the spectral behavior of the matrices in
equations~\eqref{EqnThetaHatPL}--\eqref{EqnThetaHatRAT} explicit, we
consider feature-linear kernels of the form $\Kernel(x,y) =
\KernelTil(x,y)= \phi(x)^\top \phi(y)$, where $\phi: x \mapsto \phi(x) \in
\real^D$ is some feature map.  The associated reproducing kernel
Hilbert space (RKHS) consists of feature-linear functions of the form
$f_v(x) = v^T \phi(x)$ for some weight vector $v \in \real^D$, and we
have $\|f_v\|_{\Hil} = \|v\|_2$.  We assume that the true function
$\fstar \in \Hil$, say of the form $\fstar(x) = (\vstar)^T \phi(x)$ for some
$\vstar \in \real^D$, and that its Hilbert norm is bounded as
$\|\fstar\|_\Hil = \|v^*\|_2 \leq R$.

Let $\PhiMat \in \real^{\numsource \times D}$ be the source feature
matrix with $\phi(x_i) \in \real^D$ as its $i^{th}$ row, and define
the target feature matrix $\PhiMatTil \in \real^{m \times D}$ in an
analogous manner.  These feature matrices induce the $D$-dimensional
empirical second-moment matrices $\Sigmat \defn \frac{1}{\numsource}
\PhiMat^\top \PhiMat\in\real^{D \times D}$ and $\SigmatTil \defn
\frac{1}{\numtarget} \PhiMatTil^\top \PhiMatTil\in\real^{D \times
  D}$. In our analysis, we assume that $\Sigmat$ and $\SigmatTil$
commute and therefore admit a common orthonormal eigenbasis.  Let
$\{\eval_k\}_{k=1}^D$ and $\{\evaltil_k\}_{k=1}^D$ denote their
eigenvalues in this shared basis, and suppose that they satisfy the
polynomial eigendecay conditions $\eval_k\asymp k^{-2\alpha}$ and
$\evaltil_k \asymp k^{-2\beta}$. Given this set-up, there is a unitary
matrix $\Umat\in \real^{D\times D}$ such that
\begin{subequations}
\begin{align}
  \label{EqnEigendecay}
\Sigmat & = \Umat \diag(\eval_1,\dots,\eval_D) \Umat^\top, \quad
\SigmatTil = \Umat \diag(\evaltil_1,\dots,\evaltil_D) \Umat^\top,
\quad \text{with $\eval_k\asymp k^{-2\alpha}, \quad \evaltil_k \asymp
  k^{-2\beta}$}.
\end{align}
The kernel matrices can written in terms of the feature matrices as
\begin{align}
\Kmat_{\numsource\numsource} &= \PhiMat
\PhiMat^\top\in\real^{\numsource\times\numsource}, \quad
\KmatTil_{\numtarget\numsource} = \PhiMatTil
\PhiMat^\top\in\real^{\numtarget \times \numsource}, \quad
\KmatTil_{\numtarget\numtarget} = \PhiMatTil
\PhiMatTil^\top\in\real^{\numtarget \times \numtarget} .
\end{align}
\end{subequations}
Consequently, the eigendecay assumptions on $\Sigmat$ and $\SigmatTil$
imply that the matrices $(\Kmat_{\numsource\numsource},
\KmatTil_{\numtarget\numtarget})$ exhibit $(\alpha,\beta)$-eigendecay,
which ensures that the spectral conditions required in the theorem are
satisfied.


\subsubsection{Proof of the \RAT bound~\eqref{EqnRATShiftRateCorrect}}
Let us first restate the MSE decomposition~\eqref{EqnMSEDecompRAT}
using our new representation, and the shorthand $\Sigmat_\lambda =
\Sigmat + \lambda\IdMat_D$.  The \RAT estimate has mean-squared error
$\Exs \qempnorm{ \fhatrat^{\; \stureg} - \fstar}^2 = \FinRatBias^2 +
\FinRatVar$.  In the feature-based representation, these bias and
variance terms are given by
\begin{subequations}
\begin{align}
\label{EqnDefnFinRatBias}    
\FinRatBias^2(\fstar) & \defn
\|\stureg\SigmatTil^{1/2}(\SigmatTil\Sigmat\Sigmat_\lambda^{-1} +
\stureg\IdMat_D)^{-1} v^*\|^2, \quad \mbox{and} \\
\label{EqnDefnFinRatVar}    
\FinRatVar & \defn \frac{\sigma^2}{\numsource}\Tr \Big(
\SigmatTil(\SigmatTil \Sigmat \Sigmat_\lambda^{-1} + \stureg
\IdMat_D)^{-1} \SigmatTil \Sigmat\Sigmat_\lambda^{-2} \SigmatTil (
\SigmatTil \Sigmat \Sigmat_\lambda^{-1} + \stureg \IdMat_D)^{-1}
\Big),
\end{align}
\end{subequations}
where the derivation of these expressions can be found in Section~\ref{SecProofFinRatDecomp}.
Here we recall that the true function $\fstar$ can be written as
$\fstar(x) = (\vstar)^T \phi(x)$.  The bulk of our technical work is devoted
to establish upper bounds on these two terms, which we state here.  In
the following result, we use $(c_0, c_1)$ to denote universal
constants.
\mygraybox{
  \begin{lemma}[\RAT bias and variance bounds]
\label{LemAuxiliary}    
Under the conditions defined above, 
the \RAT bias has the uniform upper bound
\begin{subequations}
  \begin{align}
\label{EqnBiasBound}    
  \sup_{\|\fstar\|_\Hil \leq R} \FinRatBias^2(\fstar) & \leq c_0 R^2
  (\lambda\stureg)^{\frac{\beta}{\alpha+\beta}}.
  \end{align}
Moreover, its variance satisfies the upper bound
\begin{align}
\label{EqnVarBound}  
\FinRatVar & \leq \frac{c_1
  \sigma^2}{\numsource}(\lambda\stureg)^{-\frac{2 \alpha - 2 \beta +
    1}{2 (\alpha+\beta)}}.
\end{align}
\end{subequations}
\end{lemma}
}
\noindent See~\Cref{SecProofEqnBiasBound}
and~\Cref{SecProofEqnVarBound} for the proofs of these two bounds,
respectively.  Taking them as given, let us complete the proof of the
\RAT upper bound in~\Cref{ThmSeparation}.

\paragraph{Balancing the terms.}

Combining the bias and variance bounds proved above yields
\begin{align}
\label{EqnMasterBound}
\MSE(\fhatratstu) \; \defn \; \Exs \qempnorm{ \fhatrat^{\; \stureg} -
  \fstar}^2 & \leq c_0 R^2
(\lambda\stureg)^{\frac{\beta}{\alpha+\beta}} + c_1
\frac{\sigma^2}{\numsource}(\lambda\stureg)^{-\frac{2\alpha-2\beta+1}{2(\alpha+\beta)}}.
\end{align}
The first term is increasing in $\stureg$, whereas the second is
decreasing in $\stureg$, so an optimal choice of $\stureg$ is obtained
by balancing these two contributions.  An order-optimal choice by
choosing $\stureg$ so that
\begin{align*}
R^2(\lambda\stureg)^{\frac{\beta}{\alpha+\beta}} =
\frac{\sigma^2}{\numsource}(\lambda\stureg)^{-\frac{2\alpha-2\beta+1}{2(\alpha+\beta)}}.
\end{align*}
Following some algebra, we arrive at the choice $\stureg = c'
\frac{1}{\lambda} \; \big(\frac{\sigma^2}{R^2 \numsource}
\big)^{\frac{2(\alpha+\beta)}{2\alpha+1}}$ for some universal constant
$c$.  Substituting this choice back into the
bound~\eqref{EqnMasterBound} yields
\begin{align*}
\MSE(\fhatratstu) & \leq c R^{2 -\frac{4\beta}{2\alpha+1}}
\sigma^{\frac{4\beta}{2\alpha+1}}
\numsource^{-\frac{2\beta}{2\alpha+1}} \; = \; c \, R^2 \Big(
\frac{\sigma^2}{\numobs R^2} \Big)^\frac{2 \beta}{2 \alpha +1 },
\end{align*}
as claimed.  Finally, the standing assumption
$\stureg\le\lambda^{\beta/\alpha}$ is satisfied whenever
$\big(\frac{\sigma^2}{R^2\numsource}
\big)^{\frac{2(\alpha+\beta)}{2\alpha+1}} \le
\lambda^{1+\beta/\alpha}$, or equivalently, whenever $\numsource \ge
c_2 \lambda^{-(2\alpha+1)/(2\alpha)}$ for a sufficiently large
constant $c_2$. \\

\noindent It remains to prove the two claims given
in~\Cref{LemAuxiliary}.


\subsubsection{Proof of the bound~\eqref{EqnBiasBound}}
\label{SecProofEqnBiasBound}

By the definition~\eqref{EqnDefnFinRatBias} of the \RAT bias term, we have
\begin{align*}
\sup_{\|\fstar\|_{\Hil}\le R} \FinRatBias^2(\fstar) &
\stackrel{(i)}{=} \sup_{\|\vstar\|_2 \le R}
\|\stureg\SigmatTil^{1/2}(\SigmatTil\Sigmat\Sigmat_\lambda^{-1} +
\stureg\IdMat_D)^{-1} \vstar\|_2^2 \\
& \stackrel{(ii)}{\leq} R^2 \;
\opnorm{\stureg\SigmatTil^{1/2}(\SigmatTil\Sigmat\Sigmat_\lambda^{-1}
  + \stureg\IdMat_D)^{-1}}^2
\end{align*}
where step (i) follows since $\fstar(x) = (\vstar)^T \phi(x)$ for some
vector $\vstar \in \real^D$, along with the equivalence
$\|\fstar\|_\Hil = \|\vstar\|_2$; and step (ii) follows from the
variational definition of the operator norm $\opnorm{\cdot}$ (i.e.,
the maximum singular value).

Since the matrices $\Sigmat$ and $\SigmatTil$ commute, they admit a
common orthonormal eigenbasis given by the columns of $\Umat$ in the
representation~\eqref{EqnEigendecay}.  In this basis, we have
$\SigmatTil \Sigmat \Sigmat_\lambda^{-1} = \Umat \diag(a_1,\dots,a_D)
\Umat^\top$, where $a_k \defn
\frac{\evaltil_k\eval_k}{\eval_k+\lambda} \geq 0$ Putting together the
pieces, we find that
\begin{align*} 
\sup_{\|\fstar\|_{\Hil}\le R} \FinRatBias^2(\fstar) \leq R^2
\opnorm{\stureg \SigmatTil^{1/2}(\SigmatTil \Sigmat
  \Sigmat_\lambda^{-1} + \stureg \IdMat_D)^{-1}}^2 & \leq R^2 \max_{k
  = 1, \ldots, D} \frac{\stureg^2 \evaltil_k}{(a_k + \stureg)^2}.
\end{align*}
Moreover, since $a_k \geq 0$, we have the upper bound
\begin{align*}
\frac{\stureg^2 \evaltil_k}{(a_k+\stureg)^2} = \frac{\stureg
  \evaltil_k}{a_k+\stureg} \cdot \frac{\stureg}{a_k + \stureg} \leq
\frac{\stureg\evaltil_k}{a_k+\stureg},
\end{align*}
from which it follows that
\begin{align*}
\sup_{\|\fstar\|_{\Hil}\le R} \FinRatBias^2(\fstar) & \leq R^2
\max_{k=1, \ldots, D} \frac{\stureg\evaltil_k}{a_k+\stureg}.
\end{align*}
In order to complete the proof of the bound\eqref{EqnBiasBound}, it
suffices to show that, under the assumed eigendecay conditions on the
eigensequences $\eval_k$ and $\evaltil_k$, there a constant $c_0 \in
(0,\infty)$ independent of $\lambda$ and $\stureg$, such that
\begin{align}
\label{EqnTechnical}  
\max_{k = 1, \ldots, D} \frac{\stureg\evaltil_k}{a_k+\stureg} \leq c_0
(\lambda\stureg)^{\frac{\beta}{\alpha+\beta}}.
\end{align}
We prove this final technical claim in~\Cref{SecProofEqnTechnical}.


\subsubsection{Proof of the bound~\eqref{EqnVarBound}}
\label{SecProofEqnVarBound}

We now prove the variance bound~\eqref{EqnVarBound}. By
definition~\eqref{EqnDefnFinRatVar} of the variance term, we have
\begin{align*}
\FinRatVar \defn \frac{\sigma^2}{\numsource}\Tr\Big(
\SigmatTil(\SigmatTil\Sigmat\Sigmat_\lambda^{-1}+\stureg\IdMat_D)^{-1}
\SigmatTil\Sigmat\Sigmat_\lambda^{-2}
\SigmatTil(\SigmatTil\Sigmat\Sigmat_\lambda^{-1}+\stureg\IdMat_D)^{-1}
\Big).
\end{align*}
Since $\Sigmat$ and $\SigmatTil$ commute, they admit a common
orthonormal eigenbasis.  Hence there exists a unitary matrix $\Umat$
which diagonalizes both $\Sigmat$ and $\SigmatTil$ simultaneously;
cf. equation~\eqref{EqnEigendecay}.  It follows that
\begin{align*}
\Sigmat_\lambda^{-1} = (\Sigmat+\lambda\IdMat_D)^{-1} = \Umat
\diag((\eval_1+\lambda)^{-1},\dots,(\eval_D+\lambda)^{-1}) \Umat^\top,
\end{align*}
and therefore
\begin{align*}
\SigmatTil\Sigmat\Sigmat_\lambda^{-1} = \Umat
\diag\!\left(\frac{\evaltil_1\eval_1}{\eval_1+\lambda},\dots,\frac{\evaltil_D\eval_D}{\eval_D+\lambda}\right)
\Umat^\top.
\end{align*}
Recalling the notation $a_k \defn
\frac{\evaltil_k\eval_k}{\eval_k+\lambda}$, we obtain
\begin{align}
  (\SigmatTil\Sigmat\Sigmat_\lambda^{-1}+\stureg\IdMat_D)^{-1} = \Umat
  \diag((a_1+\stureg)^{-1},\dots,(a_D+\stureg)^{-1}) \Umat^\top.
\end{align}
Similarly, we have $\SigmatTil\Sigmat\Sigmat_\lambda^{-2} = \Umat
\diag \big( \frac{\evaltil_1\eval_1}{(\eval_1+\lambda)^2}, \ldots,
\frac{\evaltil_D\eval_D}{(\eval_D + \lambda)^2} \big) \Umat^T$.

Substituting these diagonal representations into the trace expression
for $\FinRatVar$ and using the invariance of the trace under
conjugation, we obtain
\begin{align*}
\FinRatVar  \; = \; \frac{\sigma^2}{\numsource} \Tr\Big(\Umat \diag\!\Big(
\frac{\evaltil_k^3\eval_k}{(\eval_k+\lambda)^2}\frac{1}{(a_k+\stureg)^2}
\Big) \Umat^\top \Big) 
& \; = \; \frac{\sigma^2}{\numsource}
\sum_{k=1}^D
\frac{\evaltil_k^3\eval_k}{(\eval_k+\lambda)^2}\frac{1}{(a_k+\stureg)^2} \\
&\; = \; \frac{\sigma^2}{\numsource}\sum_{k=1}^D
\frac{a_k^2}{(a_k+\stureg)^2}\frac{\evaltil_k}{\eval_k}, 
\end{align*}
where the last equation follows from the definition of $a_k$.

In order to analyze this sum, we define the integer cutoff
$\kstar_{\stureg} \defn \lceil (\lambda\stureg)^{-1/(2(\alpha+\beta))}
\rceil$.  Given the scaling \mbox{$\eval_k \asymp k^{-2\alpha}$,} we
have $\eval_k \asymp \lambda$ for $k_\lambda \asymp \lambda^{-1/(2\alpha)}$.
The assumption $\stureg \le \lambda^{\beta/\alpha}$ implies that
\begin{align*}
\kstar_{\stureg} = (\lambda\stureg)^{-1/(2(\alpha+\beta))} \ge
\lambda^{-1/(2\alpha)} \asymp k_\lambda.
\end{align*}
Therefore, for all $k \ge \kstar_{\stureg}$, we have $\eval_k \lesssim
\lambda$, and hence
\begin{align*}
a_k = \frac{\evaltil_k\eval_k}{\eval_k+\lambda} \asymp
\frac{\evaltil_k\eval_k}{\lambda} \asymp
\lambda^{-1}k^{-2(\alpha+\beta)}.
\end{align*}

We now split the sum at $\kstar_{\stureg}$.  Since
$\frac{a_k^2}{(a_k+\stureg)^2} \le 1$ for all $k = 1, \ldots, D$, the
contribution from terms $k = 1, \ldots, \kstar_{\stureg}$ can be
bounded as
\begin{align*}
  \sum_{k=1}^{
    \kstar_{\stureg}}\frac{a_k^2}{(a_k+\stureg)^2}\frac{\evaltil_k}{\eval_k}
  & \leq \; \sum_{k=1}^{\kstar_{\stureg}} \frac{\evaltil_k}{\eval_k}.
\end{align*}
Using $\eval_k \asymp k^{-2\alpha}$ and $\evaltil_k \asymp
k^{-2\beta}$, we obtain $\frac{\evaltil_k}{\eval_k} \asymp
k^{2\alpha-2\beta}$, and therefore
\begin{align*}
\sum_{k=1}^{k_{\stureg}} \frac{\evaltil_k}{\eval_k} \le
C\int_1^{k_{\stureg}} k^{2\alpha-2\beta}dk \le C
(\kstar_{\stureg})^{2\alpha-2\beta+1},
\end{align*}
where we use $2\alpha-2\beta+1>0$, which follows from
$\beta\in(1/2,1/2+\alpha)$.

For the tail $k > \kstar_{\stureg}$, we use the bound
$\frac{a_k^2}{(a_k+\stureg)^2} \le \frac{a_k^2}{\stureg^2}$, so that
\begin{align*}
\sum_{k > \kstar_{\stureg}}
\frac{a_k^2}{(a_k+\stureg)^2}\frac{\evaltil_k}{\eval_k} \leq
\frac{1}{\stureg^2}\sum_{k > \kstar_{\stureg}}
a_k^2\frac{\evaltil_k}{\eval_k}.
\end{align*}
For $k > \kstar_{\stureg}$ we have $a_k \asymp \lambda^{-1}
k^{-2(\alpha+\beta)}$, whence $a_k^2 \frac{\evaltil_k}{\eval_k} \asymp
\lambda^{-2} k^{-4(\alpha+\beta)} k^{2\alpha-2\beta} = \lambda^{-2}
k^{-2\alpha-6\beta}$.  Putting together the pieces, it follows that
\begin{align*}
\sum_{k > \kstar_{\stureg}} a_k^2\frac{\evaltil_k}{\eval_k} & \leq
C\lambda^{-2}\int_{\kstar_{\stureg}}^\infty k^{-2\alpha-6\beta}dk \;
\leq \; C\lambda^{-2} (\kstar_{\stureg})^{-2\alpha-6\beta+1},
\end{align*}
where the integral is finite since $2\alpha+6\beta>1$.

Combining the two parts, we obtain the upper bound
\begin{align*}
\FinRatVar \leq \frac{C\sigma^2}{\numsource}\left((\kstar_{\stureg})^{2
  \alpha - 2 \beta +1} + \frac{1}{\lambda^2 \stureg^2}
(\kstar_{\stureg})^{-2\alpha-6\beta+1}\right).
\end{align*}
Substituting $\kstar_{\stureg} =
(\lambda\stureg)^{-1/(2(\alpha+\beta))}$ shows that the two terms are
of the same order, from which the claimed bound~\eqref{EqnVarBound}
follows.



\subsubsection{Proof of the \PL lower bound~\eqref{EqnPLShiftRateIncorrect}}
\label{SecProofPLLower}

We now prove the lower bound~\eqref{EqnPLShiftRateIncorrect} on the
\PL squared bias, as stated in~\Cref{ThmSeparation}.  We can formulate
the bias term of \PL in equation~\eqref{EqnPLBiasSplit} as
\begin{align}
  \PLBiasSquared(\fstar) 
  & =\big\|\lambda\SigmatTil^{1/2}\Sigmat_\lambda^{-1}v^* +
  \stureg\SigmatTil^{1/2}\SigmatTil_\stureg^{-1}\Sigmat\Sigmat_\lambda^{-1}v^*\big\|^2,
  \label{EqnNewPLBias}
\end{align}
which is derived in \Cref{SecProofFinPLBias}. 
  
Observe that both matrices $\Amat \defn
\lambda\SigmatTil^{1/2}\Sigmat_\lambda^{-1}$ and $\Bmat \defn
\stureg\SigmatTil^{1/2}\SigmatTil_\stureg^{-1}\Sigmat\Sigmat_\lambda^{-1}$
are positive semi-definite, symmetric, and commutative by assumption.
Consequently, for any vector $u$, we have $u^\top \Amat \Bmat u \geq
0$, and hence
\begin{align*}
\PLBiasSquared(\fstar) = \|(\Amat + \Bmat) v^*\|_2^2 & \geq \|\Amat
v^*\|_2^2 + \|\Bmat v^*\|_2^2 \\
& =
\underbrace{\big\|\lambda\SigmatTil^{1/2}\Sigmat_\lambda^{-1}v^*\big\|^2}_{\text{shift
    bias}} +
\underbrace{\big\|\stureg\SigmatTil^{1/2}\SigmatTil_\stureg^{-1}\Sigmat\Sigmat_\lambda^{-1}v^*\big\|^2}_{\text{ridge
    bias}}.
\end{align*}
We now take the supremum over $\|\fstar\|_\Hil \leq R$, or
equivalently over $\|\vstar\|_2 \leq R$ in the representation
$\fstar(x) = (\vstar)^T \phi(x)$.  In this way, we find that
\begin{align*}
\sup_{\|\fstar\|_\Hil \leq R} \PLBiasSquared(\fstar) & \geq
\sup_{\|v^*\|\leq R}
\big\|\lambda\SigmatTil^{1/2}\Sigmat_\lambda^{-1}v^* \big\|^2 = R^2
\underbrace{\opnorm{\lambda\SigmatTil^{1/2}\Sigmat_\lambda^{-1}}^2}_{\defn
  c_1} \; = c_1 \, R^2.
\end{align*}
Note that for any fixed $\lambda > 0$, the quantity $c_1 > 0$ is an
absolute constant, independent of $(n, R, \sigma, \stureg)$, which
completes the proof.

\subsubsection{Proof of the minimax-optimal lower bound~\eqref{EqnMinimaxShiftLower}}
\label{SecProofMinimaxLower}

We prove this lower bound via an Assouad construction, and using a
particular diagonal feature model. Let $\{e_j \}_{j=1}^\numsource$
denote the standard basis of $\real^{\numsource}$, and define a
feature map $\phi: \Xspace \rightarrow \real^\numsource$ such that
\begin{align*}
\phi(x_k)=\sqrt{\numsource}\,k^{-\alpha} e_k, \quad \mbox{and} \quad
\phi(\xtil_k)=\sqrt{\numsource}\,k^{-\beta} e_k.
\end{align*}
This feature map defines a finite-dimensional Hilbert space $\Hil$ of
functions $f_v(x) \defn \inner{v}{\phi(x)}$ with norm
$\hilnorm{f_v}=\norm{v}_2$, and kernel function $\Kerfun(x,x') =
\inprod{\phi(x)}{\phi(x')}$. By construction, with $\Qprob$ uniform
over $\{\xtil_j\}_{j=1}^m$ and $\Prob$ uniform over $\{x_i\}_{i=1}^n$,
the associated covariance operators are diagonal with eigenvalues
$k^{-2\alpha}$ and $k^{-2\beta}$, so that the kernel matrices
$(\Kmat_{nn}, \KmatTil_{mm})$ exhibit $(\alpha,\beta)$-eigendecay.

Consider the true function $\fstar(x) = f_{\vstar}(x)$.  Given our
diagonal construction, the source observations take the form $Y_k \sim
\mathcal{N}\big( \sqrt{\numsource} \, k^{-\alpha} \vstar_k,
\sigma^2)$.  For any estimator $\fhat$, define
$\vhat_k=\frac{k^\beta}{\sqrt{\numsource}}\fhat(\xtil_k)$. Then a
direct substitution yields $\qempnorm{\fhat-\fstar}^2 =
\Sum{k}{\numsource} k^{-2\beta}(\vhat_k-\vstar_k)^2$, where have used
orthogonality of the feature vectors. Thus, it suffices to lower bound
\begin{align}
\label{EqnLossDefinition}
\Loss(\vhat,\vstar) & \defn \sum_{k=1}^{\numsource} k^{-2 \beta}
(\vhat_k - \vstar_k)^2.
\end{align}
Thus, our problem has been reduced to estimating the vector $\vstar$
in a weighted Euclidean norm.

\paragraph{Lower bound via hypercube construction:}

We now construct a set of functions indexed by a Boolean vector $\tau
\in \{-1, 1\}^M$ for an integer $M \in [1, \numsource]$.  We make a
specific choice of $M$ later in the argument; in particular, see
equation~\eqref{EqnMchoice}.  Fixing $\eta = 1/4$, for each Boolean
vector $\tau \in \{-1, +1 \}^M$, we define the vector $\vtau \in
\real^\numsource$ with components
  \begin{align*}
    \vtau_k & \defn \begin{cases} \tau_k \delta_k & \mbox{for $k = 1, \ldots, M$} \\
      0 & \mbox{otherwise,}
    \end{cases}
  \end{align*}
  where $\delta_k = \eta \frac{\sigma}{\sqrt{\numsource}} k^\alpha$.
  We require $M \leq n$, and to be small enough so that
  \begin{align}
    \label{EqnMbound}
    \norm{\vtau}_2^2 & = \sum_{k=1}^M \delta_k^2 \; = \; \frac{\eta^2
      \sigma^2}{\numsource} \sum_{k=1}^M k^{2 \alpha} \leq R^2.
  \end{align}
This latter property ensures that each vector $\vtau$ indexes a
function within our class.  Letting $\Prob_\tau$ denote the law of $Y$
under $\vtau$, we have $Y \sim \mathcal{N}(\mu^\tau, \sigma^2 \IdMat)$
where $\mu^\tau_k \defn \sqrt{\numsource}\,k^{-\alpha} v^\tau_k$.

Our proof is based on two auxiliary claims.  First, for any valid $M$,
the worst-case estimation error over this family satisfies the lower
bound
\begin{subequations}
  \begin{align}
\label{EqnKeyLower}    
\sup_{\tau \in \{-1, +1\}^M} \Exs_\tau[\Loss(\vhat, v^\tau)] \ge
c_{\alpha,\beta} \frac{\sigma^2}{\numsource} M^{2\alpha-2\beta+1},
\end{align}
where $c_{\alpha, \beta} > 0$ is a constant depending only on the
exponents $(\alpha, \beta)$.  Second, the choice
\begin{align}
\label{EqnMchoice}  
M \defn \left \lfloor c_\ast \Bigl(\frac{\numsource
  R^2}{\sigma^2}\Bigr)^{\frac{1}{2\alpha+1}} \right \rfloor \qquad \mbox{for
  a sufficiently small $c_\ast > 0$}
\end{align}
\end{subequations}
is valid, meaning that $M \leq n$, and the bound~\eqref{EqnMbound}
holds.

Observe that the lower bound~\eqref{EqnMinimaxShiftLower} follows from
these two claims.  In particular, we substitute the
choice~\eqref{EqnMchoice} into inequality~\eqref{EqnKeyLower}.
Ignoring the constants, this yields a lower bound that scales as
\begin{align*}
  \frac{\sigma^2}{\numsource} M^{2\alpha-2\beta+1} \; = \;
  \frac{\sigma^2}{n} \Big(\frac{\numsource
    R^2}{\sigma^2}\Big)^{\frac{2 \alpha - 2 \beta + 1}{2\alpha+1}} & =
  R^2 \; \frac{\sigma^2}{n R^2} \Big(\frac{\numsource
    R^2}{\sigma^2}\Big)^{\frac{2 \alpha - 2 \beta + 1}{2\alpha+1}} \;
  = \; R^2 \Big(\frac{\sigma^2}{n R^2} \Big)^{\frac{2 \beta}{2 \alpha
      + 1}},
\end{align*}
as claimed in equation~\eqref{EqnMinimaxShiftLower}. \\

\noindent It remains to prove the two auxiliary results~\eqref{EqnKeyLower}
and~\eqref{EqnMchoice}. \\

\paragraph{Proof of the lower bound~\eqref{EqnKeyLower}:}

Given any estimate $\vhat$ of the vector $\vtau$, define the signs
$\widehat{\tau}_k = \operatorname{sign}(\vhat_k)$. If $\widehat{\tau}_k \neq
\tau_k$, then $|\vhat_k - v^\tau_k| \ge \delta_k$, whence $(\vhat_k -
v^\tau_k)^2 \geq \delta_k^2 \mathbf{1}\{\widehat{\tau}_k \neq
\tau_k\}$.  Summing over all indices $k=1,\dots,M$ yields the lower
bound
\begin{align*}
\Loss(\vhat, v^\tau) \ge \sum_{k=1}^M k^{-2\beta}
  \delta_k^2 \Exs_\tau \big[\mathbf{1}\{\widehat{\tau}_k \neq \tau_k\}
    \big]
\end{align*}
In order to apply Assouad's lemma~\cite{tsybakov2009introduction}, we
need to verify its testing conditions, which involve making single
flips to the Boolean vector.  For each $\tau \in \{-1,+1\}^M$ and any
$k \in [M]$, let $\tau^{(k)}$ denote the vector obtained by flipping
the $k$-th coordinate of $\tau$. Under our construction, the
Kullback--Leibler divergence can be upper bounded as $\KL(\Prob_\tau
\,\|\, \Prob_{\tau^{(k)}}) = \frac{1}{2\sigma^2}
\bigl(2\sqrt{\numsource}\,k^{-\alpha}\delta_k\bigr)^2 = 2\eta^2$.
Combining with Pinsker's inequality yields the TV upper bound
\begin{align*}
\TV(\Prob_\tau,\Prob_{\tau^{(k)}}) \leq \eta \; = \; 1/4.
\end{align*}
Applying Assouad's lemma with weights $\Delta_k =
k^{-2\beta}\delta_k^2$ then guarantees that
\begin{subequations}
\begin{align}
\label{EqnKento}  
\sup_{\tau \in \{-1,+1\}^M} \Exs_\tau \Loss(\vhat,v^\tau) & \geq
\frac{1-\eta}{2}\sum_{k=1}^M k^{-2\beta}\delta_k^2 \; = \;
\frac{3}{8}\sum_{k=1}^M k^{-2\beta}\delta_k^2,
\end{align}
where the last equality uses our choice $\eta = 1/4$.  Recalling our
choice $\delta_k^2 = \eta^2 \frac{\sigma^2}{\numsource} k^{2\alpha}$
with $\eta = 1/4$, we have
\begin{align}
\label{EqnMina}  
  \sum_{k=1}^M k^{-2\beta} \delta_k^2 = \eta^2
  \frac{\sigma^2}{\numsource} \sum_{k=1}^M k^{2\alpha-2\beta} \; = \;
  \frac{1}{16} \frac{\sigma^2}{\numsource} \sum_{k=1}^M
  k^{2\alpha-2\beta}.
\end{align}
\end{subequations}
Now our assumption that $2 \beta < 2 \alpha + 1$ implies that $2\alpha
- 2\beta > -1$, so that
\begin{align*}
  \sum_{k=1}^M k^{2\alpha-2\beta} \ge c_{\alpha,\beta}
  M^{2\alpha-2\beta+1} \qquad \mbox{ for some constant $c_{\alpha,
      \beta} > 0$.}
\end{align*}
Combining with the lower bounds~\eqref{EqnKento} and~\eqref{EqnMina}
yields the claim~\eqref{EqnKeyLower}.

\paragraph{Proof of validity of $M$:}  We now need to prove that
for a sufficiently small constant $c_\ast > 0$, the
choice~\eqref{EqnMchoice} of $M$ is valid, meaning that $M \leq n$,
and the bound~\eqref{EqnMbound} holds.  First, given our assumption
$\frac{R^2}{\sigma^2} \le n^{2\alpha}$, it follows that $M \leq c
\big(\frac{n R^2}{\sigma^2} \big)^{\frac{1}{2\alpha+1}} \leq c \;
c_\ast n \leq n$ as long as $c_\ast$ is sufficiently small.

As for the condition~\eqref{EqnMbound}, we can write
\begin{align*}
\norm{\vtau}_2^2 &= \sum_{k=1}^M \delta_k^2 = \frac{\eta^2
  \sigma^2}{\numsource} \sum_{k=1}^M k^{2\alpha} \; \le \;
\frac{\eta^2 \sigma^2}{\numsource} \left( 1 + \int_1^M x^{2\alpha} dx
\right) \le c_{\alpha} \frac{\eta^2 \sigma^2}{\numsource}
M^{2\alpha+1},
\end{align*}
for a constant $c_{\alpha} > 0$.  Substituting our
choice~\eqref{EqnMchoice} yields the upper bound
\begin{align*}
\norm{\vtau}_2^2 & \le c_{\alpha} \eta^2 \sigma^2 \cdot
\frac{1}{\numsource} \left( c_\ast \Bigl(\frac{\numsource
  R^2}{\sigma^2}\Bigr)^{\frac{1}{2\alpha+1}} \right)^{2\alpha+1} \; =
\; c_{\alpha} \eta^2 c_\ast^{2\alpha+1} R^2.
\end{align*}
Since $\eta = 1/4$ and $c_\alpha$ is universal, choosing $c_\ast$
sufficiently small ensures that $\norm{\vtau}_2^2 \le R^2$, as
claimed~\eqref{EqnMbound}.


\subsection{Proof of~\Cref{ThmComp}}
\label{SecProofThmComp}

We prove each of the two claims in turn.

\subsubsection{Proof of~\Cref{ThmComp}(a)}

Introduce the shorthand $\TopHat_\stepsize(f) = \PicardIter{f}$, so
that the algorithm generates the sequence $f^{k+1} =
\TopHat_\stepsize(f^k)$. Our proof hinges on the following claim: with
the stepsize $\stepsize = 1/\Liphat$, the update operator
$\TopHat_\stepsize$ satisfies the inequality
\begin{align}
\label{EqnApproxAverage}  
\|\TopHat_\stepsize(f) - \TopHat_\stepsize(\fhat)\|_2^2 & \leq \|f -
\fhat\|_2^2 - \frac{1}{2} \|(f - \TopHat_\stepsize(f)) - (\fhat -
\TopHat_\stepsize(\fhat)) \|_2^2 + 2 \numtarget
\frac{\epsstable^2}{\liphat^2},
\end{align}
where $\fhat$ is any \RAT fixed point.

Taking this inequality as given, let us complete the proof of the
claim~\eqref{EqnWeakPicard}.  We apply the
bound~\eqref{EqnApproxAverage} with $f = \funit{\kit}$, so that
$\TopHat_\stepsize(f) = \funit{\kit+1}$ by definition of the Picard
iteration.  Doing so, using the fact that $\fhat =
\TopHat_\stepsize(\fhat)$, and re-arranging yields
\begin{align*}
\frac{1}{2} \|\Defect(\funit{\kit})\|_2^2 = \frac{1}{2}
\|(\funit{\kit} - \TopHat_\stepsize(\funit{\kit}) \|_2^2 & \leq
\|\funit{\kit} - \fhat\|_2^2 - \|\funit{\kit +1} - \fhat\|_2^2 + 2
\numtarget \frac{\epsstable^2}{\liphat^2}
\end{align*}
Summing this recursion and using the telescoping property, we find that
\begin{align*}
  \frac{1}{2 (\Kmax +1)} \sum_{\kit = 0}^\Kmax
  \|\Defect(\funit{\kit})\|_2^2 & \leq \frac{\|\funit{0} -
    \fhat\|_2^2}{\Kmax + 1} + 2 \numtarget
  \frac{\epsstable^2}{\liphat^2}
\end{align*}
Noting that the minimum is less than the average and rescaling both
sides by $1/\numtarget$ to convert to the norm $\qempnorm{\cdot}$, the claim
follows. \\

\noindent It remains to prove our auxiliary claim.
\paragraph{Proof of the bound~\eqref{EqnApproxAverage}:}

In the standard Euclidean inner product and norm, the $(\liphat,
\epsstable)$-co-coercivity condition takes the form
\begin{align}
\label{EqnEuclidCoercive}
\inprod{f - \fhat}{\GradHat(f) - \GradHat(\fhat)} & \geq
\frac{1}{\liphat} \big \{ \|\GradHat(f) - \GradHat(\fhat) \|_2^2 -
\numtarget \epsstable^2 \big \}.
\end{align}
Introduce the shorthand $\SopHat(f) \defn f - \stepsize \GradHat(f)$.
We then have
\begin{align*}
\|\SopHat(f) - \SopHat(\fhat)\|_2^2 & = \|(f - \fhat) - \stepsize
(\GradHat(f) - \GradHat(\fhat)) \|_2^2 \notag \\
& = \|f - \fhat\|_2^2 + \stepsize^2 \|\GradHat(f) -
\GradHat(\fhat)\|_2^2 - 2 \stepsize \inprod{f - \fhat}{\GradHat(f) -
  \GradHat(\fhat)} \notag \\
& \stackrel{(i)}{\leq} \|f - \fhat\|_2^2 - \big \{ \tfrac{2}{ \stepsize
  \liphat} - 1 \big \} \stepsize^2 \|\GradHat(f) - \GradHat(\fhat)\|_2^2
+ 2 \tfrac{\stepsize \numtarget \epsstable^2}{\liphat} \notag \\
& \stackrel{(ii)}{=} \|f - \fhat\|_2^2 - \big \{ \tfrac{2}{ \stepsize
  \liphat} - 1 \big \} \|(\IdOp - \SopHat)(f) - (\IdOp - \SopHat)(\fhat)
\|_2^2 + 2 \tfrac{\stepsize \numtarget \epsstable^2}{\liphat}
\end{align*}
where step (i) uses the $(\liphat, \epsstable)$-co-coercivity
condition, and step (ii) uses the fact that $\IdOp - \SopHat =
\stepsize \GradHat$ by definition.  Setting $\stepsize = 1/\liphat$
yields
\begin{align}
\label{EqnSopBound}
\|\SopHat(f) - \SopHat(\fhat)\|_2^2 & \leq \|f - \fhat\|_2^2 -
\|(\IdOp - \SopHat)(f) - (\IdOp - \SopHat)(\fhat) \|_2^2 + 2
\numtarget \tfrac{\epsstable^2}{\liphat^2}
\end{align}

Next we use the bound~\eqref{EqnSopBound} to prove
inequality~\eqref{EqnApproxAverage}.  Since the proximal operator
$\ProxAlpha$ is firmly
non-expansive~\cite{parikh2014proximal,bauschke2017convex}, we have
\begin{align*}
\|\ProxAlpha(u) - \ProxAlpha(v)\|_2^2 & \leq \|u - v\|_2^2 - \|(\IdOp -
\ProxAlpha)(u) - (\IdOp - \ProxAlpha)(v)\|_2^2.
\end{align*}
We apply this inequality with $u = \SopHat(f)$ and $v =
\SopHat(\fhat)$.  These choices ensure that $\ProxAlpha(u) =
\TopHat_\stepsize(f)$ and $\ProxAlpha(v) = \TopHat_\stepsize(\fhat)$,
so that we obtain
\begin{align*}
\|\TopHat_\stepsize(f) - \TopHat_\stepsize(\fhat)\|_2^2 & \leq
\|\SopHat(f) - \SopHat(\fhat)\|_2^2 - \|(\IdOp - \ProxAlpha)(u) -
(\IdOp - \ProxAlpha)(v)\|_2^2.
\end{align*}
Combined with our earlier inequality~\eqref{EqnSopBound}, we have
\begin{align*}
  \|\TopHat_\stepsize(f) - \TopHat_\stepsize(\fhat)\|_2^2 & \leq \|f -
  \fhat\|_2^2 + 2 \numtarget \tfrac{\epsstable^2}{\liphat^2} -
  \DiffTerm,
\end{align*}
where $\DiffTerm \defn \| (f - \SopHat(f)) - (\fhat - \SopHat(\fhat))
\|_2^2 + \|(\SopHat(f) - \TopHat_\stepsize(f)) - (\SopHat(\fhat) -
\TopHat_\stepsize(\fhat)) \|_2^2$.  In order to complete the proof, it
suffices to show that
\begin{align*}
\DiffTerm & \geq \frac{1}{2} \|(f - \TopHat_\stepsize(f)) - (\fhat-
\TopHat_\stepsize(\fhat))\|_2^2.
\end{align*}
In terms of the shorthand $\Delta_1 = (f - \SopHat(f)) - (\fhat -
\SopHat(\fhat))$ and $\Delta_2 = (\SopHat(f) - \TopHat_\stepsize(f)) -
(\SopHat(\fhat) - \TopHat_\stepsize(\fhat))$, this inequality is
equivalent to showing that
\begin{align*}
\|\Delta_1\|_2^2 + \|\Delta_2\|_2^2 & \geq \frac{1}{2} \|\Delta_1 + \Delta_2\|_2^2
\end{align*}
which follows since $\inprod{\Delta_1}{\Delta_2} \leq \frac{1}{2}
\|\Delta_1\|_2^2 + \frac{1}{2} \|\Delta_2\|_2^2$ by Young's inequality.

\subsubsection{Proof of~\Cref{ThmComp}(b)}

We claim that with the stepsize $\stepsize =
\frac{\sconhat}{\liphat^2}$, the operator $\TopHat_\stepsize$ is
approximately contractive around any fixed point $\fhat$: in
particular, we have
\begin{subequations}
\begin{align}
\label{EqnCompContractive}
\qempnorm{\TopHat_\stepsize(f) - \TopHat_\stepsize(\fhat)}^2 & \leq
\contract \qempnorm{ f - \fhat}^2 + 4 (1 - \contract) \epsstable^2,
\end{align}
where $\contract \defn 1 - \tfrac{\sconhat^2}{\liphat^2}$.

We return to prove this property shortly.  Taking it as given, let us
prove the bound~\eqref{EqnStrongPicard}.  Applying the
inequality~\eqref{EqnCompContractive} repeatedly and unwinding the
recursion yields
\begin{align}
  \label{EqnVicuna}
  \qempnorm{\funit{\kit} - \fhat}^2 & \leq \contract^{\kit}
  \qempnorm{\funit{0} - \fhat}^2 + \frac{4 (1- \contract)
    \epsstable^2}{1 - \contract} \; = \; \contract^{\kit}
  \qempnorm{\funit{0} - \fhat}^2 + 4 \epsstable^2,
\end{align}
\end{subequations}
Noting that $\Defect(\funit{\Kmax}) = \funit{\Kmax} -
\TopHat_\stepsize(\funit{\Kmax}) = \funit{\Kmax} - \funit{\Kmax + 1}$,
we can write
\begin{align*}
\qempnorm{\Defect(\funit{\Kmax})}^2 & \stackrel{(i)}{\leq} 2 \big \{
\qempnorm{\funit{\Kmax} - \fhat}^2 + \qempnorm{\funit{\Kmax+1} -
  \fhat}^2 \big \} \; \stackrel{(ii)}{\leq} \; 2 \Big( 1 -
\tfrac{\sconhat^2}{\liphat^2} \Big)^\Kmax \qempnorm{\funit{0} - \fhat}^2
+ 8 \epsstable^2,
\end{align*}
as claimed in equation~\eqref{EqnStrongPicard}.  Here step (i) follows
from the triangle inequality, whereas step (ii) follows by applying
inequality~\eqref{EqnVicuna}.

\paragraph{Proof of the bound~\eqref{EqnCompContractive}:}

We first derive a consequence of the approximate co-coercivity
condition~\eqref{EqnEuclidCoercive}.  By re-arranging it, we can write
write
\begin{align*}
\|\GradHat(f) - \GradHat(\fhat)\|_2^2 & \leq \liphat \inprod{f -
  \fhat}{\GradHat(f) - \GradHat(\fhat)} + \numtarget \epsstable^2 \\
& \stackrel{(*)}{\leq} \tfrac{\liphat^2}{2} \|f - \fhat\|_2^2 +
\frac{1}{2} \|\GradHat(f) - \GradHat(\fhat) \|_2^2 + \numtarget \epsstable^2,
\end{align*}
where inequality $(*)$ follows from the Cauchy--Schwarz inequality.
Re-arranging yields
\begin{align}
  \label{EqnLlama}
\|\GradHat(f) - \GradHat(\fhat)\|_2^2 & \leq \liphat^2 \|f -
\fhat\|_2^2 + 2 \numtarget \epsstable^2.
\end{align}

We first show that the operator $\SopHat_\stepsize(f) = f - \stepsize
\GradHat(f)$ is $\gamma$-contractive in the given
sense~\eqref{EqnCompContractive}.  We write
\begin{align*}
  \|\SopHat_\stepsize(f) - \SopHat_\stepsize(\fhat)\|_2^2 & = \| (f -
  \fhat) - \stepsize (\GradHat(f) - \GradHat(\fhat)) \|_2^2 \\
 & = \|f - \fhat\|_2^2 + \stepsize^2 \|\GradHat(f) -
  \GradHat(\fhat)\|_2^2 - 2 \stepsize \inprod{f - \fhat}{\GradHat(f) -
    \GradHat(\fhat)} \\
& \leq \|f - \fhat\|_2^2 + \stepsize^2 \Liphat^2 \big \{ \|f -
  \fhat\|_2^2 + 2 \numtarget \epsstable^2 \big \} - 2 \stepsize \sconhat
  \big \{ \|f - \fhat\|_2^2 - \numtarget \epsstable^2 \big \},
\end{align*}
where the last step makes use of inequality~\eqref{EqnLlama}, along
with the $(\sconhat, \epsstable)$-approximate monotonicity
condition~\eqref{EqnApproxMonotone}).  Re-arranging terms yields
\begin{align*}
  \|\SopHat_\stepsize(f) - \SopHat_\stepsize(\fhat)\|^2 & \leq \big \{
  1 + \stepsize^2 \Liphat^2 - 2 \stepsize \sconhat \big \} \|f -
  \fhat\|^2 + 2 \big \{ \stepsize^2 \Liphat^2 + \stepsize \sconhat
  \big \} \numtarget \epsstable^2,
\end{align*}
and after setting $\stepsize = \sconhat/\Liphat^2$, we find that
\begin{align*}
  \|\SopHat_\stepsize(f) - \SopHat_\stepsize(\fhat)\|^2 & \leq \big \{
  1 - \tfrac{\sconhat^2}{\Liphat^2} \big \} \|f - \fhat\|^2 + 4
  \tfrac{\sconhat^2}{\Liphat^2} \epsstable^2 \; = \; \contract \|f -
  \fhat\|_2^2 + 4 (1 - \contract) \numtarget \epsstable^2,
\end{align*}
where we have made use of the definition $\contract = 1 -
(\sconhat/\liphat)^2$.  Since the proximal operator is non-expansive,
we have
\begin{align*}
  \|\TopHat_\stepsize(f) - \TopHat_\stepsize(\fhat) \|_2^2 \; = \;
  \|\ProxAlpha( \SopHat_\stepsize(f)) -
  \ProxAlpha(\SopHat_\stepsize(\fhat)) \|^2 & \leq
  \|\SopHat_\stepsize(f) - \SopHat_\stepsize(\fhat)\|_2^2 \\
  & \leq \contract \|f -
  \fhat\|_2^2 + 4 (1 - \contract) \numtarget \epsstable^2.
\end{align*}
Rescaling both sides by $1/\numtarget$ to convert to the
$\qempnorm{\cdot}$-norm completes the proof of the
bound~\eqref{EqnCompContractive}.


\section{Discussion}
\label{SecDiscussion}

We studied statistical estimation in a student--teacher setting, where
the predictions of a black-box predictive method (the teacher) are
used to guide training of a second model (the student), which might be
simpler, more interpretable, or better adapted to some form of
covariate shift.  We introduced a precise notion of a student oracle
estimand $\fdagger$, in particular via the fixed point of a proximal
update applied to a target population objective.  This perspective
naturally leads to the \emph{residual-as-teacher} (\RAT) estimator,
which mimics this proximal update by using the teacher to estimate the
student's residuals.  We proved various guarantees for this estimator,
including statistical bounds (\Cref{ThmStat}, \Cref{PropExactMSE}
and~\Cref{CorPLversusRat}), as well as computational bounds on an
iterative algorithm used to compute the \RAT fixed point
(\Cref{ThmComp}).  In addition, we also provided theory for, and
comparisons with, the standard student soft-matching (\SM) approach to
this problem.  This theory reveals that the two methods differ
fundamentally in how they handle bias or mis-specification present in
the teacher (cf.~\Cref{CorPLversusRat} and~\Cref{PropExactMSE}).
Moreover, for kernel-based student--teacher pairs, we established a
separation result (\Cref{ThmSeparation}): when the student
regularization parameter $\stureg$ is appropriately tuned, then the
\RAT estimator achieves the minimax-optimal rate, whereas the \SM
estimator is inconsistent for any choice of student tuning parameter.

Our work leaves open a number of interesting questions.  First, we
have shown that \RAT achieves minimax-optimal rates for a specific
class of student--teacher problems.  Whether \RAT remains
minimax-optimal beyond this class, particularly for more general
student--teacher pairs and non-kernel settings, is an important open
question.  Another important direction is to better understand the
statistical properties of teacher-based gradient estimation under
covariate shift, including the distinction between benign and malign
covariate shift.  We gave a precise characterization in one setting
(\Cref{ThmSeparation}), and suspect that qualitatively similar results
hold more generally.  Finally, a more general understanding of the
interaction between the student and teacher function classes would be
valuable.  Our results show that the ability of \RAT to mitigate bias
depends on this interaction, but a systematic characterization remains
open.

\subsubsection*{Acknowledgements}  This work was partially funded by National Science
Foundation Grant NSF DMS-2311072; Office of Naval Reseach ONR Grant
N00014026-1-2116, and the Ford Professorship to MJW.  
KY was supported by the Takenaka Scholarship Foundation.


\printbibliography

\appendix

\section{Proximal operators and fixed points}
\label{AppConsistency}

In this appendix, we collect together various properties of proximal
updates, pertinent to both the oracle function $\fdagger$ and the \RAT
estimate $\fhatrat$.  As in the proof of~\Cref{ThmStat}, we define the
function $\PenSmall(u) \defn \min \limits_{ \{ f \in \Fclass \mid
  f(\xtil_1^\numtarget) = u \} } \Pen(f)$.

\subsection{Proximal fixed point for $\fdagger$}
Let us clarify why $\fdagger$ satisfies the proximal fixed point
equation. Consider a function $\ftil$ that satisfies the proximal
fixed point relation $\ftil = \ProxAlpha \Big(\ftil(\xtil_1^m) -
\stepsize \nabla \LossBarM(\ftil) \Big)$.  Letting $\ztil \in \partial
\PenSmall(\ftil(\xtil_1^m))$, the optimality conditions associated
with the proximal update imply that $\nabla \LossBarM(\ftil) + \ztil =
0$.  But this is exactly the zero sub-gradient condition that defines
the student oracle estimand $\fdagger$.

\subsection{\RAT fixed points}
\label{SecRatFix}
Recall that we have defined the \RAT estimator via the fixed point
equation
\begin{align}
\label{EqnAppRat}
\fhatrat & = \ProxAlpha \Big( \fhatrat(\xtil_1^m) - \stepsize \GradHat(\fhatrat) \Big)
\end{align}
for some stepsize $\stepsize > 0$.  In order for this definition to be
meaningful, it should be the case that the set of fixed points is
independent of the choice of stepsize.  Here we establish this basic
fact.

Consider any function $\fhat$ that satisfies the fixed point
relation~\eqref{EqnAppRat} for some stepsize $\stepsize > 0$.
Expanding the definition of the proximal update, we have
\begin{align*}
\fhat & \in \arg \min_{f \in \Fclass} \Big \{ \frac{1}{2 \stepsize}
\|f(\xtil_1^m) - u\|_2^2 + \PenSmall(f(\xtil_1^m)) \Big \},
\end{align*}
where $u = \fhat(\xtil_1^m) - \stepsize \GradHat(\fhat)$.  The
zero-order optimality conditions for this optimization problem imply
that there exists some $\zhat \in \partial
\PenSmall(\fhat(\xtil_1^m))$ such that
\begin{align*}
0 & = \frac{1}{\stepsize} \big \{ \fhat(\xtil_1^m) - u \big \} + \zhat
\; = \; \GradHat(\fhat) + \zhat.
\end{align*}
Observe that the right-hand side is independent of $\stepsize$, so that the
set of fixed points does not depend on $\stepsize$, as claimed.


\section{Additional proofs}
\label{SecAdditionalProofs}

In this appendix, we collect together various proofs of auxiliary
results, including the proof of~\Cref{PropPL}
in~\Cref{SecProofPropPL}; the proof of~\Cref{CorPLversusRat}
in~\Cref{SecProofCorPLversusRat}; and the proof of~\Cref{PropExactMSE}
in~\Cref{SecProofPropExactMSE}.

  
\subsection{Proof of~\Cref{PropPL}}
\label{SecProofPropPL}
The claim of this proposition follows by arguments analogous to those
used to analyze the \RAT estimator in the proof of~\Cref{ThmStat}.
In particular, beginning with the definition~\eqref{EqnDefnPLEst} of
the \SM estimator, it follows that it satisfies a proximal fixed point
equation of the form
\begin{align}
  \label{EqnProxSM}
\fhatpl & = \ProxAlpha \big(\fhatpl(\xtil_1^m) - \stepsize \nabla
\LossPL(\fhatpl) \big).
\end{align}
Consequently, it can be viewed as a variant of the \RAT estimator, in
which the approximate gradient $\GradHat(f)$ used in \RAT is replaced
by $\nabla \LossPL(f)$.  The arguments used in the proof
of~\Cref{ThmStat} do not exploit any specific properties of the
gradient approximation used, as long as it satisfies a proximal fixed
point equation of the form~\eqref{EqnProxSM}.  Consequently, we can
recapitulate these same arguments, \emph{mutatis mutandis}, in order
to derive the analogous guarantees for the \SM estimator.


\subsection{Proof of~\Cref{CorPLversusRat}}
\label{SecProofCorPLversusRat}

We divide our proof into two parts, corresponding to each of the two
claims~\eqref{EqnSimplePLBound} and~\eqref{EqnSimpleRatBound}. Recall
that the least-squares loss is strongly convex with $\scon = 1$, so
that the bounds~\eqref{EqnPLEstBound} and~\eqref{EqnEstBound} are in
force for the \SM and \RAT estimators, respectively.  Moreover, for
the least-squares loss, the oracle gradient $\MyGrad(f) \in \real^m$
has entries
\begin{align}
\label{EqnAppOracleGradient}  
\big[ \MyGrad(f) \big]_j & = f(\xtil_j) - \Exs[Y \mid \xtil_j] \; = \;
f(\xtil_j) - \big \{\fdagger(\xtil_j) + \gstar(\xtil_j) \big \},
\end{align}
where we have inserted the decomposition $\Exs[Y \mid x] = \fdagger(x)
+ \gstar(x)$.

\paragraph{Proof of the \SM bound~\eqref{EqnSimplePLBound}:}
The least-squares loss is strongly convex with $\scon = 1$, so that
the bound~\eqref{EqnPLEstBound} is in force.  The \SM gradient
$\nabla \LossPL(f) \in \real^m$ takes the form
\begin{align*}
[\nabla \LossPL(f)]_j & = f(\xtil_j) - \yhat_j \quad \mbox{for $j = 1,
  \ldots, m$,}
\end{align*}
where $\yhat \defn \Teacher(y) \in \real^m$ are the pseudo-responses
computed by the teacher.  Substituting these choices into the
bound~\eqref{EqnPLEstBound} yields
\begin{align}
\label{EqnInter}  
  \qempnorm{\fhatpl - \fdagger}^2 & \leq \qempinner{\fhatpl -
    \fdagger}{\Teacher(y) - \big \{\fdagger(\xtil_1^m) +
    \gstar(\xtil_1^m) \big \} }.
\end{align}
We now decompose the difference $\Teacher(y) - \big \{
\fdagger(\xtil_1^m) + \gstar(\xtil_1^m) \big \}$ as the sum
\begin{align*}
  \underbrace{\Teacher \big(\fdagger(x_1^n) + \gstar(x_1^n) + w \big)
    - \TeacherBar(\fdagger + \gstar)}_{\equiv \newnoisepl} \; + \;
  \Big \{ \TeacherBar(\fdagger + \gstar) - \big [\fdagger(\xtil_1^m) +
    \gstar(\xtil_1^m) \big] \Big \}.
\end{align*}
Combined with Young's inequality, inequality~\eqref{EqnInter} then
implies that
\begin{align*}
\qempnorm{\fhatpl - \fdagger}^2 & \leq \qempinner{\fhatpl -
  \fdagger}{\newnoisepl} + \frac{1}{2} \qempnorm{\fhatpl - \fdagger}^2
+ \frac{1}{2} \NewBiasPL, 
\end{align*}
where $\NewBiasPL \defn \qempnorm{\TeacherBar(\fdagger(x_1^n) +
  \gstar(x_1^n)) - (\fdagger + \gstar)}^2$.  Re-arranging terms
completes the proof.

\paragraph{Proof of the \RAT bound~\eqref{EqnSimpleRatBound}:}
In this case, the bound~\eqref{EqnSimpleRatBound} is in force, and
again using the gradient representation~\eqref{EqnAppOracleGradient},
we arrive at the upper bound
\begin{align*}
\qempnorm{\fhatrat - \fdagger}^2 & \leq \qempinner{\fhatrat -
  \fdagger}{\fhatrat(\xtil_1^m) - \fstar(\xtil_1^m) -
  \GradHat(\fhatrat)}.
\end{align*}
Equivalently, in terms of the shorthand $\Delta = \fhatrat - \fdagger$
and the representation $\GradHat(\fhatrat) = \Teacher(\fhatrat(x_1^n)
- y)$, we have
\begin{align}
\label{EqnInterDelta}  
\qempnorm{\Delta}^2 & \leq \qempinner{\Delta}{\Teacher(\fhatrat(x_1^n)
  - y) + \Delta(\xtil_1^m) - \gstar(\xtil_1^m)}.
\end{align}
Now observing that $\Teacher(\fhatrat(x_1^n) - y) =
\Teacher(\Delta(x_1^n) - \gstar(x_1^n) - w)$, we decompose the
difference $\Teacher(\fhatrat(x_1^n) - y) + \Delta(\xtil_1^m) -
\gstar(\xtil_1^m)$ into the sum
\begin{align*}
\underbrace{\Teacher \big(\fhatrat(x_1^n) - y \big) -
  \TeacherBar(\Delta(x_1^n) - \gstar(x_1^n))}_{\equiv \newnoiserat} \;
\; + \; \; \Big \{ \TeacherBar(\Delta(x_1^n) - \gstar(x_1^n)) - \big
\{ \Delta(\xtil_1^m) - \gstar(\xtil_1^m) \big \} \Big \}.
\end{align*}
Substituting this decomposition into the upper bound~\eqref{EqnInterDelta}
and applying Young's inequality yields
\begin{align*}
\qempnorm{\Delta}^2 & \leq \qempinner{\Delta}{ \newnoiserat} +
\frac{1}{2} \qempnorm{\Delta}^2 + \frac{1}{2} \NewBiasRat,
\end{align*}
where $\NewBiasRat \defn \qempnorm{\TeacherBar(\Delta(x_1^n) -
  \gstar(x_1^n)) - (\Delta - \gstar)}^2$.  Re-arranging terms yields
the claim.

\subsection{Proof of~\Cref{PropExactMSE}}
\label{SecProofPropExactMSE}

We first derive the claimed forms of the \PL and \RAT weight vectors
$\thetahatpl$ and $\thetahatrat$, as given in
equations~\eqref{EqnThetaHatPL} and~\eqref{EqnThetaHatRAT},
respectively.  Recall that functions in the student class can be
written as $f_\theta(\cdot) = \sum_{j=1}^m \theta_j \KernelTil(\cdot,
\xtil_j)$ for some weight vector $\theta \in \real^m$.

\paragraph{Proof of relation~\eqref{EqnThetaHatPL}:}  By definition,
the PL estimate in this case performs $\stureg$-regularized KRR on the
teacher's output vector $\UseTeachMat(y) \in \real^m$.  By standard
results on kernel ridge regression, the resulting estimate is given by
the student function $f_{\thetahatpl}$, where the weight vector
takes the claimed form 
$\thetahatpl = \Big(\KmatTil_{mm} + m \stureg \IdMat)^{-1}
\UseTeachMat(y)$.

\paragraph{Proof of relation~\eqref{EqnThetaHatRAT}:}

Recall that in this section, the student penalty takes the form
$\Pen(f) = \frac{\numtarget \stureg}{2}\|f\|_\Hil^2$.  Consequently,
when the proximal operator $\ProxAlpha$ is applied to a vector $u \in
\real^m$, it returns the student function $f_v$, where the coefficient
vector $v \in \real^m$ is given by
\begin{align*}
v & \defn \big(\KmatTil_{\numtarget\numtarget} + \numtarget \stureg
\stepsize \IdMat_{\numtarget}\big)^{-1}u.
\end{align*}
Now let $\fhatrat \equiv f_{\thetahatrat}$ be a \RAT fixed point.  For
the least-squares loss, the source residual vector is $\fhatrat(x_1^n)
- y = \KmatTil_{\numsource\numtarget}\thetahatrat - y$, and hence the
teacher-estimated gradient is
\begin{align*}
\GradHat(\fhatrat) & =
\TeachMat\big(\KmatTil_{\numsource\numtarget}\thetahatrat - y\big).
\end{align*}
Substituting these relations into the fixed point
equation~\eqref{EqnFhatRatFix}, we find that
\begin{align}
\label{EqnRATUpdateTheta}
\thetahatrat & = \big(\KmatTil_{\numtarget\numtarget} + \numtarget
\stureg \stepsize \IdMat_{\numtarget}\big)^{-1} \Big(
\KmatTil_{\numtarget\numtarget}\thetahatrat - \stepsize
\TeachMat\big(\KmatTil_{\numsource\numtarget}\thetahatrat - y\big)
\Big).
\end{align}
Using the response-linearity of the teacher, this becomes
\begin{align*}
\thetahatrat & = \big(\KmatTil_{\numtarget\numtarget} + \numtarget
\stureg \stepsize \IdMat_{\numtarget}\big)^{-1} \Big(
\KmatTil_{\numtarget\numtarget}\thetahatrat - \stepsize \Aop
\thetahatrat + \stepsize \TeachMat(y) \Big),
\end{align*}
where we recall the definition $\Aop \defn \TeachMat
\KmatTil_{\numsource\numtarget}$.  Multiplying both sides by
$\KmatTil_{\numtarget\numtarget} + \numtarget \stureg \stepsize
\IdMat_{\numtarget}$ yields
\begin{align*}
\big(\KmatTil_{\numtarget\numtarget} + \numtarget \stureg \stepsize
\IdMat_{\numtarget}\big)\thetahatrat &=
\KmatTil_{\numtarget\numtarget}\thetahatrat - \stepsize \Aop
\thetahatrat + \stepsize \TeachMat(y).
\end{align*}
Cancelling the common term
$\KmatTil_{\numtarget\numtarget}\thetahatrat$ from both sides, we
obtain $\numtarget \stureg \stepsize \, \thetahatrat = -\stepsize \Aop
\thetahatrat + \stepsize \TeachMat(y)$.  Finally, dividing both sides
by $\stepsize > 0$ and rearranging yields
\begin{align*}
\big(\Aop + \numtarget \stureg \IdMat_{\numtarget}\big)\thetahatrat
&=
\TeachMat(y).
\end{align*}
Assuming that $\Aop + \numtarget \stureg \IdMat_{\numtarget}$ is
invertible, we conclude that $\thetahatrat = \big(\Aop + \numtarget
\stureg \IdMat_{\numtarget}\big)^{-1} \TeachMat(y)$, as
claimed in equation~\eqref{EqnThetaHatRAT}. \\


\noindent We now turn to the proofs of the exact MSEs given
in~\Cref{PropExactMSE}.

\paragraph{Proof of the MSE relation~\eqref{EqnMSEDecompPL}:}

By definition of the observation model, we have the relation $y =
\KmatTil_{\numsource \numtarget} \thetastar + w$.  Recalling our
shorthand notation $\Aop = \UseTeachMat \KmatTil_{nm}$ and
$\Bmat_\stureg = \KmatTil_{mm} + m \stureg \IdMat_m$, we can write
\begin{align*}
\thetahatpl - \thetastar & = \Bmat_\stureg^{-1} \UseTeachMat(y) -
\thetastar \; = \; \big \{ \Bmat_\stureg^{-1} \Aop - \IdMat_m \big \}
\thetastar + \Bmat_{\stureg}^{-1} \UseTeachMat w.
\end{align*}
Recall that $\qempnorm{\fhatpl - \fstar}^2  = (1/m) \|\KmatTil
(\thetahatpl - \thetastar)\|_2^2$.  We thus have
\begin{align*}
  \Exs_w \qempnorm{\fhatpl - \fstar}^2 & = \frac{1}{m} \| \KmatTil_{mm}
  \big \{\Bmat_\stureg^{-1} \Aop - \IdMat_m \big \} \thetastar \|_2^2
  + \frac{\sigma^2}{m} \frobnorm{\KmatTil_{mm} \Bmat_{\stureg}^{-1} \UseTeachMat}^2,
\end{align*}
using the fact that $\cov(w) = \sigma^2 \IdMat_n$.


\paragraph{Proof of the MSE relation~\eqref{EqnMSEDecompRAT}:}

Similarly, we can write
\begin{align*}
  \thetahatrat - \thetastar &= \big \{ (\Aop + \numtarget
  \stureg\IdMat)^{-1} \Aop - \IdMat \big \} \thetastar + (\Aop +
  \numtarget \stureg \IdMat)^{-1} \TeachMat(w).
\end{align*}
Using the identity $(\Aop + \numtarget\stureg \IdMat)^{-1} \Aop =
\IdMat - \numtarget\stureg(\Aop + \numtarget\stureg \IdMat)^{-1}$, we
have
\begin{align*}
  \thetahatrat - \thetastar = -\numtarget\stureg(\Aop +
  \numtarget\stureg \IdMat)^{-1} \thetastar + (\Aop +
  \numtarget\stureg \IdMat)^{-1} \TeachMat(w)
\end{align*}
We again use the relation $\qempnorm{\fhatrat - \fstar}^2 = (1/m)
\|\KmatTil_{mm} (\thetahatrat - \thetastar)\|_2^2$.  Computing the
squared bias and variance terms as before yields the claim.


\section{Auxiliary results for~\Cref{ThmSeparation}}

This appendix is devoted to various auxiliary results associated with
the separation result given in~\Cref{ThmSeparation}.

\subsection{Proof of equations~\eqref{EqnDefnFinRatBias} and~\eqref{EqnDefnFinRatVar}}
\label{SecProofFinRatDecomp}

Under the finite-dimensional feature representation introduced above,
the Gram matrices admit the forms
$\Kmat_{\numsource\numsource}=\PhiMat\PhiMat^\top$ and
$\KmatTil_{\numtarget\numsource}=\PhiMatTil\PhiMat^\top,\qquad
\KmatTil_{\numtarget\numtarget}=\PhiMatTil\PhiMatTil^\top$.
Substituting these expressions into the definitions of the operators
$\Aop$ and $\TeachMat_\lambda \TeachMat_\lambda^\top$ allows us to
express them in terms of the feature matrices.  A key step in these
calculations is the push-through identity
\begin{align}
  (\PhiMat\PhiMat^\top + \numsource \lambda
  \IdMat_{\numsource})^{-1}\PhiMat =\PhiMat (\PhiMat^\top \PhiMat +
  \numsource \lambda \IdMat_D)^{-1}.
\end{align}
We start from the definition $\TeachMat_\lambda =
\Kmat_{\numtarget\numsource}(\Kmat_{\numsource\numsource} + \numsource
\lambda \IdMat_{\numsource})^{-1}$.  Substituting the feature
representations and applying the push-through identity yields
\begin{align*}
\TeachMat_\lambda & = \PhiMatTil \PhiMat^\top (\PhiMat \PhiMat^\top +
\numsource\lambda\IdMat_{\numsource})^{-1} = \PhiMatTil (\PhiMat^\top
\PhiMat + \numsource \lambda\IdMat_D)^{-1} \PhiMat^\top.
\end{align*}
Using $\PhiMat^\top \PhiMat = \numsource \Sigmat$, we obtain the
compact representation
\begin{align}
  \TeachMat_\lambda =\frac{1}{\numsource} \PhiMatTil( \Sigmat +
  \lambda \IdMat_D)^{-1} \PhiMat^\top =\frac{1}{\numsource} \PhiMatTil
  \Sigmat_\lambda^{-1} \PhiMat^\top,
\end{align}
where we define the shorthand $\Sigmat_\lambda = \Sigmat + \lambda\IdMat_D$.

Similar calculations applied to the operator $\Aop$ and
$\KmatTil_{\numtarget\numtarget}(\Aop+\numtarget\stureg\IdMat_{\numtarget})^{-1}$
yield the representations
\begin{align*}
\Aop =\PhiMatTil\Sigmat\Sigmat_\lambda^{-1}\PhiMatTil^\top, \qquad
\KmatTil_{\numtarget\numtarget}(\Aop+\numtarget\stureg\IdMat_{\numtarget})^{-1}
=\frac{1}{\numtarget}\PhiMatTil\big(\SigmatTil\Sigmat\Sigmat_\lambda^{-1}+\gamma\IdMat_D\big)^{-1}\PhiMatTil^\top.
\end{align*}

Finally, substituting the above expressions into the definitions of
the bias and variance terms yields the desired representations in
equation~\eqref{EqnMSEDecompRAT}.  For the bias term, we have
\begin{align*}
  \NewBiasRat^2 & \defn \frac{1}{\numtarget}\| \numtarget\gamma
  \KmatTil_{\numtarget \numtarget} (\Aop + \numtarget \gamma
  \IdMat_{\numtarget})^{-1}\thetastar \|_2^2 \\
& = \frac{1}{\numtarget}\Big\| \gamma \PhiMatTil \big(\SigmatTil
  \Sigmat \Sigmat_\lambda^{-1}+\gamma\IdMat_D\big)^{-1}
  \PhiMatTil^\top \thetastar \Big\|_2^2
  \\
  & = \frac{\gamma^2}{\numtarget} (\thetastar)^\top \PhiMatTil \big(
  \SigmatTil\Sigmat\Sigmat_\lambda^{-1}+\gamma\IdMat_D\big)^{-\top}
  (\PhiMatTil^\top\PhiMatTil)
  \big(\SigmatTil\Sigmat\Sigmat_\lambda^{-1}+\gamma\IdMat_D\big)^{-1}
  \PhiMatTil^\top\thetastar \\
  & = \gamma^2 (\PhiMatTil^\top\thetastar)^\top
  \big(\SigmatTil\Sigmat\Sigmat_\lambda^{-1}+\gamma\IdMat_D\big)^{-\top}
  \SigmatTil
  \big(\SigmatTil\Sigmat\Sigmat_\lambda^{-1}+\gamma\IdMat_D\big)^{-1}
  (\PhiMatTil^\top\thetastar),
\end{align*}
where we used $\PhiMatTil^\top\PhiMatTil=\numtarget\SigmatTil$.
Defining $\vstar=\PhiMatTil^\top\thetastar$, this becomes
\begin{align*}
  \NewBiasRat^2 =\gamma^2 (\vstar)^\top
  \big(\SigmatTil\Sigmat\Sigmat_\lambda^{-1}+\gamma\IdMat_D\big)^{-\top}
  \SigmatTil
  \big(\SigmatTil\Sigmat\Sigmat_\lambda^{-1}+\gamma\IdMat_D\big)^{-1}
  \vstar \\ =\gamma^2\big\| \SigmatTil^{1/2}
  \big(\SigmatTil\Sigmat\Sigmat_\lambda^{-1}+\gamma\IdMat_D\big)^{-1}
  \vstar\big\|_2^2.
\end{align*}

Similarly, for the variance term, we have
\begin{align*}
\NewVarRat & \defn \frac{\sigma^2}{\numtarget} \frobnorm{
  \KmatTil_{\numtarget\numtarget}
  (\Aop+\numtarget\gamma\IdMat_{\numtarget})^{-1} \TeachMat_\lambda
}^2 \\
& = \frac{\sigma^2}{\numtarget} \frobnorm{
  \frac{1}{\numsource\numtarget} \PhiMatTil
  \big(\SigmatTil\Sigmat\Sigmat_\lambda^{-1}+\gamma\IdMat_D\big)^{-1}
  \PhiMatTil^\top \PhiMatTil \Sigmat_\lambda^{-1} \PhiMat^\top }^2
\\
& = \frac{\sigma^2}{\numtarget\,\numsource^2} \frobnorm{ \PhiMatTil
  \big(\SigmatTil\Sigmat\Sigmat_\lambda^{-1}+\gamma\IdMat_D\big)^{-1}
  \SigmatTil \Sigmat_\lambda^{-1} \PhiMat^\top }^2,
\end{align*}
where we used $\PhiMatTil^\top\PhiMatTil=\numtarget\SigmatTil$.

Now expand the Frobenius norm via $\|M\|_F^2=\Tr(M^\top M)$, from
cyclicity of trace,
\begin{align*}
\NewVarRat & = \frac{\sigma^2}{\numtarget\,\numsource^2} \Tr\Big(
\PhiMat^\top \PhiMat \Sigmat_\lambda^{-1} \SigmatTil
\big(\SigmatTil\Sigmat\Sigmat_\lambda^{-1}+\gamma\IdMat_D\big)^{-\top}
\PhiMatTil^\top \PhiMatTil
\big(\SigmatTil\Sigmat\Sigmat_\lambda^{-1}+\gamma\IdMat_D\big)^{-1}
\SigmatTil \Sigmat_\lambda^{-1} \Big) \\
& =\frac{\sigma^2}{\numsource} \Tr\Big( \Sigmat \Sigmat_\lambda^{-1}
\SigmatTil
\big(\SigmatTil\Sigmat\Sigmat_\lambda^{-1}+\gamma\IdMat_D\big)^{-\top}
\SigmatTil
\big(\SigmatTil\Sigmat\Sigmat_\lambda^{-1}+\gamma\IdMat_D\big)^{-1}
\SigmatTil \Sigmat_\lambda^{-1} \Big),
\end{align*}
where in the second line we again used
$\PhiMatTil^\top\PhiMatTil=\numtarget\SigmatTil$ and
$\PhiMat^\top\PhiMat=\numsource\Sigmat$.


\subsubsection{Proof of equation~\eqref{EqnNewPLBias}}
\label{SecProofFinPLBias}
Starting from \eqref{EqnPLBiasSplit}, we decompose the bias vector as
\begin{align*}
\PLBiasSquared(\fstar) =\frac{1}{\numtarget} \big\|
(\KmatTil_{\numtarget\numtarget}-\Aop)\thetastar + \numtarget\stureg
(\KmatTil_{\numtarget\numtarget}+\numtarget\stureg\IdMat_{\numtarget})^{-1}
\Aop\thetastar \big\|^2.
\end{align*}
Using the feature representations, this becomes
\begin{align}
  \PLBiasSquared(\fstar)
  &=\frac{1}{\numtarget}\big\|
    \PhiMatTil(\IdMat_D-\Sigmat\Sigmat_\lambda^{-1})\PhiMatTil^\top\thetastar
    + \numtarget\stureg(\PhiMatTil\PhiMatTil^\top +
    \numtarget\stureg\IdMat_{\numtarget})^{-1}\PhiMatTil\Sigmat\Sigmat_\lambda^{-1}\PhiMatTil^\top\thetastar
  \big\|^2 \nonumber \\
  &=\frac{1}{\numtarget}\big\|
    \PhiMatTil\big(
      \lambda\Sigmat_\lambda^{-1}
      + \stureg\SigmatTil_\stureg^{-1}\Sigmat\Sigmat_\lambda^{-1}
    \big)v^*
  \big\|^2 \nonumber \\
  &= (v^*)^\top
  \big(
      \lambda\Sigmat_\lambda^{-1}
      + \stureg\SigmatTil_\stureg^{-1}\Sigmat\Sigmat_\lambda^{-1}
  \big)^\top
  \SigmatTil
  \big(
      \lambda\Sigmat_\lambda^{-1}
      + \stureg\SigmatTil_\stureg^{-1}\Sigmat\Sigmat_\lambda^{-1}
  \big)v^* \\
  &=\big\|
    \SigmatTil^{1/2}
    \big(
      \lambda\Sigmat_\lambda^{-1}
      + \stureg\SigmatTil_\stureg^{-1}\Sigmat\Sigmat_\lambda^{-1}
    \big)v^*
  \big\|^2, \nonumber
\end{align}
where we used the identities
$(\PhiMatTil\PhiMatTil^\top+\numtarget\stureg\IdMat)^{-1}\PhiMatTil
=\PhiMatTil(\PhiMatTil^\top\PhiMatTil+\numtarget\stureg\IdMat)^{-1}$,
$\IdMat_D-\Sigmat\Sigmat_\lambda^{-1}=\lambda\Sigmat_\lambda^{-1}$,
and wrote $v^*=\PhiMatTil^\top\thetastar$ and
$\PhiMatTil^\top\PhiMatTil=\numtarget\SigmatTil$. This concludes the
proof of the bias representation~\eqref{EqnNewPLBias}.


\subsection{Proof of the bound~\eqref{EqnTechnical}}
\label{SecProofEqnTechnical}

Fix constants
$c_{\alpha,-},c_{\alpha,+},c_{\beta,-},c_{\beta,+}\in(0,\infty)$ such
that
\begin{align*}
c_{\alpha,-} k^{-2\alpha} \le \eval_k \le c_{\alpha,+} k^{-2\alpha},
\qquad c_{\beta,-} k^{-2\beta} \le \evaltil_k \le c_{\beta,+}
k^{-2\beta}
\end{align*}
for all $k = 1, \ldots, D$.  We now derive a uniform lower bound on
$a_k$.  If $\eval_k \ge \lambda$, then $\eval_k + \lambda \leq 2
\eval_k$, and hence $a_k = \frac{\evaltil_k \eval_k}{\eval_k +
  \lambda} \ge \frac{1}{2}\evaltil_k$.  If instead $\eval_k <
\lambda$, then $\eval_k + \lambda \le 2\lambda$, and hence $a_k =
\frac{\evaltil_k \eval_k}{\eval_k+\lambda} \ge \frac{\evaltil_k
  \eval_k}{2\lambda}$.  Therefore, for every $k = 1, \ldots, D$, we
have
\begin{align*}
a_k \ge \frac{1}{2} \min \!  \left\{ \evaltil_k, \frac{\evaltil_k
  \eval_k}{\lambda} \right\}.
\end{align*}
Substituting the eigendecay bounds gives
\begin{align*}
a_k \ge \frac{1}{2}
\min\!\left\{c_{\beta,-}k^{-2\beta},\frac{c_{\alpha,-}c_{\beta,-}}{\lambda}k^{-2(\alpha+\beta)}\right\}.
\end{align*}

Consequently, we have
\begin{align*}
\frac{\stureg\evaltil_k}{a_k+\stureg} \le
\min\!\left\{\evaltil_k,\frac{\stureg\evaltil_k}{a_k}\right\} \le
\min\!\left\{c_{\beta,+}k^{-2\beta},\frac{2c_{\beta,+}\stureg
  k^{-2\beta}}{\min\{c_{\beta,-}k^{-2\beta},(c_{\alpha,-}c_{\beta,-}/\lambda)k^{-2(\alpha+\beta)}\}}\right\}.
\end{align*}
Using $\min\{u,v\}\le u$ and $\min\{u,v\}\le v$, this implies
\begin{align*}
\frac{\stureg\evaltil_k}{a_k+\stureg} \le
\min\!\left\{c_{\beta,+}k^{-2\beta},\frac{2c_{\beta,+}}{c_{\beta,-}}\stureg,\frac{2c_{\beta,+}}{c_{\alpha,-}c_{\beta,-}}\lambda\stureg
k^{2\alpha}\right\}.
\end{align*}
Since the middle term is dominated by the maximum of the first and
third terms at the balancing scale, it suffices to bound
\begin{align*}
\frac{\stureg \evaltil_k}{a_k+\stureg} \le \min\!\left
\{c_{\beta,+}k^{-2\beta},\frac{2c_{\beta,+}}{c_{\alpha,-}c_{\beta,-}}\lambda\stureg
k^{2\alpha} \right\}.
\end{align*}

Define the functions $A_1(k) \defn c_{\beta,+} k^{-2\beta}$ and
$A_2(k) \defn \frac{2c_{\beta,+}}{c_{\alpha,-}c_{\beta,-}} \lambda
\stureg k^{2\alpha}$.  Observe that $A_1$ is decreasing whereas $A_2$
is increasing in $k$, so that the maximum of $\min\{A_1(k),A_2(k)\}$
is attained when the two terms are of comparable size.  Choose $k^* >
0$ satisfy $A_1(k^*)=A_2(k^*)$, so that
\begin{align*}
(k^*)^{2 (\alpha + \beta)} = \frac{c_{\alpha,-}
    c_{\beta,-}}{2}\frac{1}{\lambda\stureg}.
\end{align*}
Substituting this relation yields
\begin{align*}
A_1(k^*) = A_2(k^*) = c_{\beta,+}\Big(\frac{2}{c_{\alpha,-}
  c_{\beta,-}}\Big)^{\frac{\beta}{\alpha+\beta}}(\lambda
\stureg)^{\frac{\beta}{\alpha+\beta}}.
\end{align*}
Putting together the pieces yields the claim~\eqref{EqnTechnical}.



\section{Analysis for response-linear teachers}
\label{SecTeacherEffect}

In this section, we discuss techniques and typical scalings of the
stochastic error in the bound~\eqref{EqnSimpleUpper}.  In particular,
it involves the stochastic error term $\qempinner{\fhat -
  \fdagger}{\UseTeachMat(w)}$, with a term of this form shared by the
\RAT and \PL estimates.

For understanding, it is convenient to introduce the rescaled noise
variables $\newnoise_j \defn \sqrt{ \frac{n}{m}} [\TeachMat(w)]_j$ for
\mbox{$j = 1, \ldots, m$.}  We clarify the motivation for the
$\sqrt{n/m}$ rescaling in a moment.  With this definition, for any
function $f \in \Fstudent$, we have the equivalence
\begin{align}
\label{EqnStudentNoiseTerm}  
\qempinner{f - \fdagger}{\UseTeachMat(w)} & =
\frac{1}{\sqrt{\numobs}} \sum_{j=1}^m \newnoise_j
\frac{\big(f(\xtil_j) - \fdagger(\xtil_j) \big)}{\sqrt{m}}.
\end{align}
This term can be recognized as a variant of the standard noise
complexity involved in analysis of $M$-estimators.  It involves the
standard rescaling $1/\sqrt{\numobs}$ in source sample size $n$, but
differs in that the summation is over the $m$ target samples.
Nonetheless, we can bound it using standard techniques in empirical
process theory.  The novel and interesting ingredient is the structure
of the noise vector $\newnoise \in \real^m$, which has been
transformed by the teacher from the original vector $w \in
\real^\numobs$.

For a general student function class, it is standard to use
discretization arguments, via metric entropy or VC dimension, to
reduce the problem to studying a finite maximum.  Thus, to gain
intuition, it is natural to study the behavior of the student noise
term~\eqref{EqnStudentNoiseTerm} when the student function $f$ ranges
over a finite class $\{f^1, \ldots, f^N \}$, say with each function is
uniformly bounded as $\|f^\ell\|_\infty \leq 1$.  If the original
noise variables are i.i.d. $w_i \sim N(0, \sigma^2)$, the new noise
vector $\newnoise \in \real^m$ is Gaussian with covariance
\begin{subequations}
  \begin{align}
\label{EqnNewMat}    
  \NewMat \defn \frac{n}{m} \UseTeachMat \UseTeachMat^\top \in
  \real^{m \times m}
\end{align}
Introduce the shorthand $\Delta^\ell \defn \frac{1}{\sqrt{m}} \big(
f^\ell(\xtil_1^m) - \fdagger(\xtil_1^m) \big) \in \real^m$, and
observe that $\|\Delta^\ell\|_2 \leq \sqrt{2}$ by construction.  With
this notation, we have the bound
\begin{align}
\label{EqnSimpleBound}  
  \Exs \Big[ \qempinner{\fhat - \fdagger}{\UseTeachMat(w)} \Big] &
  \stackrel{(i)}{\leq} \max_{\ell=1, \ldots N} \sqrt{
    \inprod{\Delta^\ell}{\NewMat \Delta^\ell}} \; \sqrt{ \frac{2
      \sigma^2 \log N}{n}} \nonumber \\
  & \stackrel{(ii)}{\leq} 2 \sqrt{ \opnorm{\NewMat} \; \sigma^2 \;
    \frac{\log N}{n}},
\end{align}
\end{subequations}
where $\opnorm{\NewMat}$ is the maximum eigenvalue of $\NewMat$.  To
be clear, we note that both bounds (i) and (ii) are quite crude, and
moreover, the bound could be further improved by localization of the
empirical process.  However, we omit these technical refinements,
since our main goal is to understand the student-teacher interaction.

Thus, we are left to understand the structure of the covariance matrix
$\NewMat$, which depends entirely on the choice of rescaled teacher
$\sqrt{n/m} \, \UseTeachMat$.  To gain intuition, we discuss two
standard choices:

\begin{figure}[h]
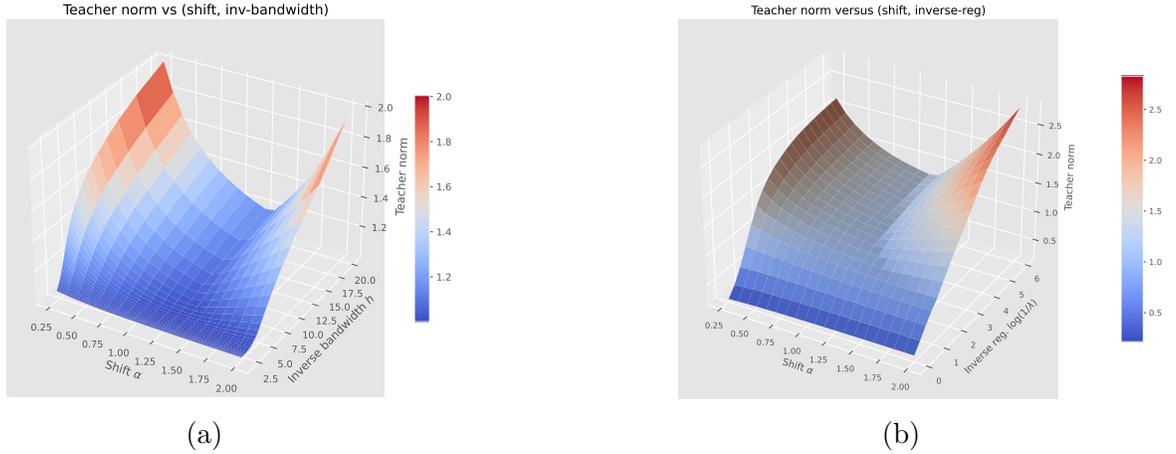

  \begin{center}
    \begin{tabular}{ccc}
      \widgraph{0.5\textwidth}{\myfigdir/fig_nw_teacher_opnorm} &&
      \widgraph{0.52\textwidth}{\myfigdir/fig_krr_teacher_opnorm}
      \\ (a) && (b)
    \end{tabular}
    \caption{Plots of the operator norm $\opnorm{\NewMat}$ from
      equation~\eqref{EqnNewMat} for two different classes of
      response-linear teachers, target distribution $\Qprob$ uniform
      on $[-1, 1]$, and source distributions $\Prob$ from a Beta
      distribution (shifted to $[-1,1]$) with parameter $\alpha \in
      [0.25, 2.0]$.  The setting $\alpha = 1$ corresponds to no
      covariate shift.  (a) Nadaraya--Watson smoother with bandwidth
      $h$.  Plots of $\opnorm{\NewMat}$ versus covariate shift
      $\alpha$ and inverse bandwidth $1/h$.  (b) Kernel ridge
      regression (KRR) teacher using a Gaussian kernel and
      regularization parameter $\lambda$.  Plots of $\opnorm{\NewMat}$
      versus covariate shift $\alpha$ and inverse regularization
      $1/\lambda$.  }
    \label{FigTeacherNorm}
  \end{center}
\end{figure}

\paragraph{Kernel ridge teacher:}
Recall the form~\eqref{EqnKRRTeacher} of the kernel ridge regression (KRR)
teacher with regularization parameter $\regpar > 0$.  Note that
\begin{align*}
\sqrt{\frac{\numobs}{m}} \UseTeachMat_\regpar & =
\frac{\Kmat_{mn}}{\sqrt{m \, n}} \Big( \frac{\Kmat_{nn}}{n} + \regpar
\IdMat \Big)^{-1}.
\end{align*}
The operator norm $\opnorm{\SigMat}$ corresponds to the squared
maximum singular value of this matrix, and by our choice of scaling
ensures that it is an order-one quantity in terms of the two sample
sizes $(n,m)$.  In particular, the $n$-dimensional matrix
$\tfrac{\Kmat_{nn}}{n}$ is an empirical approximation to the kernel
integral operator, so that by standard results, its spectrum will
converge to that of the kernel operator.  Similar comments apply to
the rescaled target-source matrix $\Kmat_{mn}/\sqrt{m n}$, which has
dimension $m \times n$, and approximates the target-source
cross-moment operator.

\paragraph{NW-smoothing teacher:}  The Nadaraya--Watson
smoother~\cite{watson1964smooth,nadaraya1964estimating} is another
standard example.  Let $\phi$ be a base-kernel function that
integrates to one, with a standard example being the Gaussian density
function.  For a bandwidth parameter $\bandwidth > 0$, the associated
teacher matrix has $(j,i)$-th entry given by
\begin{align}
\label{EqnNWTeacher}  
[\UseTeachMat_\bandwidth]_{ji} & = \frac{\phi_h(\xtil_j -
  x_i)}{\sum_{\ell=1}^\numobs \phi_\bandwidth(\xtil_j - x_\ell)}
\qquad \mbox{where $\phi_\bandwidth(t) \defn \frac{1}{\bandwidth}
  \phi(t/\bandwidth)$.}
\end{align}
Thus, the NW teacher matrix $\UseTeachMat_\bandwidth$ is
row-stochastic, and typically acts like a local averaging operator.

\end{document}

%% file: title_rat.tex
Residual-as-Teacher: Mitigating Bias Propagation in Student--Teacher Estimation

%% file: abstract_rat.tex
    We study statistical estimation in a \mbox{student--teacher}
    setting, where predictions from a pre-trained teacher are used to
    guide a student model. A standard approach is to train the student
    to directly match the teacher's outputs, which we refer to as
    student soft matching (\texttt{SM}).  This approach directly
    propagates any systematic bias or mis-specification present in the
    teacher, thereby degrading the student's predictions.  We propose
    and analyze an alternative scheme, known as residual-as-teacher
    (\texttt{RaT}), in which the teacher is used to estimate residuals
    in the student’s predictions.  Our analysis shows how the student
    can thereby emulate a proximal gradient scheme for solving an
    oracle optimization problem, and this provably reduces the effect
    of teacher bias.  For general \mbox{student--teacher} pairs, we
    establish non-asymptotic excess risk bounds for any \texttt{RaT}
    fixed point, along with convergence guarantees for the
    student-teacher iterative scheme.  For kernel-based
    student--teacher pairs, we prove a sharp separation: the
    \texttt{RaT} method achieves the minimax-optimal rate, while the
    \texttt{SM} method incurs constant prediction error for any sample
    size.  Experiments on both synthetic data and ImageNette
    classification under covariate shift corroborate our theoretical
    findings.